\documentclass[11pt]{article}

\usepackage[utf8]{inputenc}
\usepackage[T1]{fontenc}
\usepackage[margin=2.5cm]{geometry}
\usepackage{amsmath,amssymb}
\usepackage{graphicx}
\usepackage{booktabs}
\usepackage{multirow}
\usepackage{tabularx}
\usepackage{array}
\usepackage{enumitem}
\usepackage{tikz}
\usetikzlibrary{positioning,arrows.meta,calc,fit,backgrounds}
\usepackage{pgfplots}
\pgfplotsset{compat=1.18}
\usepgfplotslibrary{groupplots}
\usepackage{xcolor}
\usepackage[round,authoryear]{natbib}
\usepackage[hidelinks]{hyperref}

\graphicspath{{figures/}}



\title{How Much Do Foundation Models Add to Gradient Boosting on Financial Event Sequences?\\[0.5ex]
\large A Paired, Label-Budgeted Evaluation on Public and Proprietary Data,\\ and a Re-Examination of Two Public Benchmarks}
\author{Carlos Eduardo Martins Relvas\\WeCogno\\\texttt{carlos.relvas@wecogno.ai}
\and Gustavo Giovanni di Bernardo\\WeCogno\\\texttt{gustavo.bernardo@wecogno.ai}}
\date{}

\begin{document}

\maketitle

\begin{abstract}
Foundation models pretrained on raw transaction sequences are proposed for credit, fraud and churn modeling, with reported gains over gradient boosting, but the data and baselines behind these gains are rarely public.
We measure how much the pretraining and adaptation recipes of these models add to a baseline of point-in-time window aggregates and LightGBM, and we propose TBS (Temporal Boosting--Sequence ensemble), which averages the ranks of a boosting model enriched with label-free event-transition features and of a pretrained sequence network fine-tuned jointly with the aggregates.
On a synthetic credit-card dataset, the public MBD-mini banking dataset and two proprietary datasets of trade credit and B2B churn, we compare 27 methods at three label budgets with five seeds and paired bootstrap intervals, using sequence models of about 0.4 million parameters pretrained per dataset.
Frozen self-supervised embeddings rarely improve the baseline, and a fine-tuned sequence network alone clears it in 10 of 21 target--budget cells.
TBS clears it in 19 of 21 cells (18 after FDR adjustment), with a median gain of +2.1 AUC points, most of it from combining boosting with a fine-tuned sequence network and a smaller part from pretraining it.
Gains shrink as labels grow and with rich tabular data: with both, the gain on the proprietary datasets is +0.7 and +0.8 points, the latter with an interval including zero.
We also re-examine two benchmarks.
On IBM TabFormer, the test follows a change in the fraud pattern and the published baseline has no history; with the training period the vendor states, our history-aware LightGBM and our own foundation model stay below the published scores of a commercial model.
On IBM AML, a simulation artifact lets a date rule reach F1 0.482, and a point-in-time LightGBM reaches the F1 of graph neural networks whose test graph includes later transactions.

\end{abstract}

\section{Introduction}
\label{sec:introduction}

Credit scoring models decide, for a large number of people and firms, who obtains credit, how much and at what price~\citep{hand1997creditscoring}.
Small changes in these models can shift both the volume of losses of a lender and the distribution of credit across groups of borrowers~\citep{fuster2022predictably}.
The same holds for the models that guide collections, fraud prevention and customer retention, whose errors translate directly into financial losses or into the wrong treatment of a client.

The quality of such a model depends on several factors at once.
It depends on the data, that is, on which events of a client's history are recorded and on when each of them becomes known.
It depends on the modeling choices, from the way the history is summarized to the learning algorithm.
And it depends on the infrastructure that computes the inputs consistently at training and scoring time and keeps the model in production, a part of machine learning systems that is often more costly than the model itself~\citep{sculley2015debt}.

In practice, the history of events of a client, such as card transactions, invoices and payments or monthly revenue and support tickets, is usually summarized into window aggregates, such as counts, sums and averages over the last weeks or months~\citep{whitrow2009aggregation,bahnsen2016feature}.
These aggregates are then fed to a tree ensemble, a family of models that ranks among the strongest classifiers in credit scoring benchmarks~\citep{lessmann2015benchmark}.
This is also the starting point of the production models of the two companies whose data we use (Section~\ref{sec:setup}).
The approach works well, but it requires choosing by hand which aggregates to compute, and it can lose information that lies in the order and timing of the events.

Foundation models offer a different answer to the same problem.
They are models pretrained on large amounts of unlabeled data and then adapted to many downstream tasks~\citep{bommasani2021foundation}.
Applied to financial event sequences, they learn a representation of the raw history without labels, which promises to reduce the manual feature engineering, to use the unlabeled history of clients and to reuse a single model across credit, fraud and churn tasks~\citep{padhi2021tabbert,babaev2022coles}.
Financial institutions and vendors have recently reported gains of such models over gradient boosting on their internal tasks~\citep{braithwaite2026nuformer,ostroukhov2026pragma}, and foundation models are becoming a concrete option for the credit models of a financial institution.

These reports are difficult to act on.
The data and baselines of in-house studies are not public, so it is not possible to check how strong the baseline was or what information each model received.
The public benchmarks used in this line of work, IBM TabFormer~\citep{padhi2021tabbert} and IBM AML~\citep{altman2023aml}, are synthetic, and the comparisons published on them combine models evaluated by different authors under protocols that are not always described in full.
Public studies that compare pretrained sequence models with gradient boosting on aggregates~\citep{babaev2022coles,mollaev2025mbd,burakov2026eventgpt} feed the pretrained representation to the boosting model or average the two scores, but none of them combines a point-in-time protocol, paired uncertainty across label budgets and controls that separate the contribution of pretraining from that of combining two models.
Moreover, differences of one or two AUC points can be of the same order as the variation between random seeds~\citep{bouthillier2021variance}, and the value of pretraining is expected to depend on how many labels are available.
A practitioner deciding whether to adopt a foundation model therefore needs to know how much it adds to a strong aggregate-and-boosting model, at the number of labels they actually have, with an estimate of uncertainty.

This paper has two objectives.
The first is to measure how much the recipes of foundation models for financial event sequences, that is, self-supervised pretraining of a sequence model followed by its use as a frozen encoder or by fine-tuning on the target, add to a baseline of point-in-time window aggregates and gradient boosting, and how this contribution depends on the number of labels and on the tabular information already available.
The second is to propose TBS (Temporal Boosting--Sequence ensemble), a hybrid recipe that combines, at the level of their scores, a point-in-time boosting model enriched with event-transition features mined without the label and a pretrained sequence network fine-tuned jointly with the aggregates, and to identify through controlled comparisons which of its components explain its performance.
Each component of TBS has precedents, from blends of boosting and recurrent networks to the joint fusion of nuFormer and the rank averaging of a pretrained model with boosting~\citep{jxzly2022amex,braithwaite2026nuformer,burakov2026eventgpt}; what TBS adds is their combination, which keeps the boosting model that joint fusion cannot train end to end~\citep{braithwaite2026nuformer}, and its evaluation against controls that separate the contribution of pretraining from that of combining two models (Section~\ref{sec:related}).
As a complementary objective, we examine how the protocols of two public benchmarks affect the comparisons between foundation models and boosting published on them.

The study is deliberately restricted to a scale that a single team can run.
Our sequence models have about 0.4 million parameters and are pretrained per dataset on tens of thousands of clients, so we evaluate the pretraining and adaptation recipes of foundation models at this scale, and we do not test a large model pretrained once on a broad corpus and reused across tasks.

We organize the study around five research questions:
\begin{enumerate}[label=\textbf{RQ\arabic*}, leftmargin=*]
  \item\label{rq:frozen} Do pretrained representations, used as frozen features, add information to the aggregates used by the boosting model?
  \item\label{rq:finetune} How much does pretraining contribute to a sequence network fine-tuned on the target, and how does this network compare with the boosting model?
  \item\label{rq:hybrid} Does TBS improve on both of its components, the boosting model and the sequence network, and which of its components explain the gain?
  \item\label{rq:budget} How does the gain vary with the label budget and with the richness of the tabular attributes available to every method?
  \item\label{rq:protocol} How do protocol choices affect the conclusions drawn from the public TabFormer and IBM AML benchmarks?
\end{enumerate}

For each question we state a hypothesis, formulated after earlier results on these datasets (Section~\ref{sec:threats-internal}), so the study is exploratory:
\begin{enumerate}[label=\textbf{H\arabic*}, leftmargin=*]
  \item\label{h:frozen} Representations learned without the label, appended as frozen features in the configurations we evaluate, add little to the performance of a boosting model that already receives the window aggregates.
  \item\label{h:finetune} Fine-tuning the pretrained network on the target is worth more than using it frozen, pretraining improves the fine-tuned network over the same network trained from scratch, and the fine-tuned network alone does not consistently outperform the boosting model.
  \item\label{h:hybrid} Because the boosting model and the sequence network use the history in different ways, their errors are partly complementary, so TBS is more consistent than either of its components alone, and its gain depends on the network being pretrained rather than on the combination of two models by itself.
  \item\label{h:budget} The gain of the sequence models decreases as the number of labels grows and as the baseline receives richer tabular information.
  \item\label{h:protocol} Choices of training cut-off, of the information given to the baseline, of the decision threshold and of the information visible at test time can produce differences between models larger than those obtained under a point-in-time protocol.
\end{enumerate}

To examine these hypotheses, we compare 27 methods, from frozen self-supervised embeddings to fine-tuned sequence models, TBS and its variants and ablations, on four datasets: a synthetic credit-card dataset whose latent variables are known, the public MBD-mini banking dataset~\citep{mollaev2025mbd} and two proprietary datasets of commercial credit and B2B churn.
Every method is run at three label budgets with five seeds paired with the baseline, under a point-in-time, out-of-time protocol, and compared through a paired bootstrap of the test set.
We then re-examine the TabFormer and IBM AML benchmarks under the same principles and measure the compute cost of the models.
The authors work for a company that offers TBS commercially; the conflict of interest and its consequences for the interpretation of the results are discussed in Sections~\ref{sec:threats} and~\ref{sec:declarations}.

The rest of the paper is organized as follows.
Section~\ref{sec:background} introduces the concepts used throughout the paper, and Section~\ref{sec:related} reviews related work.
Section~\ref{sec:methodology} describes the evaluation methodology, Section~\ref{sec:models} the models and the comparisons between them, and Section~\ref{sec:setup} the datasets and the experimental setup.
Section~\ref{sec:results} presents the results, answering RQ1 to RQ4 on our four datasets, RQ5 through the re-examination of the public benchmarks, and closing with the compute cost of the models.
Section~\ref{sec:discussion} discusses the results, Section~\ref{sec:threats} the threats to validity, and Section~\ref{sec:conclusion} concludes.

\section{Background}
\label{sec:background}

This section introduces the concepts used in the rest of the paper.
It describes how a prediction problem on event histories is defined, how the histories are usually turned into tabular inputs, how they can instead be represented as sequences of tokens and encoded by a neural network, how such a network is pretrained and adapted to a target, and how two models are compared when their difference is small.
The specific choices made in this study are given in Sections~\ref{sec:methodology} to~\ref{sec:setup}.

\subsection{Event histories and point-in-time prediction}
\label{sec:bg-pit}

The unit of prediction is an entity, such as a client, a contract or an invoice, observed at a reference date that we call the as-of date.
The input is the history of events of the entity before that date, where each event is a record with a timestamp, categorical fields, such as a merchant group or a channel, and numeric fields, such as an amount.
The target is an outcome observed after the as-of date, such as a default, a late payment, a purchase or a cancellation.

A model is used in production at the as-of date, so its inputs may only contain what was known at that moment.
This requirement is called point-in-time correctness, and it is more subtle than it seems, because the date on which an event happens is not always the date on which it becomes known.
A payment is known on the day it is made, but a delinquency is known only when it is confirmed, and the revenue of a month only after the month closes.
Using the date an event refers to instead of the date it became known lets information from the future reach the inputs, a form of leakage that is a common source of inflated results in applied machine learning~\citep{kapoor2023leakage}.
The same reasoning applies to labels: a training label can only be used if it was already known when the model is trained, a condition we call label maturity.

Models are evaluated out of time, on as-of dates later than those used for training, and preferably out of sample, on entities that the model has not seen.
This reproduces the situation in which the model is deployed and exposes it to changes in the population over time.

\subsection{Window aggregates and gradient boosting}
\label{sec:bg-agg}

The most common way to turn an event history into a fixed-size input is to compute window aggregates~\citep{whitrow2009aggregation,bahnsen2016feature}.
For each numeric field and each time window before the as-of date, such as the last 30 or 90 days, one computes statistics such as the sum, the mean, the count, the minimum and the maximum, and for each categorical field one counts the events of each level.
The result is a table with one row per entity, which can be fed to any tabular model.

Gradient boosting builds an ensemble of decision trees, each fitted to correct the errors of the previous ones~\citep{friedman2001gbm}.
Efficient implementations such as LightGBM~\citep{ke2017lightgbm} handle hundreds of heterogeneous and partly redundant columns, missing values and monotone transformations without preprocessing.
On tabular data, tree ensembles remain hard to outperform with deep learning~\citep{grinsztajn2022tree,shwartzziv2022tabular}, and they rank among the strongest classifiers in credit scoring~\citep{lessmann2015benchmark}.

Window aggregates are robust and easy to audit, but they summarize each window as an unordered set of events.
Two clients with the same purchases and withdrawals in a month have the same aggregates whether the withdrawals came right after the purchases or weeks later, so a pattern defined by the order or by the timing of events cannot be expressed unless someone writes a feature for it.
Recovering this kind of information without hand-written features is the main promise of sequence models.

\subsection{Representing events as tokens}
\label{sec:bg-tokens}

A sequence model receives the history as an ordered list of vectors, one per element, in the same way as a language model receives a sentence as a list of tokens.
The first design decision is therefore what a token is and how it is turned into a vector, which is called tokenization.
In text, a token is a word or a piece of a word and has a single identity.
A financial event is different: it has several categorical fields, several numeric fields and a timestamp, and the relation between consecutive timestamps carries information of its own.

Two families of designs have been proposed for such records.
In the first, each event is serialized into a short text, such as ``merchant=grocery amount=120 channel=online'', and the text is tokenized as in a language model, so that every field becomes one or more tokens and an event spans several positions of the sequence~\citep{braithwaite2026nuformer}.
This design reuses the tooling of language models, but it makes sequences several times longer than the number of events.
In the second, each field is mapped to an embedding of its own, and the embeddings of the fields of an event are combined into a single vector, so that one event occupies one position of the sequence~\citep{padhi2021tabbert,luetto2023unittab}.
TabBERT combines the fields with a small Transformer applied inside each event, before a second Transformer is applied across events, a hierarchical design that lets the fields of an event interact before they are mixed with other events~\citep{padhi2021tabbert}.

This study follows the second family, with one event per position, which keeps the sequence length equal to the number of events and lets every model share the same input.
Numeric values and time are handled in two complementary ways.
They are discretized into quantile bands computed on training data, so that they can be embedded like categorical fields, and they are also kept as standardized continuous values after a logarithmic transform.
Time enters through three quantities: the hour of day, the gap since the previous event of the same entity, and the recency, that is, the time between the event and the as-of date.

Formally, let event $i$ have categorical fields $x_{i1},\dots,x_{iF}$, which include the bands of amount, gap and recency and the hour of day, and a vector $c_i$ of continuous values.
Each field $f$ has an embedding table $E_f$ with one row per level of its vocabulary, and the vector of the event is
\begin{equation}
  e_i = \sum_{f=1}^{F} E_f[x_{if}] + W c_i,
  \label{eq:event-embedding}
\end{equation}
where $W$ is a learned linear projection.
A learned positional embedding is added to $e_i$ before the sequence enters the encoder.
In the hierarchical variant, the sum in Equation~\eqref{eq:event-embedding} is replaced by a small Transformer over the field embeddings of the event, whose outputs are concatenated and projected back to the model width.

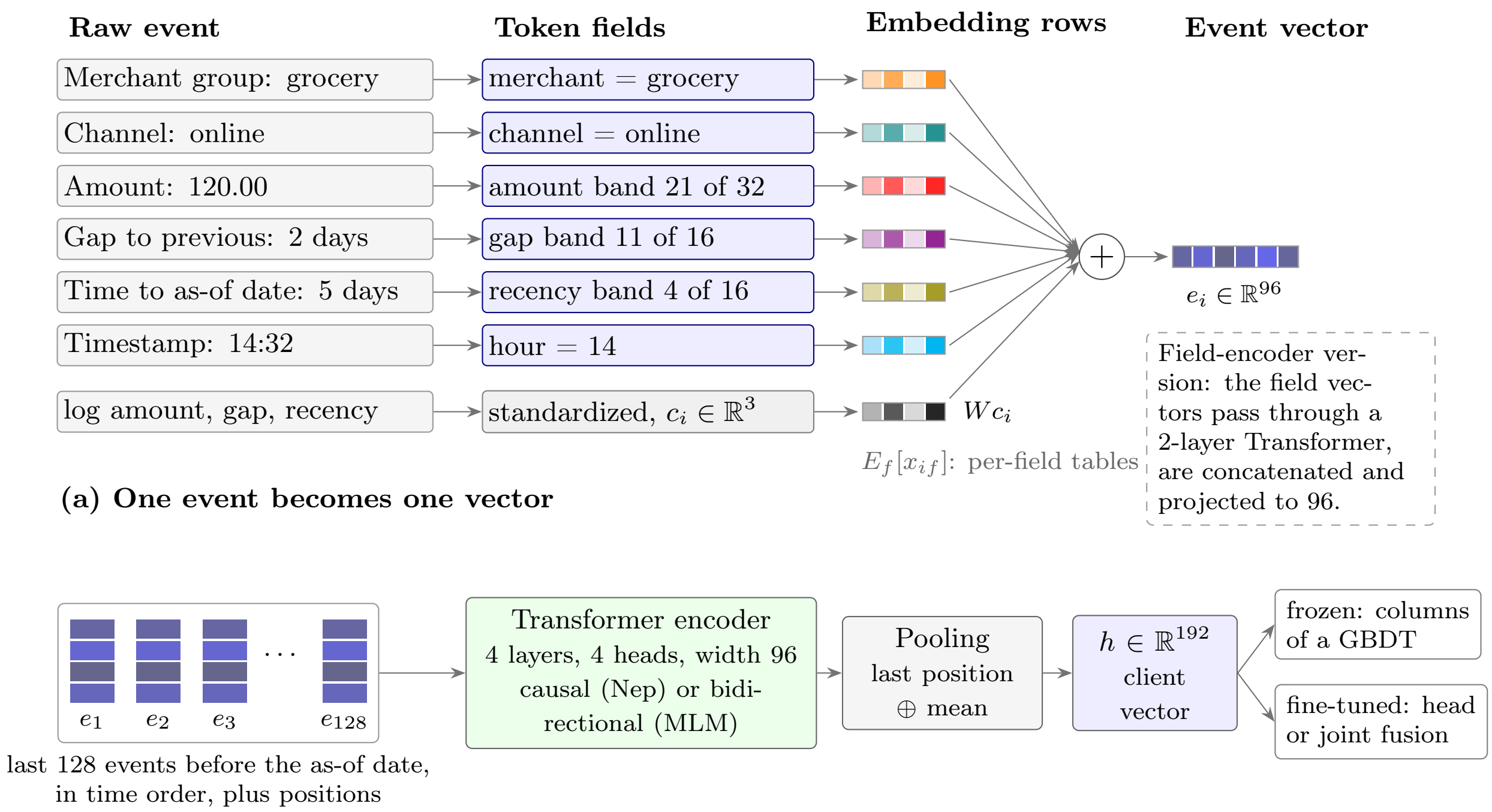
\begin{figure}[t]
  \centering
  \resizebox{\linewidth}{!}{\begin{tikzpicture}[
  font=\footnotesize, >=Stealth,
  raw/.style={draw=black!40, fill=black!4, rounded corners=1.5pt, text width=4.1cm, minimum height=0.46cm, anchor=west, align=left, inner sep=2pt},
  tok/.style={draw=blue!50!black, fill=blue!7, rounded corners=1.5pt, text width=3.6cm, minimum height=0.46cm, anchor=west, align=left, inner sep=2pt},
  box/.style={draw=black!60, rounded corners=2pt, align=center, minimum height=0.9cm},
  hdr/.style={font=\footnotesize\bfseries, anchor=south west},
  arr/.style={->, draw=black!55},
]
  \newcommand{\embvec}[2]{%
    \foreach \k/\s in {0/30,1/65,2/15,3/85}{\fill[#2!\s] ($(#1)+(\k*0.24,-0.1)$) rectangle ++(0.21,0.2);}
    \draw[black!40] ($(#1)+(0,-0.1)$) rectangle ++(0.93,0.2);}

  \node[hdr] at (0,0.35) {Raw event};
  \node[hdr] at (4.8,0.35) {Token fields};
  \node[hdr] at (9.0,0.35) {Embedding rows};
  \node[hdr] at (12.6,0.35) {Event vector};

  \foreach \i/\r/\t/\c in {
      0/{Merchant group: grocery}/{merchant = grocery}/orange,
      1/{Channel: online}/{channel = online}/teal,
      2/{Amount: 120.00}/{amount band 21 of 32}/red,
      3/{Gap to previous: 2 days}/{gap band 11 of 16}/violet,
      4/{Time to as-of date: 5 days}/{recency band 4 of 16}/olive,
      5/{Timestamp: 14:32}/{hour = 14}/cyan}{
    \node[raw] (r\i) at (0,-0.6*\i) {\r};
    \node[tok] (t\i) at (4.8,-0.6*\i) {\t};
    \draw[arr] (r\i.east) -- (t\i.west);
    \embvec{9.1,-0.6*\i}{\c}
    \draw[arr] (t\i.east) -- (9.05,-0.6*\i);
    \draw[arr] (10.08,-0.6*\i) -- (11.55,-1.95);
  }
  \node[raw] (r6) at (0,-3.75) {$\log$ amount, gap, recency};
  \node[tok, fill=black!6, draw=black!50] (t6) at (4.8,-3.75) {standardized, $c_i \in \mathbb{R}^3$};
  \draw[arr] (r6.east) -- (t6.west);
  \embvec{9.1,-3.75}{black}
  \node[anchor=west, font=\scriptsize] at (10.1,-3.75) {$W c_i$};
  \draw[arr] (t6.east) -- (9.05,-3.75);
  \draw[arr] (10.08,-3.6) -- (11.55,-2.05);
  \node[font=\scriptsize, black!60, anchor=north west] at (8.95,-4.05) {$E_f[x_{if}]$: per-field tables};

  \node[circle, draw=black!70, inner sep=1.2pt, font=\normalsize] (sum) at (11.8,-2.0) {$+$};
  \foreach \k/\s in {0/40,1/70,2/25,3/55,4/85,5/35}{\fill[blue!\s!black!60] ($(12.6,-2.0)+(\k*0.24,-0.12)$) rectangle ++(0.21,0.24);}
  \draw[black!40] (12.6,-2.12) rectangle ++(1.42,0.24);
  \node[font=\scriptsize] at (13.3,-2.42) {$e_i \in \mathbb{R}^{96}$};
  \draw[arr] (sum.east) -- (12.55,-2.0);
  \node[draw=black!40, dashed, rounded corners=2pt, align=left, font=\scriptsize, text width=3.0cm, anchor=north west] at (12.3,-2.85)
    {Field-encoder version: the field vectors pass through a 2-layer Transformer, are concatenated and projected to 96.};
  \node[font=\footnotesize\bfseries, anchor=west] at (-0.1,-4.75) {(a) One event becomes one vector};

  \begin{scope}[yshift=-6.3cm]
    \foreach \j/\lab in {0/1,1/2,2/3} {
      \foreach \k/\s in {0/40,1/70,2/25,3/55}{\fill[blue!\s!black!60] ($(0.15+\j*0.75,0.2)+(0,-\k*0.24)$) rectangle ++(0.5,-0.21);}
      \node[font=\scriptsize] at (0.4+\j*0.75,-0.95) {$e_{\lab}$};
    }
    \node at (2.55,-0.2) {$\cdots$};
    \foreach \k/\s in {0/40,1/70,2/25,3/55}{\fill[blue!\s!black!60] ($(3.0,0.2)+(0,-\k*0.24)$) rectangle ++(0.5,-0.21);}
    \node[font=\scriptsize] at (3.25,-0.95) {$e_{128}$};
    \node[draw=black!40, rounded corners=2pt, fit={(0.05,0.35) (3.6,-1.15)}, inner sep=1pt] (seq) {};
    \node[font=\scriptsize, align=center, anchor=north] at (seq.south) {last 128 events before the as-of date,\\in time order, plus positions};

    \node[box, fill=green!8, text width=3.7cm] (enc) at (6.6,-0.4) {Transformer encoder\\{\scriptsize 4 layers, 4 heads, width 96}\\{\scriptsize causal (Nep) or bidirectional (MLM)}};
    \node[box, fill=black!4, text width=2.0cm] (pool) at (10.0,-0.4) {Pooling\\{\scriptsize last position $\oplus$ mean}};
    \node[box, fill=blue!7, text width=1.6cm] (h) at (12.4,-0.4) {$h \in \mathbb{R}^{192}$\\{\scriptsize client vector}};
    \draw[arr] (seq.east) -- (enc.west);
    \draw[arr] (enc.east) -- (pool.west);
    \draw[arr] (pool.east) -- (h.west);
    \node[draw=black!50, rounded corners=2pt, font=\scriptsize, align=left, anchor=west] (fr) at (13.75,0.15) {frozen: columns\\of a GBDT};
    \node[draw=black!50, rounded corners=2pt, font=\scriptsize, align=left, anchor=west] (ft) at (13.75,-0.95) {fine-tuned: head\\or joint fusion};
    \draw[arr] (h.east) -- (fr.west);
    \draw[arr] (h.east) -- (ft.west);
    \node[font=\footnotesize\bfseries, anchor=west] at (-0.1,-2.45) {(b) The sequence of event vectors becomes one client vector};
  \end{scope}
\end{tikzpicture}}
  \caption{From events to a client vector. (a) Each field of an event is mapped to a token, using training-period vocabularies and quantile bands, and each token selects a row of its embedding table; the rows and a projection of the continuous values are summed into the event vector (Equation~\ref{eq:event-embedding}), or combined by a small Transformer in the field-encoder version. Values are illustrative. (b) The last 128 event vectors, with their positions, pass through the Transformer encoder, and the outputs are pooled into one vector that is used frozen or fine-tuned.}
  \label{fig:tokenization}
\end{figure}

Figure~\ref{fig:tokenization} summarizes the whole path, from one event to its vector and from the sequence of event vectors to a single vector per client.

Table~\ref{tab:token-example} illustrates the procedure on a card purchase.
The values are illustrative and the band indices depend on the quantiles of each dataset.
Each categorical field contributes one row of its embedding table, the three continuous values contribute a linear projection, and the sum is a single vector of the model width that represents the event.

\begin{table}[ht]
  \centering
  \caption{Illustrative tokenization of one card purchase into a single event vector. Values and band indices are examples; bands are computed from training-period quantiles of each dataset.}
  \label{tab:token-example}
  \small
  \begin{tabular}{llll}
    \toprule
    Raw attribute & Raw value & Token field & Contribution to $e_i$ \\
    \midrule
    Merchant group & grocery & merchant group = grocery & row of $E_{\text{merchant}}$ \\
    Channel & online & channel = online & row of $E_{\text{channel}}$ \\
    Amount & 120.00 & amount band 21 of 32 & row of $E_{\text{amount}}$ \\
    Gap to previous event & 2 days & gap band 11 of 16 & row of $E_{\text{gap}}$ \\
    Time to the as-of date & 5 days & recency band 4 of 16 & row of $E_{\text{recency}}$ \\
    Timestamp & 14:32 & hour = 14 & row of $E_{\text{hour}}$ \\
    \midrule
    \multicolumn{2}{l}{$\log$ amount, gap and recency, standardized} & continuous vector $c_i$ & $W c_i$ \\
    \midrule
    \multicolumn{3}{l}{Event vector} & $e_i$, the sum of all rows above \\
    \bottomrule
  \end{tabular}
\end{table}

\subsection{Sequence encoders}
\label{sec:bg-encoders}

A sequence encoder maps the list of event vectors to a representation of the whole history.
The Transformer~\citep{vaswani2017attention} does so with self-attention, in which each position builds its new vector as a weighted combination of the vectors of the other positions, with weights that depend on their content.
With bidirectional attention, each event sees the whole history; with causal attention, each event sees only the events before it, which is what a model needs when it is trained to predict the next event.
Recurrent networks such as the GRU~\citep{cho2014gru} instead read the sequence one event at a time and update a hidden state, and remain strong supervised baselines on event sequences~\citep{osin2025ebes}.

Self-attention compares every position with every other, so its cost grows with the square of the sequence length.
For this reason, and because recent events usually carry most of the information about a near-term outcome, histories are often truncated to their most recent events before they enter the encoder.
Self-attention also ignores the order of its inputs, so the position of each event is supplied explicitly, for example by adding a learned positional embedding to its vector, and the time between events can be supplied through the tokens themselves, as described above.
To obtain one vector per entity, the outputs of the encoder are pooled, for example by taking the output at the last position, the mean over all positions, or both.
This vector is the representation of the history that is either used as a frozen feature or passed to a prediction head, as described in Section~\ref{sec:bg-adaptation}.

\subsection{Self-supervised pretraining and foundation models}
\label{sec:bg-pretraining}

A foundation model is a model pretrained on a large amount of data without task labels and later adapted to many tasks~\citep{bommasani2021foundation}.
Pretraining relies on a self-supervised objective, a task whose answer can be computed from the data itself.
Three objectives are common for event sequences.
In masked modeling, some events or fields are hidden and the model learns to recover them from the rest of the sequence, as in masked language models~\citep{devlin2019bert} and in TabBERT and PRAGMA~\citep{padhi2021tabbert,ostroukhov2026pragma}.
In next-event prediction, a causal model learns to predict every field of the next event from the events before it, which is the objective of autoregressive language models~\citep{radford2018gpt} and of nuFormer~\citep{braithwaite2026nuformer}.
In contrastive learning, two views of the same entity, such as two sub-sequences of its history, are pulled together in the representation space and views of different entities are pushed apart~\citep{oord2018cpc,babaev2022coles}.
A simpler and older practice is to train an autoencoder that compresses the history into a vector and reconstructs it.

None of these objectives uses the target, so the entities without labels, and the history of labeled entities before the training period, can all be used for pretraining.
This is the main reason to expect pretraining to help when labels are scarce.

\subsection{Adapting a pretrained model to a target}
\label{sec:bg-adaptation}

A pretrained encoder can be used in two ways.
In the frozen use, the encoder is kept fixed and its representation of each entity is appended as extra columns to the inputs of another model, such as a boosting model on the aggregates.
In fine-tuning, the encoder is trained further on the labeled data together with a prediction head, so that its representation adapts to the target.
Joint fusion is a form of fine-tuning in which the sequence encoder and a network over tabular features are trained together on the label, so that the model sees both the raw history and the aggregates~\citep{braithwaite2026nuformer}.

A pretrained model can also be combined with a boosting model instead of replacing it.
A simple combination averages the ranks of the scores of the two models, which requires no training and puts both scores on the same scale, at the cost of not producing a calibrated probability.
A second combination uses the score of the boosting model as a fixed offset and trains the sequence network to add a correction to it, so that the combined model starts from the boosting score and moves away from it only where the sequence brings information.

Pretraining is not restricted to sequences.
TabPFN~\citep{hollmann2025tabpfn} is a network pretrained on a large number of synthetic tabular problems that predicts on a new table in a single forward pass, without training on it, and is competitive on small datasets.
Applied to window aggregates, it provides a pretrained alternative to boosting that does not use the order of events.

\subsection{Comparing models with small differences}
\label{sec:bg-evaluation}

The models in this study are binary classifiers evaluated by their ability to rank positives above negatives.
The area under the ROC curve (AUC) is the probability that a random positive receives a higher score than a random negative; in credit, it is often reported as the Gini coefficient, equal to $2\,\mathrm{AUC}-1$, or complemented by the Kolmogorov--Smirnov statistic.
When positives are rare, as in fraud, average precision, which summarizes the precision--recall curve, is more informative.

The differences between strong models on the same task are often of one or two AUC points, which can be of the same order as the variation caused by the random seed, the subsample of labels or the split used for early stopping~\citep{bouthillier2021variance}.
Two practices make such differences measurable.
The first is pairing: both models are trained with the same seeds and evaluated on the same test rows, so that the variation they share cancels out in their difference.
The second is the bootstrap~\citep{efron1993bootstrap}, which resamples the test rows with replacement, recomputes the difference on each resample and uses the spread of the resampled differences as a confidence interval.
When many comparisons are made, some will appear significant by chance, and procedures such as that of Benjamini and Hochberg~\citep{benjamini1995fdr} control the expected share of false discoveries among the comparisons declared significant.

\section{Related Work}
\label{sec:related}

This study draws on four lines of work: foundation models for financial event sequences, which are the models whose added value we measure; studies and benchmarks that compare sequence models with gradient boosting; hybrids of boosting and neural networks, the family to which TBS belongs; and work on evaluation practice, which motivates our protocol and our re-examination of two public benchmarks.
The section closes with a comparison of the closest studies with ours.

\subsection{Foundation models for financial event sequences}
\label{sec:rw-fm}

TabBERT~\citep{padhi2021tabbert} introduced a hierarchical Transformer for sequences of tabular records, with a field-level encoder inside each transaction and a sequence-level encoder across transactions, pretrained by masked field modeling, and released with it the synthetic TabFormer card dataset.
UniTTab~\citep{luetto2023unittab} and FATA-Trans~\citep{zhang2023fatatrans} extended the design to heterogeneous records and to time-aware attention.
CoLES~\citep{babaev2022coles} took a contrastive approach, learning client representations by bringing together sub-sequences of the same client.
Generative pretraining on transactions was explored by \citet{skalski2023foundation}, who trained an autoregressive model on card purchases as a step towards a foundation purchasing model, and \citet{shestov2025llm4es} used large language models to embed event sequences.

More recently, financial institutions have described foundation models trained on their own data.
nuFormer~\citep{braithwaite2026nuformer} serializes transactions into text tokens, pretrains a causal Transformer by next-token prediction and fine-tunes it jointly with a tabular network, a recipe its authors call joint fusion, and reports gains over a LightGBM production model on internal tasks.
PRAGMA~\citep{ostroukhov2026pragma} pretrains masked models with up to a billion parameters on a large transaction corpus and adapts them with linear probes and fine-tuning, and TREASURE~\citep{yeh2026treasure} describes a Transformer-based foundation model for payment transactions.
EventGPT~\citep{burakov2026eventgpt} pretrains a hierarchical causal Transformer by next-event prediction on the public Multimodal Banking Dataset and evaluates it zero-shot, through frozen embeddings and in an ensemble with gradient boosting.

These models differ in tokenization, objective and scale, but most of the evidence on their value comes from data and baselines that are not public, which makes it difficult to know how much they add to a well-built aggregate-and-boosting model.
We do not reproduce any of these systems; we implement their pretraining and adaptation recipes at a common small scale, with a shared tokenizer and backbone, so that differences come from the recipe rather than from capacity (Section~\ref{sec:models}).

\subsection{Sequence models and gradient boosting on financial data}
\label{sec:rw-comparisons}

Several studies have compared sequence models with gradient boosting on aggregates.
\citet{babaev2019etrnn} trained a recurrent network on raw card transactions for credit scoring and compared it with LightGBM on thousands of aggregates, including learning curves over the size of the training set.
CoLES compares its embeddings, fed to LightGBM, with LightGBM on hand-crafted aggregates on public datasets that include a credit default task, and also in a semi-supervised setup in which only a fraction of the labels is used~\citep{babaev2022coles}.
\citet{jurgovsky2018sequence} found that an LSTM and a random forest on aggregated features detect different frauds.
nuFormer compares joint fusion with LightGBM and with early and late fusion, in which pretrained embeddings are fed to a downstream model~\citep{braithwaite2026nuformer}.
EventGPT compares frozen embeddings with CatBoost on engineered features and reports that the embeddings alone score slightly below the features~\citep{burakov2026eventgpt}, while \citet{butvinik2026tct} report no gain from adding contrastive sequence embeddings to the engineered features of a boosting model for financial crime detection.

Benchmarks of event-sequence models follow a different path.
EBES~\citep{osin2025ebes} compares sequence models under a common protocol and finds that simple supervised recurrent models are hard to beat and that contrastive pretraining helps inconsistently, without a boosting baseline.
The Multimodal Banking Dataset~\citep{mollaev2025mbd} provides public banking sequences with product-propensity targets, together with baselines that include gradient boosting on aggregated features and on CoLES embeddings and a credit default task on a second public dataset.

Outside event sequences, tree ensembles remain strong on tabular data~\citep{grinsztajn2022tree,shwartzziv2022tabular}, and TabPFN~\citep{hollmann2025tabpfn} is a pretrained tabular model that is competitive on small datasets without training.
Our question lies between these lines of work: whether a model of the raw event sequence adds information to aggregates that a tree ensemble already uses well, and how the answer changes with the number of labels.

\subsection{Hybrids of boosting and neural networks}
\label{sec:rw-hybrids}

Combining gradient boosting with neural networks is an established practice.
\citet{he2014practical} fed the leaves of boosted trees to a linear model, Wide \& Deep~\citep{cheng2016wide} trained a linear and a deep model jointly, DeepGBM~\citep{ke2019deepgbm} distilled boosted trees into a neural network, and GrowNet~\citep{badirli2020grownet} used shallow networks as the weak learners of boosting.
In BGNN~\citep{ivanov2021bgnn}, boosted trees and a graph neural network are trained together, and stacking~\citep{wolpert1992stacked} uses out-of-fold predictions of one model as inputs of another.
For sequences, \citet{leontjeva2016combining} and \citet{chen2017hybrid} combined features learned by recurrent models with static features in a tree-based classifier, and CoLES feeds its embeddings to LightGBM together with hand-crafted aggregates~\citep{babaev2022coles}.
In credit, the winning solution of the American Express default prediction competition blended LightGBM models on aggregates with a recurrent network over the raw statements~\citep{jxzly2022amex}, without self-supervised pretraining.

Score-level combination is the simplest form of such hybrids, and rank-level combination of classifiers has a long history~\citep{ho1994decision}.
The closest precedent to TBS is the ensemble of EventGPT, which averages the ranks of the pretrained model, used through frozen embeddings, and of a boosting model on engineered features~\citep{burakov2026eventgpt}.
EventGPT differs from our study in several respects: its pretrained model is used mainly through frozen embeddings, with fine-tuning reported for a single seed; its pretraining period overlaps the label months of the test clients, whose own histories are excluded, an overlap it measures with a separate out-of-time run; and it is evaluated on product-propensity and churn targets of a single public dataset, without credit outcomes, transition features or a residual variant.
It also covers aspects that our study does not, such as zero-shot scoring from the next-event distribution, a larger model of about 23 million parameters, geolocation events and the public release of its weights.
Joint fusion, on the other hand, cannot include a boosting model, because there are no gradients to propagate from the trees to the Transformer~\citep{braithwaite2026nuformer}.

On the side of the inputs, window aggregates are the standard representation of transaction histories for tree models~\citep{whitrow2009aggregation,bahnsen2016feature}, and \citet{lucas2020hmm} added features derived from hidden Markov models of the sequence to capture its dynamics.
TBS combines these elements in a configuration we did not find in prior work: a boosting model on point-in-time aggregates, enriched with label-free counts of ordered event transitions, combined at the level of the score with a pretrained network fine-tuned jointly with the aggregates.

\subsection{Evaluation practice and public benchmarks}
\label{sec:rw-evaluation}

The variation caused by random seeds and data splits can exceed the differences that benchmarks report~\citep{bouthillier2021variance}, and leakage between training and test information is a common source of inflated results in applied machine learning~\citep{kapoor2023leakage}.
Both concerns apply to financial event sequences, where small differences in AUC are the norm and where the time at which an event becomes known is easy to get wrong.

Two synthetic datasets from IBM are often used to report results of transaction models.
TabFormer, released with TabBERT~\citep{padhi2021tabbert}, is used by vendors of foundation models to report fraud detection results~\citep{neospace2026cortex}.
The anti-money-laundering datasets of \citet{altman2023aml} were used to propose graph neural networks for directed multigraphs, Multi-GIN and Multi-PNA~\citep{egressy2024multignn}, which build on GIN~\citep{xu2019gin} and PNA~\citep{corso2020pna}.
Section~\ref{sec:benchmarks} examines what the published comparisons on these two benchmarks measure.

\subsection{Positioning}
\label{sec:rw-positioning}

Table~\ref{tab:related} compares the closest studies with ours along the dimensions that matter for the questions of Section~\ref{sec:introduction}.
No single study among them measures the value of pretrained sequence models against a strong boosting baseline on credit outcomes, at several label budgets, with paired uncertainty, and with controls that separate the contribution of pretraining from that of combining two models.

The table also shows how the evidence is distributed.
Most studies that include a boosting model use it as the downstream learner of frozen embeddings, and only EventGPT combines the two models at the level of their scores.
Uncertainty estimates are reported by five of the eight studies, and none of them reports that its main pretraining is restricted to a period before the evaluation, a condition that matters when the pretraining corpus and the test set come from the same institution.
Our study fills the gap at a small scale and on four datasets, two of them proprietary, and the conclusions should be read with these limits in mind (Section~\ref{sec:threats}).

\begin{table}[ht]
  \centering
  \caption{Comparison of the closest studies with ours, as described in each paper. Credit: a credit-risk target is evaluated. SSL: self-supervised pretraining. Fine-tuned: the pretrained model is fine-tuned on the target. Temporal pretraining: pretraining data are restricted to a period before the evaluation period. GBDT: comparison with gradient boosting on aggregates or engineered features. Combination with GBDT: pretrained embeddings fed to boosting (feature) or scores combined (score). Budgets: several numbers of labels. Uncertainty: standard deviations, intervals or tests on the reported results. n.r.: not reported; --: not applicable (no pretraining); partly: only in a secondary experiment.}
  \label{tab:related}
  \footnotesize
  \setlength{\tabcolsep}{3.5pt}
  \begin{tabular}{lccccccccc}
    \toprule
    Study & Credit & \shortstack{Public\\data} & SSL & \shortstack{Fine-\\tuned} & \shortstack{Temporal\\pretraining} & GBDT & \shortstack{Combination\\with GBDT} & Budgets & Uncertainty \\
    \midrule
    TabBERT (2021) & no & yes & yes & no & n.r. & no & no & no & no \\
    E.T.-RNN (2019) & yes & no & no & -- & -- & yes & no & yes & yes \\
    CoLES (2022) & yes & yes & yes & yes & n.r. & yes & feature & yes & yes \\
    EBES (2025) & no & yes & yes & yes & n.r. & no & no & no & yes \\
    MBD (2025) & yes & yes & yes & no & n.r. & yes & feature & no & yes \\
    nuFormer (2026) & yes & no & yes & yes & n.r. & yes & feature & yes & no \\
    PRAGMA (2026) & yes & no & yes & yes & n.r. & no & no & no & no \\
    EventGPT (2026) & no & yes & yes & partly & partly & yes & score & yes & yes \\
    This study & yes & partly & yes & yes & yes & yes & score & yes & yes \\
    \bottomrule
  \end{tabular}
\end{table}

\section{Evaluation Methodology}
\label{sec:methodology}

This section states the rules that every comparison of the study follows.
Their purpose is to make a small difference between two models detectable when it exists, and to keep the comparison honest when it does not.

The evaluation has two parts.
The main study compares 27 methods with a common baseline of point-in-time window aggregates and LightGBM on four datasets, three label budgets and five paired seeds, and answers RQ1 to RQ4.
All the rules of this section apply to it in full.
The re-examination of the public TabFormer and IBM AML benchmarks answers RQ5.
It follows the same principles of point-in-time inputs, out-of-time testing and thresholds chosen on validation data, but each benchmark keeps its own split and metrics so that our results can be compared with the published ones, and its experiments use fewer seeds; the differences are stated in Section~\ref{sec:benchmarks}.
The rest of this section describes the temporal constraints on inputs and labels, the label budgets and the pairing of runs, the metrics and the statistical analysis, and the rules we follow to interpret the results; the datasets, dates and configurations to which these rules are applied are described in Section~\ref{sec:setup}.

\subsection{Temporal and information constraints}
\label{sec:method-pit}

Each labeled example is an entity at an as-of date, and every input, whether an aggregate, a token or a pretraining sequence, is computed only from events that were known strictly before that date.
On the two proprietary datasets, each event carries the date on which it became known, not the date it refers to: a payment enters on the day it was made, a delinquency on the day it was confirmed and the monthly revenue after the month closed.
The distinction matters in practice: in one of the proprietary datasets, a field with the days of delay of an installment entered the sequence on the date of the installment rather than on the date the delay became known, and it inflated the results until it was dated by when it became known.
We verified on each dataset that no event at or after the as-of date reaches the inputs.

Every vocabulary of event codes, category encoding, quantile band and scaler is computed only on development data: training and unlabeled entities, with events before the first test date.
Levels that do not occur there fall in an ``other'' category or a missing code on the test set.
The one exception is TBS, whose ranks are taken among the scored test rows (Section~\ref{sec:models-tbs}).

A training label must also be known before the model is used.
For each dataset we fix the date on which the last training label becomes known, and we require that it is not later than the first as-of date of the test set.
Where the label of an example matures after the end of the training window, as for invoices that fall due after the test begins, the example is excluded from training by a rule that depends only on dates fixed in advance, so that the rule does not select examples by their outcome.

The test set contains later as-of dates and entities that were seen neither by supervised training nor by self-supervised pretraining, and each test entity appears once.
Pretraining uses only entities and events from the training period, labeled or not.
The test is therefore both out of time and out of sample, which reproduces the use of a model on new clients after it is trained.

\subsection{Label budgets and paired runs}
\label{sec:method-budgets}

Practitioners rarely know in advance how many labels a new problem will have, and the value of pretraining is expected to depend on it (H4).
We therefore train every supervised component with 1,000 labels, with 4,000 labels and with all available training labels, drawn by stratified sampling from the same training pool.
The test set does not change with the budget, so that differences between budgets come from the training data alone.

Every cell of the study, defined by a dataset, a scenario, a target, a budget and a method, is run with five seeds.
A seed fixes the label subsample at the two smaller budgets, the split used for early stopping, the initialization of the networks and the randomness of the boosting model.
The baseline runs with the same seeds, so that each method is compared with the baseline seed by seed, on the same training rows and the same test rows.

\subsection{Metrics and statistical analysis}
\label{sec:method-stats}

The primary metric of the main study is the AUC, which does not depend on a decision threshold and is the usual measure of ranking quality in credit.
In the re-examination of the fraud and anti-money-laundering benchmarks, we follow their practice and report the average precision (AP), the non-interpolated summary of the precision--recall curve, which those benchmarks call AUPRC, and the F1 score.
F1 depends on a decision threshold, which we always choose on validation data; a threshold that maximizes F1 on the test set is used only to reproduce published numbers, and is identified as such.

For a candidate $c$ and a reference $r$, seed $s$ gives the paired difference
\begin{equation}
  \Delta^{(s)} = \mathrm{AUC}^{(s)}_{c} - \mathrm{AUC}^{(s)}_{r},
  \label{eq:paired-diff}
\end{equation}
and we report its mean over the five seeds, $\bar{\Delta}$, in AUC points, where one point is 0.01 AUC.
Its uncertainty is estimated with a paired bootstrap of the test set~\citep{efron1993bootstrap}.
We draw $B = 2{,}000$ resamples of the test rows, use the same resamples for both models and for every seed, and compute in each resample $b$ the mean difference over seeds $\bar{\Delta}^{*}_{b}$.
Because each entity appears once in the test set, resampling rows resamples entities.
The 95\% interval is given by the 2.5th and 97.5th percentiles of $\bar{\Delta}^{*}_{b}$, and the two-sided p-value by
\begin{equation}
  p = \min\left\{1,\ 2\min\left[\frac{1 + \#\{\bar{\Delta}^{*}_{b} \le 0\}}{B + 1},\ \frac{1 + \#\{\bar{\Delta}^{*}_{b} \ge 0\}}{B + 1}\right]\right\}.
  \label{eq:pvalue}
\end{equation}

These intervals hold the trained models fixed.
They measure the uncertainty that comes from the finite test set, conditional on our five training runs, and not the variation that new training samples, initializations or splits would produce.
As a sensitivity analysis, a second interval also resamples the five seeds with replacement inside each test resample, which adds the variation among our runs; with five seeds, it is a coarse estimate.
We also report how many of the five seeds favor the candidate, which exposes differences that depend on a single seed~\citep{bouthillier2021variance}.

The study makes many comparisons, and we control the false discovery rate with the Benjamini--Hochberg procedure~\citep{benjamini1995fdr} in two families.
The first family contains the 1,005 comparisons with the baseline, covering all methods, both scenarios and the ensemble controls.
The second contains the 273 comparisons that support our conclusions about where the gains come from: pretraining, joint fusion against the frozen encoder, and TBS against its components and its controls.
Each family is adjusted within itself, and we give counts of cells before and after the adjustment.
The adjustment and the seed resampling are separate analyses, so we also count the cells that pass both.

\subsection{Interpretation rules}
\label{sec:method-interpretation}

An interval that covers zero means that the test could not separate the difference from chance, and we do not read it as evidence that two models are equivalent.
How small a difference the test can resolve depends on the number of test positives; on the proprietary churn dataset, with 600 positives, differences of about 0.01 AUC are below what the test can detect.

Cells are not independent.
The three budgets of a target share its test set, and the targets of a dataset share entities and pretraining, so counts of cells summarize the evidence rather than add independent confirmations.

No method, the baseline included, received a hyperparameter search.
All methods use a fixed configuration, given in Section~\ref{sec:setup}, with early stopping on a validation split carved from the training rows.
This reflects how a model is first tried on a new dataset, but a fixed configuration is not neutral, because it can suit one family better than another.
For example, the networks train for a small number of epochs with early stopping, and a pretrained encoder may reach a good solution within that budget more easily than a randomly initialized one, so part of the measured value of pretraining may be ease of optimization.
Our results are therefore comparisons at fixed configurations, and a tuned comparison could move individual numbers.

\section{Models and Comparison Design}
\label{sec:models}

We compare 27 methods, grouped by how the event history reaches the final model, and four ensemble controls built from their saved predictions (Table~\ref{tab:methods}).
The neural models are our own small-scale versions of published recipes, not reproductions of the original systems.
They share one tokenizer and one backbone, so that differences between them come from the training recipe rather than from capacity, and TabPFN v2 is the only published model used as released.
Each method has a role in the design: the baseline is the reference of every comparison, the alternatives are the foundation-model recipes and supervised models whose added value we measure, TBS and its variant are the method we propose, and the ablations and controls isolate the contribution of each component.

This section describes the methods in that order.
The concepts behind them are introduced in Section~\ref{sec:background}, and the full configurations are given in Appendix~\ref{app:config}.

\begin{table}[ht]
  \centering
  \caption{Methods compared, grouped by how the event history reaches the final model, with their role in the design. Ensemble controls are rank averages of two saved predictions, built like TBS.}
  \label{tab:methods}
  \footnotesize
  \begin{tabularx}{\linewidth}{>{\raggedright\arraybackslash}p{3.3cm}X>{\raggedright\arraybackslash}p{2.2cm}}
    \toprule
    Group & Methods & Role \\
    \midrule
    Baseline & Aggregates + GBDT (Agg) & reference \\
    Frozen representation + GBDT & AE, Transformer, masked numeric; AE, Transformer, + one-hot; AE, MLP, whole sequence; AE, MLP, masked events; MLM (TabBERT-style); Nep (nuFormer-style); CoLES; Mlm-Cl & alternatives (RQ1) \\
    End-to-end & GRU; JF-rand (no pretraining); JF (pretrained Nep + joint fusion); TabPFN v2 & alternatives and ablation (RQ2) \\
    Expert features & Expert features alone; Agg + expert features & reference \\
    Transition features & Agg + Trans; Agg + Trans + Mlm-Cl & components of TBS \\
    TBS and hybrids & TBS; TBS-R (boosting residual); TBS-R, no pretraining; TBS-R, no Trans in offset; TBS-R, Nep encoder; JF + Trans & proposed method, variants and ablations (RQ3) \\
    Field encoder & Nep, field encoder; JF, field encoder; JF + Trans, field encoder; TBS, field encoder & variants \\
    Ensemble controls & Agg with Agg + Trans; Agg with JF; Agg + Trans with JF-rand; Agg + Trans with GRU & controls (RQ3) \\
    \bottomrule
  \end{tabularx}
\end{table}

\subsection{Baseline and reference features}
\label{sec:models-baseline}

The baseline, Agg, summarizes each history with point-in-time aggregates over windows of 15, 30, 60, 90 and 180 days before the as-of date.
For each numeric field it computes the sum, mean, count, standard deviation, minimum, maximum and last value, and for each categorical field it counts the events of each level.
This is the aggregation a practitioner would configure without domain knowledge, and it is the starting point of the production models of the two companies whose data we use.
The model is LightGBM~\citep{ke2017lightgbm} with a fixed configuration and early stopping after 100 rounds without improvement on a stratified validation split of 20\% of the training rows.
Every method that ends in a boosting model uses the same configuration and the same split, so that differences between them come from their inputs.

Two datasets also have features written with domain knowledge, which serve as a reference for what such knowledge buys.
On the synthetic data, 96 features were written by someone who knows the generated patterns, for example the number of bets followed by a cash withdrawal within two hours.
On client B, they are the twelve variables of the company's production model.
They are reported alone and added to the aggregates; the other two datasets have no such features.

\subsection{Shared sequence representation}
\label{sec:models-backbone}

All sequence models receive the same tokens, built as described in Section~\ref{sec:bg-tokens}.
Each entity is represented by its last $L = 128$ events before the as-of date, and each event by a set of categorical fields and three continuous values.
The categorical fields are the event codes of the dataset, such as merchant group, channel, event type and subtype, currency or counterparty type, with vocabularies capped at the most frequent levels of the training period, together with the amount, the gap since the previous event and the recency in quantile bands, and the hour of day.
The continuous values are the standardized logarithms of the amount, of the time since the previous event and of the time to the as-of date.
Bands and standardization use training-period data only.
Events without an amount, such as the support tickets and survey answers of client B, receive an amount of zero in the tokens, while the aggregates treat the amount as missing.

The backbone is a pre-norm Transformer encoder~\citep{vaswani2017attention} with model width 96, four layers, four heads, feed-forward width 256, dropout 0.1, learned positional embeddings and a final layer normalization, for about 0.4 million parameters.
In the default event encoder, the embeddings of the fields of an event and a linear projection of its continuous values are summed, as in Equation~\eqref{eq:event-embedding}.
A second version replaces the sum with a two-layer Transformer over the fields of each event, whose outputs are concatenated and projected, the hierarchical field-then-event design of TabBERT~\citep{padhi2021tabbert}; we call it the field encoder.
The representation of a sequence is the concatenation of the output at the last position and the mean over the valid positions, a vector of 192 dimensions.

\subsection{Frozen representations}
\label{sec:models-frozen}

Each method of this group produces a fixed embedding per entity, which is appended to the aggregates as extra columns of the boosting model.
The first four are generic autoencoders from our production platform and stand for the common practice of compressing histories with a reconstruction objective.
Two of them are small Transformers, of width 64 and two layers over the last 50 events, that reconstruct masked numeric values, the second also with one-hot categories; the other two are MLP autoencoders, one reconstructing the flattened sequence and one reconstructing masked events.

The other four use the shared backbone or its embeddings and follow the objectives of the literature.
MLM uses bidirectional attention and is pretrained to recover masked events (15\%) and masked individual fields (10\%), with a cross-entropy loss per field and a squared-error loss for the continuous values, in the line of TabBERT and PRAGMA~\citep{padhi2021tabbert,ostroukhov2026pragma}.
Nep uses causal attention and is pretrained to predict every field of the next event, in the line of nuFormer~\citep{braithwaite2026nuformer}; the published model serializes events into text tokens, and summing field embeddings is our adaptation.
CoLES trains a GRU over the same event embeddings so that random contiguous sub-sequences of the same entity are close and those of different entities are far~\citep{babaev2022coles}.
Mlm-Cl trains the bidirectional backbone with the masked loss plus a contrastive term between two views of the same entity: one view drops 0 to 30\% of the most recent events, which imitates an earlier as-of date, and the other is a contiguous crop of 50 to 100\% of the history.
All pretraining uses AdamW with a one-cycle schedule, batches of 128 and 8 to 12 epochs, on training-period entities only.

\subsection{Supervised and fine-tuned models}
\label{sec:models-supervised}

The supervised GRU reads the event embeddings without pretraining; it is the strongest simple baseline of the EBES benchmark~\citep{osin2025ebes} and represents a sequence model trained only on the labels.
Joint fusion (JF) trains the pretrained Nep backbone together with an MLP over the quantile-normalized aggregates on the label, the recipe that nuFormer calls joint fusion~\citep{braithwaite2026nuformer}.
The encoder uses a smaller learning rate than the rest of the network, and training stops early on the validation AUC.
JF-rand is the same network without pretraining, and isolates the contribution of pretraining to the fine-tuned model.

TabPFN v2~\citep{hollmann2025tabpfn} is applied to the aggregates and provides a pretrained tabular alternative to boosting that does not use the order of events.
It runs at the two budgets within its limit of 10,000 training rows, and when there are more than 300 aggregates we keep the 300 with the largest LightGBM gain on the same training rows.

\subsection{The proposed method: TBS}
\label{sec:models-tbs}

TBS combines two models that use the event history in different ways.
The first is a LightGBM on the aggregates of the baseline plus a set of event-transition features, Trans, which recover part of the information on the order and timing of events that window aggregates lose (Section~\ref{sec:bg-agg}).
Trans mines ordered pairs of event types without looking at the label.
An event type is the merchant group crossed with the decline flag on the synthetic data, and one of the 30 most frequent event codes elsewhere, with the source type on the proprietary data and the remaining codes pooled into one level.
A pair $A \to B$ is counted when an event of type $B$ immediately follows an event of type $A$ of the same entity within a gap $\Delta t$.
We keep the pairs whose support, the share of training-period entities that have the pair within 24 hours in the last 180 days, is at least 1\%, up to the 150 most supported pairs, and count their occurrences for $\Delta t \in \{1\,\mathrm{h}, 24\,\mathrm{h}\}$ in windows of 30, 90 and 180 days before the as-of date.
In the same windows, we add rhythm features: the number of events and of distinct types, the median and the coefficient of variation of the gaps, the number of events within ten minutes of the previous one, the number of repeats of the same type within one hour, and the maximum number of events in one hour and in one day.
A final count gives the number of event types that appear in the last 30 days and never before.

The second model is the fine-tuned joint-fusion network JF of Section~\ref{sec:models-supervised}.
The score of TBS is the average of the ranks of the two scores on the scored population.
Ranks are used instead of probabilities because the two scores are not on the same scale.
For example, if the scores of the network range from 0.09 to 0.71 and those of the boosting model from 0.011 to 0.091, an average of the probabilities is decided almost entirely by the network, while an average of the ranks gives both models the same weight: an entity ranked first by the network and third by the boosting model gets an average rank of two, which is not necessarily the second position in the final order.
It has no combination weights to fit, although the choice of architecture and components can still adapt to the validation data.
It does not produce a calibrated probability, and a calibration step is needed where probabilities are required.
Ranks are taken among the scored test rows, so the score is defined for a batch; used one record at a time it would need a reference distribution from earlier data, which we did not evaluate.
TBS also preserves the explainability of its tabular component.
Its branch is explained with SHAP values~\citep{lundberg2017shap}, including the transition features, which are named columns.
The shift that the network causes in the rank of each entity shows the effect of the combination, but it does not explain which events led the network to its score, which requires additional analysis.

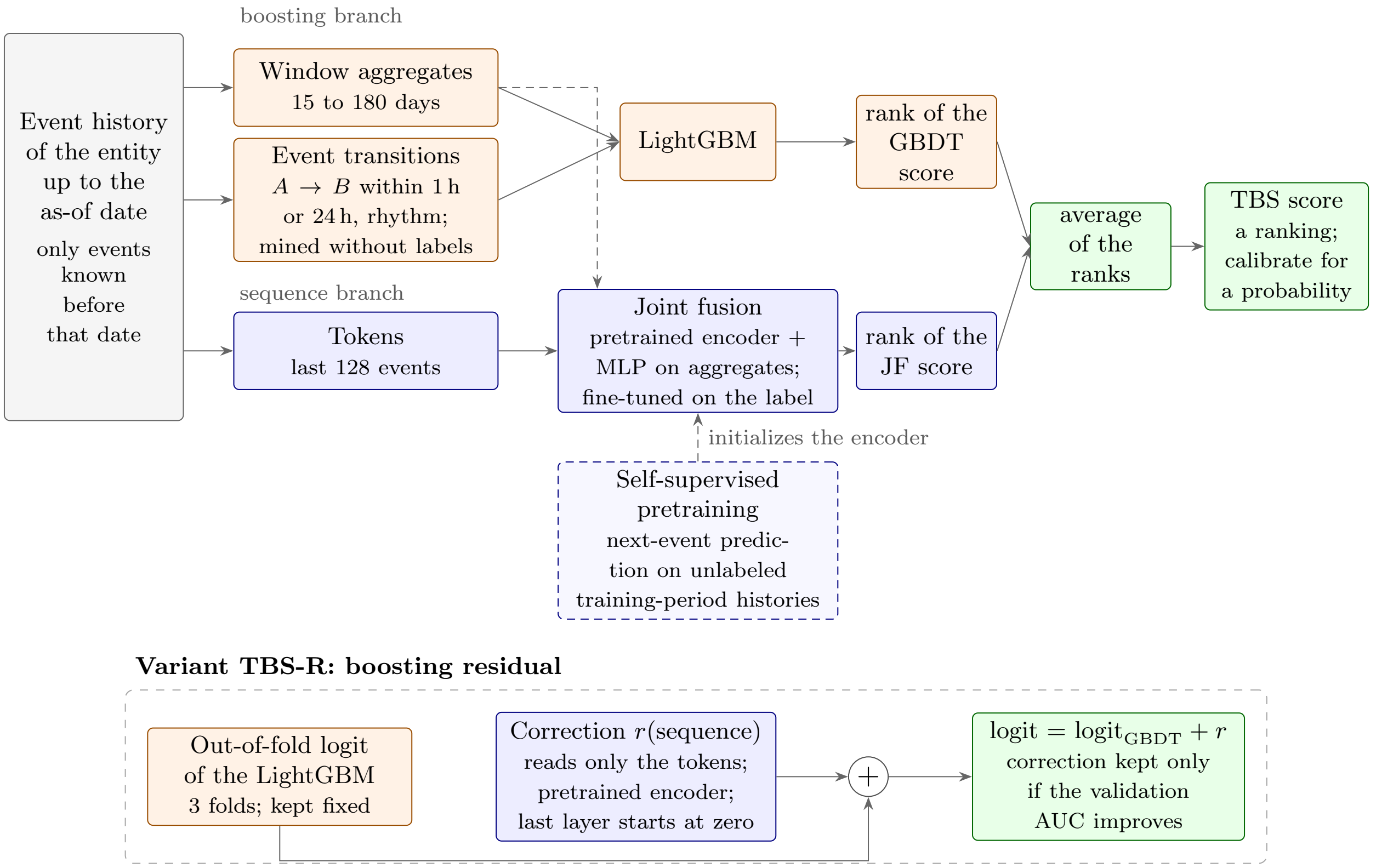
\begin{figure}[t]
  \centering
  \resizebox{\linewidth}{!}{\begin{tikzpicture}[
  font=\footnotesize, >=Stealth,
  box/.style={draw=black!60, rounded corners=2pt, align=center, minimum height=1.0cm, inner sep=3pt},
  gb/.style={box, fill=orange!10, draw=orange!60!black},
  nn/.style={box, fill=blue!7, draw=blue!50!black},
  fin/.style={box, fill=green!9, draw=green!40!black},
  arr/.style={->, draw=black!60},
  lab/.style={font=\scriptsize, black!65},
]
  \node[box, fill=black!4, text width=2.1cm, minimum height=5.0cm] (hist) at (0,0.2) {Event history\\of the entity\\up to the\\as-of date\\[4pt]{\scriptsize only events known\\before that date}};

  \node[lab, anchor=south west] at (1.75,2.65) {boosting branch};
  \node[gb, text width=3.2cm, anchor=west] (agg) at (1.8,2.0) {Window aggregates\\{\scriptsize 15 to 180 days}};
  \node[gb, text width=3.2cm, anchor=west] (trans) at (1.8,0.55) {Event transitions\\{\scriptsize $A \to B$ within 1\,h or 24\,h, rhythm;}\\{\scriptsize mined without labels}};
  \node[gb, text width=1.8cm] (lgb) at (7.8,1.3) {LightGBM};
  \draw[arr] (hist.east |- agg) -- (agg.west);
  \draw[arr] (hist.east |- trans) -- (trans.west);
  \draw[arr] (agg.east) -- (lgb.west);
  \draw[arr] (trans.east) -- (lgb.west);

  \node[nn, text width=3.2cm, anchor=west] (tok) at (1.8,-1.4) {Tokens\\{\scriptsize last 128 events}};
  \node[nn, text width=3.4cm] (jf) at (7.8,-1.4) {Joint fusion\\{\scriptsize pretrained encoder $+$ MLP on aggregates;}\\{\scriptsize fine-tuned on the label}};
  \node[lab, anchor=south west] at (1.75,-0.93) {sequence branch};
  \draw[arr] (hist.east |- tok) -- (tok.west);
  \draw[arr] (tok.east) -- (jf.west);
  \draw[arr, densely dashed] (agg.east) -| ($(jf.north)+(-1.3,0)$);

  \node[nn, fill=blue!3, densely dashed, text width=3.4cm] (pre) at (7.8,-3.85) {Self-supervised pretraining\\{\scriptsize next-event prediction on unlabeled training-period histories}};
  \draw[arr, densely dashed] (pre.north) -- node[lab, right] {initializes the encoder} (jf.south);

  \node[gb, text width=1.6cm] (rg) at (10.75,1.3) {rank of the\\GBDT score};
  \node[nn, text width=1.6cm] (rn) at (10.75,-1.4) {rank of the\\JF score};
  \draw[arr] (lgb.east) -- (rg.west);
  \draw[arr] (jf.east) -- (rn.west);
  \node[fin, text width=1.6cm] (avg) at (13.0,-0.05) {average\\of the ranks};
  \draw[arr] (rg.east) -- (avg.west);
  \draw[arr] (rn.east) -- (avg.west);
  \node[fin, text width=1.9cm] (tbs) at (15.4,-0.05) {TBS score\\{\scriptsize a ranking; calibrate for a probability}};
  \draw[arr] (avg.east) -- (tbs.west);

  \begin{scope}[yshift=-6.9cm]
    \node[gb, text width=3.2cm] (off) at (2.4,0) {Out-of-fold logit of the LightGBM\\{\scriptsize 3 folds; kept fixed}};
    \node[nn, text width=3.4cm] (res) at (7.0,0) {Correction $r(\text{sequence})$\\{\scriptsize reads only the tokens; pretrained encoder; last layer starts at zero}};
    \node[circle, draw=black!70, inner sep=1.2pt, font=\normalsize] (plus) at (10.0,0) {$+$};
    \node[fin, text width=3.3cm] (lg) at (13.1,0) {$\mathrm{logit} = \mathrm{logit}_{\mathrm{GBDT}} + r$\\{\scriptsize correction kept only if the validation AUC improves}};
    \draw[arr] (res.east) -- (plus.west);
    \draw[arr] (off.south) -- ++(0,-0.45) -| (plus.south);
    \draw[arr] (plus.east) -- (lg.west);
    \node[draw=black!35, dashed, rounded corners=3pt, fit=(off)(res)(plus)(lg), inner sep=8pt] (inset) {};
    \node[font=\footnotesize\bfseries, anchor=south west] at (inset.north west) {Variant TBS-R: boosting residual};
  \end{scope}
\end{tikzpicture}}
  \caption{TBS. Top: the boosting branch reads window aggregates and label-free event transitions; the sequence branch reads the tokens of the last 128 events with an encoder pretrained without labels and fine-tuned on the label together with an MLP on the aggregates (joint fusion). Each branch ranks the entities, and TBS averages the two ranks. Bottom: the variant TBS-R keeps the out-of-fold boosting logit fixed and adds a correction computed from the sequence, which is kept only if it improves the validation AUC.}
  \label{fig:tbs}
\end{figure}

Figure~\ref{fig:tbs} shows the two branches of TBS and their combination.

The variant TBS-R combines the same two sources through a boosting residual.
The out-of-fold logit of the LightGBM on the aggregates and Trans, computed with three folds, enters as a fixed offset, and a sequence network adds a correction, so that $\mathrm{logit} = \mathrm{logit}_{\mathrm{GBDT}} + r(\text{sequence})$.
The last layer of $r$ starts at zero, so the model starts exactly at the boosting score, and training keeps a correction only if it improves the AUC on a validation split of 20\% of the training entities; otherwise the output remains the boosting score.
The encoder of $r$ is Mlm-Cl by default.
In TBS-R, the contribution of the network is the correction it adds to the logit of each entity.
A third hybrid, JF + Trans, adds the transition features to the tabular branch of joint fusion and tests whether a single network can replace the combination of two models.

\subsection{Ablations and controls}
\label{sec:models-ablations}

The hypotheses of Section~\ref{sec:introduction} are tested through targeted comparisons, each of which changes as few components as possible (Table~\ref{tab:comparisons}).
The contribution of pretraining to the fine-tuned network (H2) is measured by JF against JF-rand, which differ only in the initialization of the encoder.
The comparison of JF with the frozen Nep encoder fed to boosting also bears on H2, but it changes more than fine-tuning: the frozen embedding enters a boosting model, while joint fusion feeds the aggregates to an MLP trained together with the encoder, so it does not isolate the effect of fine-tuning.
The same joint-fusion network with a frozen encoder would isolate it, and we did not run it.

For TBS (H3), the comparisons with its two components, Agg + Trans and JF, measure whether the combination improves on each of them.
Four ensemble controls, built from the saved predictions exactly as TBS, separate the sources of the gain.
The ensemble of Agg with Agg + Trans combines two boosting models without any sequence network and measures what combining alone is worth.
The ensemble of Agg with JF removes the transition features, the ensemble of Agg + Trans with JF-rand removes pretraining, and the ensemble of Agg + Trans with the GRU replaces the pretrained network with a supervised sequence model.
For TBS-R, the ablations replace the Mlm-Cl encoder with an untrained one or with the Nep encoder, or drop Trans from the offset.
Finally, the field-encoder versions of Nep, JF, JF + Trans and TBS measure how much a richer event encoder changes each recipe.

\begin{table}[ht]
  \centering
  \caption{Targeted comparisons and the component each isolates.}
  \label{tab:comparisons}
  \footnotesize
  \begin{tabularx}{\linewidth}{>{\raggedright\arraybackslash}p{5.2cm}X>{\raggedright\arraybackslash}p{1.6cm}}
    \toprule
    Comparison & What it isolates & Hypothesis \\
    \midrule
    JF vs.\ JF-rand & pretraining of the fine-tuned network & H2 \\
    JF vs.\ frozen Nep + GBDT & fine-tuning, together with a change of predictor & H2 \\
    TBS vs.\ Agg + Trans & adding the pretrained network to the boosting model & H3 \\
    TBS vs.\ JF & adding the boosting model to the network & H3 \\
    TBS vs.\ ensemble of Agg + Trans and JF-rand & pretraining, within the combination & H3 \\
    TBS vs.\ ensemble of Agg and JF & transition features, within the combination & H3 \\
    Ensemble of Agg and Agg + Trans vs.\ Agg & combining two boosting models, without a network & H3 \\
    TBS-R vs.\ TBS-R, no pretraining & pretraining of the residual encoder & H3 \\
    Field-encoder version vs.\ sum & the event encoder & -- \\
    \bottomrule
  \end{tabularx}
\end{table}

\section{Experimental Setup}
\label{sec:setup}

The main study uses four datasets with different sources of signal (Table~\ref{tab:datasets}).
Two of them are public or reproducible, a synthetic credit-card dataset that we generated and the public MBD-mini banking dataset, and two are proprietary, from a consumer-goods company that extends trade credit to its points of sale and from a business-software company.
Together they cover credit, churn and product propensity, at the level of clients, invoices and contracts.

This section describes each dataset and its prediction tasks, the temporal splits and the rules that make labels mature before they are used, the two information scenarios in which every method is run, and the configuration and reproducibility of the experiments.
All four datasets follow the methodology of Section~\ref{sec:methodology}.

\begin{table}[ht]
  \centering
  \caption{Datasets. Entities are clients (synthetic, MBD-mini), invoices of points of sale (client A) or contracts (client B). Labeled entities are split by time into training and test; the remaining entities are used only for pretraining. Test positives are given per target.}
  \label{tab:datasets}
  \footnotesize
  \begin{tabular}{lrrrrll}
    \toprule
    Dataset & Entities & Events & Train labels & Test & Test positives & Targets \\
    \midrule
    Synthetic & 48,000 & 6.8 million & 17,879 & 6,121 & 610, 601, 618 & Default, Seq.\ trigger, Churn \\
    MBD-mini & 54,000 & 10.7 million & 22,000 & 8,000 & 416, 153 & Any product, Product 1 \\
    Client A & 56,000 & 1.4 million & 24,000 & 8,000 & 563 & Late payment \\
    Client B & 35,292 & 18.8 million & 15,376 & 10,000 & 600 & Churn (6m) \\
    \bottomrule
  \end{tabular}
\end{table}

\subsection{Datasets and prediction tasks}
\label{sec:setup-datasets}

The synthetic dataset contains card transactions, a customer register and monthly statements for 48,000 clients over twelve monthly reference dates.
Half of the clients are labeled, and the other half are used only for pretraining.
Each transaction has a merchant group, an amount, a number of installments, a channel, flags for cash withdrawals and declines, and a time of day.
The three targets were designed to represent different sources of signal.
Default depends on a latent risk visible in the register and the statements, on recent stress and on utilization.
The sequence trigger is driven by an order pattern, a bet followed by a cash withdrawal within two hours and then by another bet, and decoy clients have the same bets and withdrawals far apart in time, so that window counts alone cannot separate them.
Churn follows a gradual decline in activity.

Each synthetic label is drawn from a logistic function of a weighted sum of latent variables plus Gaussian noise, so the probability of the label given the latent variables is monotone in that sum, and its test AUC is the best achievable by a model that knows them.
We call it the latent-information oracle; it is 0.777 for default, 0.779 for the sequence trigger and 0.685 for churn.
We compute it directly from the weighted sum in the generator, not from a model fitted to the latent variables, and its values are empirical AUCs on our test clients, so a model can exceed them by chance on a finite test set.
The oracle is a reference, not a ceiling for the models we compare: some latent variables, such as the credit risk and the credit limit, are visible in the register but not in the transactions, so the best AUC achievable from transactions alone is lower.

The Multimodal Banking Dataset~\citep{mollaev2025mbd} is a public release of anonymized banking event sequences under the CC BY 4.0 license, and we use its reduced MBD-mini version.
We sample 54,000 clients with 10.7 million transactions.
The targets are whether the client buys any of four products in the month after its history ends and whether the client buys product 1, the most frequent.

Client A is a consumer-goods company that extends trade credit to its points of sale.
The unit is an invoice at its issue date, and the target is a payment more than seven days late.
The history of each point of sale contains invoice issuance, payment by delay band, maturity and confirmed delinquency events, each dated when it became known.
Product codes are encoded as the 40 most frequent codes in development data plus one category for the rest.

Client B is a business-software company that scores each contract, defined as a client and one of its products, on the first day of a month $t$ from the history of earlier months, and predicts whether a cancellation is recorded for it in month $t+6$.
The target is a positive net cancellation amount in the accounting records of that month, which can be a partial cancellation; cancellations in months $t+1$ to $t+5$ do not count, so the target is not a cancellation at any time within six months.
Population and target are those of the company's production model.
The history contains the monthly recurring revenue, known after the month closes, support tickets, dated by their creation, answers to satisfaction surveys and past cancellations.
The proprietary datasets are not released, and we report only aggregate results without naming the companies.

\subsection{Temporal splits, sampling and label maturity}
\label{sec:setup-splits}

Table~\ref{tab:asof} gives, for each dataset, the as-of dates of training and test, the label and the date on which the last training label becomes known.
On the synthetic data, reference dates from the last three months form the out-of-time test.
On MBD-mini, as in the protocol of the dataset~\citep{mollaev2025mbd}, the as-of date is the first day of the month whose purchases form the target and the input history contains only events before it, so training targets fall in April to October 2022 and test targets in November 2022 to January 2023.
In every dataset, test entities are disjoint from training entities.

\begin{table}[ht]
  \centering
  \caption{As-of dates and label maturity. The as-of date is the first day after the input history ends. On client B, $t$ is the month that starts at the as-of date. The last training label known should not be later than the first test as-of date. Synthetic labels are drawn from latent variables when the data are generated, and no events after the as-of date are simulated, so maturity does not apply.}
  \label{tab:asof}
  \footnotesize
  \begin{tabularx}{\linewidth}{llXll}
    \toprule
    Dataset & Training as-of dates & Label & Last training label known & Test as-of dates \\
    \midrule
    Synthetic & 2025-01-01 to 2025-09-01 & drawn at generation & n/a & 2025-10-01 to 2025-12-01 \\
    MBD-mini & 2022-04-01 to 2022-10-01 & purchase in the month that starts at the as-of date & 2022-11-01 & 2022-11-01 to 2023-01-01 \\
    Client A & 2019-03-01 to 2019-05-31 & known at payment, or at due date $+$ 8 days & 2019-06-15 & 2019-06-15 to 2019-07-08 \\
    Client B & 2023-06-01 to 2025-03-01 & net cancellation recorded in month $t+6$ & 2025-10-01 & 2025-10-01 to 2025-12-01 \\
    \bottomrule
  \end{tabularx}
\end{table}

On MBD-mini and client B, the date of the last training label is the first test as-of date.
On client A, invoices issued at the end of the training window fall due after the test begins, so training uses only invoices whose due date plus eight days is not later than the first test as-of date, 2019-06-15.
The due date is fixed when the invoice is issued, so this rule does not select invoices by how fast they were paid: without it, 657,450 invoices from 59,134 training points of sale would be eligible, with a late-payment rate of 6.4\%, and with it, 617,226 invoices from 58,968 points of sale, with the same rate.
The training sample is drawn from the eligible invoices, one invoice at random per point of sale for 24,000 points of sale, so its size does not depend on the rule, and neither the test set nor the unlabeled pretraining pool depends on it.

On MBD-mini and client B, each entity enters the labeled sample with a single month.
Entities with a positive event in some month enter with one such month as a positive, and the others with a random month as a negative, with positives capped at about 6\% of the sample.
This sampling changes the population, because negatives come only from entities that never had the event, and makes the task easier than month-by-month scoring of the whole portfolio.
Absolute AUCs on these two datasets should therefore not be compared with population-level evaluations; the comparison between methods holds for this sample, and we did not verify that it carries over to population scoring.

\subsection{Information scenarios}
\label{sec:setup-scenarios}

Every method is run in a transactions-only scenario, in which it sees only the event history of the entity.
This scenario isolates the question of what a model of the raw sequence adds to aggregates of the same events, and it is available for the seven targets of the four datasets.

Three datasets also have a full scenario, in which tabular information beyond the events is added to the inputs of every method: the register and statement aggregates on the synthetic data, the static attributes of the point of sale on client A, and the twelve variables of the production model on client B.
This scenario measures what the sequence adds when good tabular data already exist, which is the usual situation of a mature credit model (H4).
It covers five targets, so 15 target--budget cells can be compared between the two scenarios.

\subsection{Configuration and reproducibility}
\label{sec:setup-config}

No method received a hyperparameter search, and all configurations were fixed before the runs reported here (Appendix~\ref{app:config}).
Boosting models use early stopping after 100 rounds on a stratified 20\% split of the training rows, and the joint-fusion networks train for at most 12 epochs with a patience of two epochs on the validation AUC of the same kind of split.
All results of the main study come from one execution of a single pipeline per dataset, on one GPU per dataset (NVIDIA A10G; L4 for client A), with Python 3.13.18, PyTorch 2.12.1 (CUDA 13.0), LightGBM 4.7.0, scikit-learn 1.9.1, pandas 2.3.3, NumPy 2.3.5 and tabpfn 9.2.0.
The tables and every number in the text are generated by scripts from the per-seed predictions, and none was typed by hand.

The synthetic data generator, the computation of its latent-information oracle and the code of the paired statistics are released under the Apache 2.0 license.\footnote{\url{https://github.com/datarisk-io/fm-vs-gbdt-eval}}
The release also contains the per-seed test predictions and labels of the synthetic and MBD-mini datasets, from which the released code, with the resampling seeds of the paper, reproduces the intervals and p-values reported for those two datasets.
The generator is deterministic given its seeds and reproduces our synthetic files byte for byte.
The predictions, labels and comparison statistics of the proprietary datasets, and the rest of the pipeline, are not released.

\section{Results}
\label{sec:results}

This section answers the research questions in order.
Sections~\ref{sec:res-frozen} to~\ref{sec:res-oracle} report the main study on our four datasets and answer RQ1 to RQ4, Section~\ref{sec:benchmarks} re-examines the two public benchmarks and answers RQ5, and Section~\ref{sec:cost} reports the compute cost of the models.
Differences are given in AUC points, where one point is 0.01 AUC, as the mean over seeds of the paired difference with the baseline (Equation~\ref{eq:paired-diff}).

Table~\ref{tab:res-all} gives the difference in test AUC against the baseline for every method and target with all labels in the transactions-only scenario, and Table~\ref{tab:res-summary} summarizes all cells over the seven targets and three budgets.
Table~\ref{tab:res-summary} counts the cells whose interval against the baseline lies above zero in three ways: nominally, after the Benjamini--Hochberg adjustment, and with the seeds resampled as well as the test rows.
Counts in the text are nominal unless stated otherwise, and a cell passes both checks when it passes the adjustment and the seed resampling (Section~\ref{sec:method-stats}).
TBS and its field-encoder version are the only methods of the study whose interval lies above zero in nearly every cell: TBS in 19 of 21, with 17 passing both checks and a median gain of +2.1 points.
Excluding the synthetic dataset, TBS is above zero in 11 of 12 cells, with a median gain of +1.7 points.
In the most mature case we tested, with all labels and the rich tabular data of the full scenario on real data, the gain is +0.7 points on client A and +0.8 on client B, the latter with an interval that includes zero (Table~\ref{tab:res-full}).
TBS-R with the Nep encoder follows (15 of 21, median +1.2), but only 9 of its cells hold with the seeds resampled, so its gains depend more on the particular runs.
TabPFN is above the baseline in 10 of its 14 cells at the two budgets where it runs (8 with the seeds resampled).
The transition features alone clear the baseline in 7 of 21 cells, the fine-tuned network alone in 10, and frozen representations in at most 9.
Only the reconstruction autoencoders, the GRU and joint fusion without pretraining fall below the baseline, with intervals below zero, in more than one cell.
Appendix~\ref{app:results} gives the corresponding tables for 1,000 and 4,000 labels, the full scenario and the absolute AUCs.

\begin{table}[p]
  \centering
  \caption{Difference in test AUC against the baseline (Agg), in AUC points ($100 \times \Delta\mathrm{AUC}$), transactions-only scenario, all training labels. Mean over seeds of the paired difference. $\uparrow$/$\downarrow$: 95\% bootstrap interval above/below zero (bold when above); $*$: Benjamini--Hochberg $q < 0.05$ over all comparisons with the baseline. The first row gives the baseline AUC.}
  \label{tab:res-all}
  \footnotesize
  \setlength{\tabcolsep}{3.5pt}
  \begin{tabular}{lrrrrrrr}
    \toprule
    & \multicolumn{3}{c}{Synthetic} & \multicolumn{2}{c}{MBD-mini} & \multicolumn{1}{c}{Client A} & \multicolumn{1}{c}{Client B} \\
    \cmidrule(lr){2-4} \cmidrule(lr){5-6} \cmidrule(lr){7-7} \cmidrule(lr){8-8}
    Method & Default & \shortstack{Seq.\\trigger} & Churn & \shortstack{Any\\product} & \shortstack{Product\\1} & \shortstack{Late\\payment} & \shortstack{Churn\\(6m)} \\
    \midrule
    Agg AUC & 0.654 & 0.733 & 0.653 & 0.741 & 0.719 & 0.846 & 0.609 \\
    \multicolumn{8}{l}{\textit{Frozen representation + GBDT}} \\
    AE, Transformer, masked numeric & +0.5 & +0.0 & $-$0.4 & $-$0.1 & +0.5 & $-$0.2 & +0.1 \\
    AE, Transformer, + one-hot & +0.3 & +0.9 & $-$0.4 & $-$0.1 & +0.5 & +0.0 & +0.5 \\
    AE, MLP, whole sequence & $-$0.2 & $-$0.2 & $-$0.6 & $-$0.2 & +1.4 & $-$0.2 & +0.2 \\
    AE, MLP, masked events & +0.3 & $-$0.7$^{\downarrow*}$ & $-$0.2 & $-$0.1 & +0.7 & $-$0.1 & +0.2 \\
    MLM (TabBERT-style) & +0.6 & \textbf{+1.5$^{\uparrow*}$} & $-$0.1 & +0.4 & \textbf{+1.7$^{\uparrow}$} & $-$0.2 & +0.7 \\
    Nep (nuFormer-style) & +0.5 & \textbf{+1.1$^{\uparrow*}$} & +0.4 & \textbf{+0.8$^{\uparrow}$} & +1.6 & +0.3 & +1.3 \\
    CoLES & +0.7 & \textbf{+3.4$^{\uparrow*}$} & $-$0.7 & \textbf{+1.0$^{\uparrow*}$} & \textbf{+1.9$^{\uparrow*}$} & +0.5 & \textbf{+1.9$^{\uparrow*}$} \\
    Mlm-Cl & \textbf{+1.1$^{\uparrow*}$} & \textbf{+1.4$^{\uparrow*}$} & $-$0.2 & \textbf{+1.1$^{\uparrow*}$} & +0.7 & +0.2 & \textbf{+1.4$^{\uparrow}$} \\
    \multicolumn{8}{l}{\textit{End-to-end}} \\
    GRU & +0.7 & \textbf{+2.3$^{\uparrow*}$} & +0.1 & +0.2 & +1.6 & +0.8 & \textbf{+2.2$^{\uparrow*}$} \\
    JF-rand (no pretraining) & $-$0.1 & \textbf{+1.8$^{\uparrow*}$} & $-$0.6 & \textbf{+1.1$^{\uparrow*}$} & +1.7 & +0.0 & +0.5 \\
    JF (pretrained Nep + joint fusion) & +0.8 & \textbf{+2.7$^{\uparrow*}$} & $-$0.2 & +0.8 & +2.2 & +0.5 & +0.8 \\
    \multicolumn{8}{l}{\textit{Expert features}} \\
    Expert features alone & +0.4 & \textbf{+3.6$^{\uparrow*}$} & +0.0 & -- & -- & -- & $-$1.7 \\
    Agg + expert features & +0.3 & \textbf{+3.3$^{\uparrow*}$} & +0.5 & -- & -- & -- & -- \\
    \multicolumn{8}{l}{\textit{Transition features}} \\
    Agg + Trans & $-$0.1 & \textbf{+2.4$^{\uparrow*}$} & $-$0.2 & +0.2 & $-$0.3 & +0.2 & \textbf{+1.1$^{\uparrow*}$} \\
    Agg + Trans + Mlm-Cl & \textbf{+1.0$^{\uparrow}$} & \textbf{+2.2$^{\uparrow*}$} & $-$0.1 & \textbf{+1.5$^{\uparrow*}$} & +0.9 & +0.2 & \textbf{+1.8$^{\uparrow*}$} \\
    \multicolumn{8}{l}{\textit{TBS and hybrids}} \\
    JF + Trans & +0.6 & \textbf{+3.0$^{\uparrow*}$} & $-$0.4 & +1.1 & \textbf{+2.5$^{\uparrow}$} & +0.7 & +1.0 \\
    TBS & \textbf{+1.1$^{\uparrow*}$} & \textbf{+2.6$^{\uparrow*}$} & +0.2 & \textbf{+1.3$^{\uparrow*}$} & \textbf{+2.1$^{\uparrow*}$} & \textbf{+0.8$^{\uparrow*}$} & \textbf{+1.6$^{\uparrow*}$} \\
    TBS-R & \textbf{+1.0$^{\uparrow*}$} & \textbf{+2.7$^{\uparrow*}$} & $-$0.5 & \textbf{+1.7$^{\uparrow*}$} & +1.4 & +0.2 & \textbf{+1.8$^{\uparrow*}$} \\
    TBS-R, no pretraining & +0.0 & \textbf{+2.3$^{\uparrow*}$} & $-$0.3 & \textbf{+1.0$^{\uparrow*}$} & +1.1 & +0.2 & \textbf{+1.1$^{\uparrow*}$} \\
    TBS-R, no Trans in offset & \textbf{+1.1$^{\uparrow*}$} & \textbf{+2.9$^{\uparrow*}$} & $-$0.3 & \textbf{+1.3$^{\uparrow*}$} & \textbf{+1.3$^{\uparrow}$} & +0.2 & \textbf{+1.3$^{\uparrow*}$} \\
    TBS-R, Nep encoder & \textbf{+0.8$^{\uparrow}$} & \textbf{+2.8$^{\uparrow*}$} & $-$0.4 & \textbf{+1.4$^{\uparrow*}$} & +1.2 & +0.5 & \textbf{+1.4$^{\uparrow*}$} \\
    \multicolumn{8}{l}{\textit{Field encoder (second version)}} \\
    Nep, field encoder & +0.5 & \textbf{+1.4$^{\uparrow*}$} & $-$0.8 & \textbf{+1.4$^{\uparrow*}$} & +1.4 & +0.2 & +0.9 \\
    JF, field encoder & \textbf{+1.2$^{\uparrow}$} & \textbf{+2.9$^{\uparrow*}$} & $-$0.6 & +0.7 & \textbf{+2.0$^{\uparrow}$} & \textbf{+0.9$^{\uparrow}$} & \textbf{+1.7$^{\uparrow}$} \\
    JF + Trans, field encoder & +0.8 & \textbf{+2.6$^{\uparrow*}$} & $-$0.5 & +1.0 & +1.9 & \textbf{+1.0$^{\uparrow*}$} & +1.6 \\
    TBS, field encoder & \textbf{+1.3$^{\uparrow*}$} & \textbf{+2.7$^{\uparrow*}$} & $-$0.1 & \textbf{+1.4$^{\uparrow*}$} & \textbf{+2.0$^{\uparrow*}$} & \textbf{+1.0$^{\uparrow*}$} & \textbf{+2.2$^{\uparrow*}$} \\
 
    \bottomrule
  \end{tabular}
\end{table}

\begin{table}[p]
  \centering
  \caption{Summary over the seven targets and three label budgets of the transactions-only scenario: number of cells, cells whose 95\% interval against Agg lies above zero (nominal; with Benjamini--Hochberg $q < 0.05$; with the seeds resampled as well as the test rows; and with both), cells whose nominal interval lies below zero, and median difference in AUC points. BH and Seeds res.\ are separate analyses, and Both counts the cells that pass the two. Cells are not independent: the three budgets of a target share its test set. Ensemble controls are rank averages built from the saved predictions (Section~\ref{sec:models-ablations}).}
  \label{tab:res-summary}
  \footnotesize
  \begin{tabular}{lrrrrrrr}
    \toprule
    & & \multicolumn{4}{c}{Above zero} & & \\
    \cmidrule(lr){3-6}
    Method & Cells & Nominal & BH & Seeds res. & Both & Below & Median \\
    \midrule
    AE, Transformer, masked numeric & 21 & 0 & 0 & 0 & 0 & 0 & +0.1 \\
    AE, Transformer, + one-hot & 21 & 1 & 1 & 1 & 1 & 0 & +0.2 \\
    AE, MLP, whole sequence & 21 & 0 & 0 & 0 & 0 & 0 & $-$0.2 \\
    AE, MLP, masked events & 21 & 0 & 0 & 0 & 0 & 4 & $-$0.1 \\
    MLM (TabBERT-style) & 21 & 7 & 4 & 3 & 3 & 0 & +0.6 \\
    Nep (nuFormer-style) & 21 & 4 & 3 & 2 & 2 & 0 & +0.7 \\
    CoLES & 21 & 8 & 7 & 7 & 7 & 0 & +0.6 \\
    Mlm-Cl & 21 & 9 & 4 & 3 & 3 & 0 & +0.9 \\
    GRU & 21 & 3 & 3 & 3 & 3 & 4 & +0.2 \\
    JF-rand (no pretraining) & 21 & 6 & 5 & 3 & 3 & 3 & +0.2 \\
    JF (pretrained Nep + joint fusion) & 21 & 10 & 9 & 6 & 6 & 0 & +1.6 \\
    TabPFN v2 & 14 & 10 & 10 & 8 & 8 & 0 & +1.9 \\
    Expert features alone & 12 & 5 & 5 & 4 & 4 & 0 & +0.7 \\
    Agg + expert features & 9 & 5 & 5 & 5 & 5 & 0 & +1.9 \\
    Agg + Trans & 21 & 7 & 5 & 4 & 4 & 1 & +0.2 \\
    Agg + Trans + Mlm-Cl & 21 & 10 & 7 & 7 & 7 & 0 & +1.0 \\
    JF + Trans & 21 & 10 & 8 & 4 & 4 & 0 & +1.5 \\
    TBS & 21 & 19 & 18 & 17 & 17 & 0 & +2.1 \\
    TBS-R & 21 & 13 & 12 & 11 & 11 & 0 & +1.4 \\
    TBS-R, no pretraining & 21 & 8 & 8 & 5 & 5 & 1 & +0.4 \\
    TBS-R, no Trans in offset & 21 & 15 & 13 & 12 & 11 & 1 & +1.1 \\
    TBS-R, Nep encoder & 21 & 15 & 14 & 9 & 9 & 0 & +1.2 \\
    Nep, field encoder & 21 & 6 & 4 & 3 & 3 & 0 & +0.6 \\
    JF, field encoder & 21 & 15 & 8 & 8 & 6 & 0 & +1.4 \\
    JF + Trans, field encoder & 21 & 12 & 12 & 10 & 10 & 0 & +1.7 \\
    TBS, field encoder & 21 & 19 & 19 & 18 & 18 & 0 & +2.0 \\
    \multicolumn{8}{l}{\textit{Ensemble controls}} \\
    Ensemble of Agg and Agg + Trans & 21 & 16 & 15 & 11 & 11 & 0 & +0.6 \\
    Ensemble of Agg and JF & 21 & 19 & 19 & 17 & 17 & 0 & +2.1 \\
    Ensemble of Agg + Trans and JF-rand & 21 & 14 & 13 & 12 & 12 & 0 & +1.4 \\
    Ensemble of Agg + Trans and GRU & 21 & 15 & 14 & 12 & 12 & 1 & +1.3 \\
 
    \bottomrule
  \end{tabular}
\end{table}

\subsection{Frozen representations (RQ1)}
\label{sec:res-frozen}

Appending a frozen embedding to the aggregates changes the boosting model little.
The four reconstruction autoencoders have median differences between $-0.2$ and $+0.2$ points.
The encoders of the literature do better, but none clears the baseline in more than 9 of 21 cells.
Their largest gains are on the synthetic sequence trigger, whose signal is the order of events (CoLES +3.4 points with all labels), and on product 1 of MBD-mini and the B2B churn of client B (CoLES +1.9 on both).

The answer to RQ1 is therefore that frozen representations improve on the aggregates in a minority of cells, most clearly where the order of events carries signal that window counts miss, and even there by less than the methods that use the label.
This supports H1 in the form we state it: in the configurations we evaluated, representations learned without the label added little to what the boosting model already obtains from the aggregates, which does not show that they hold little additional information, and the generic reconstruction objectives, common in practice, were the least useful of all.

\subsection{Pretraining and fine-tuning (RQ2)}
\label{sec:res-finetune}

Joint fusion with the pretrained Nep backbone tends to do better than the same backbone frozen and fed to boosting, with a median of +0.5 points and no interval below zero, but only 3 of 21 cells pass both checks (Table~\ref{tab:res-targeted-summary}).
At our fixed configuration, pretraining improves joint fusion: the pretrained network beats the same network trained from scratch in 10 of 21 cells (6 passing both checks) and is below it in 1, and without pretraining the network is below the baseline in 3 cells.

Yet the fine-tuned network alone is an unreliable replacement for the boosting model.
Its median gain over the baseline is +1.6 points, but its interval clears zero in only 10 of 21 cells.
The supervised GRU, which uses the sequence without pretraining, is below the baseline in more cells than above it.
The answer to RQ2 has three parts, and H2 is only partly supported.
Pretraining contributes most of the value of the fine-tuned network in the joint-fusion configuration we tested.
The network on its own does not consistently outperform the boosting model.
The isolated effect of fine-tuning over using the network frozen is not identified by our contrast (Section~\ref{sec:models-ablations}).

\begin{table}[tbp]
  \centering
  \caption{Summary of the targeted comparisons (Tables~\ref{tab:res-targeted} and~\ref{tab:res-controls} in the appendix) over the seven targets and three label budgets of the transactions-only scenario, with the columns of Table~\ref{tab:res-summary}. BH adjusts within the family of the 273 targeted comparisons.}
  \label{tab:res-targeted-summary}
  \footnotesize
  \setlength{\tabcolsep}{4pt}
  \begin{tabular}{>{\raggedright\arraybackslash}p{5.4cm}rrrrrrr}
    \toprule
    & & \multicolumn{4}{c}{Above zero} & & \\
    \cmidrule(lr){3-6}
    Comparison & Cells & Nominal & BH & Seeds res. & Both & Below & Median \\
    \midrule
    JF vs.\ JF-rand (pretraining) & 21 & 10 & 8 & 6 & 6 & 1 & +0.8 \\
    JF vs.\ frozen Nep in boosting & 21 & 6 & 5 & 3 & 3 & 0 & +0.5 \\
    TBS vs.\ Agg + Trans & 21 & 13 & 13 & 13 & 13 & 0 & +1.2 \\
    TBS vs.\ JF & 21 & 8 & 7 & 3 & 3 & 0 & +0.5 \\
    TBS-R vs.\ Agg + Trans & 21 & 12 & 10 & 9 & 9 & 1 & +0.7 \\
    TBS-R vs.\ TBS-R without pretraining & 21 & 8 & 7 & 6 & 6 & 1 & +0.3 \\
    JF, field encoder vs.\ sum & 21 & 5 & 3 & 0 & 0 & 1 & +0.2 \\
    TBS, field encoder vs.\ sum & 21 & 3 & 0 & 0 & 0 & 1 & +0.1 \\
    \multicolumn{8}{l}{\textit{Ensemble controls}} \\
    Ensemble of Agg and Agg + Trans vs.\ Agg & 21 & 16 & 15 & 11 & 11 & 0 & +0.6 \\
    Ensemble with JF-rand vs.\ Agg + Trans & 21 & 11 & 9 & 9 & 8 & 1 & +0.5 \\
    TBS vs.\ ensemble with JF-rand (pretraining) & 21 & 8 & 5 & 6 & 5 & 1 & +0.3 \\
    TBS vs.\ ensemble of Agg and JF (transitions) & 21 & 4 & 2 & 3 & 2 & 0 & +0.1 \\
    TBS vs.\ ensemble of Agg and Agg + Trans (network) & 21 & 14 & 12 & 11 & 11 & 0 & +1.0 \\
 
    \bottomrule
  \end{tabular}
\end{table}

\subsection{TBS and the sources of its gain (RQ3)}
\label{sec:res-tbs}

Combining the two models works better than choosing between them.
TBS improves on its own boosting component, the aggregates with transition features, in 13 of 21 cells, and on its own network component in 8, mostly at the smaller budgets where the network alone is unstable; it is never below either component with an interval below zero.
TBS-R, which starts from the boosting score and keeps a correction only when validation improves, clears the baseline in 13 of 21 cells.
Its pretrained encoder improves on the same residual without pretraining in 8 of 21 cells, spread over the three budgets, and the variants with the Nep encoder and without transitions in the offset clear the baseline most often, in 15 of 21 cells each.

Against the baseline, TBS changes three things at once: it adds the transition features, it adds the network, and it averages two models.
The ensemble controls separate the three (Table~\ref{tab:res-controls} in the appendix and the last rows of Table~\ref{tab:res-summary}).
Averaging the baseline with the boosting model on aggregates and transitions, a purely tabular ensemble, is above the baseline in 16 of 21 cells but with a median of only +0.6 points.
Replacing the pretrained network by the same network trained from scratch leaves a median gain of +1.4 points over the baseline and +0.5 over the boosting component.
Removing the transition features instead, so that the baseline itself is averaged with the network, costs less: that ensemble is above the baseline in 19 of 21 cells.
Compared cell by cell, the full TBS is ahead of it by a median of +0.1 points, with an interval above zero in 4 cells and 2 passing both checks.
This paired median, not the difference between the two medians over the baseline, is the contribution of the transition features.

The medians point to combining boosting with a fine-tuned sequence network as the main source of the gain, with pretraining that network adding a smaller part (paired median +0.3 points); once the network is present, the transition features add a small amount, and averaging two boosting models gives a smaller but consistent gain.
Cell by cell the evidence is weaker than the medians: with both checks, TBS beats the control with the untrained network in 5 of 21 cells and the tabular ensemble in 11 (Table~\ref{tab:res-targeted-summary}).
It is never below the tabular ensemble, and its interval is below the untrained-network control in 1 cell, product 1 of MBD-mini with 1,000 labels, which does not hold with the seeds resampled.
Encoding the fields of each event with a small Transformer instead of summing their embeddings, the change that helped most on TabFormer (Section~\ref{sec:res-tabformer}), makes the fine-tuned network more consistent on its own: joint fusion with the field encoder clears the baseline in 15 of 21 cells, against 10 with the sum.
TBS barely changes with the field encoder (median +0.1 points compared directly with the sum), because the combination with boosting already recovers most of what the better encoder adds.
The answer to RQ3 is that TBS improves on both of its components and that most of its gain comes from adding a fine-tuned sequence network to the boosting model, more than from the transition features or from averaging two boosting models; pretraining that network adds a smaller part, so H3 is only partly supported.

\subsection{Label budgets and tabular information (RQ4)}
\label{sec:res-budget}

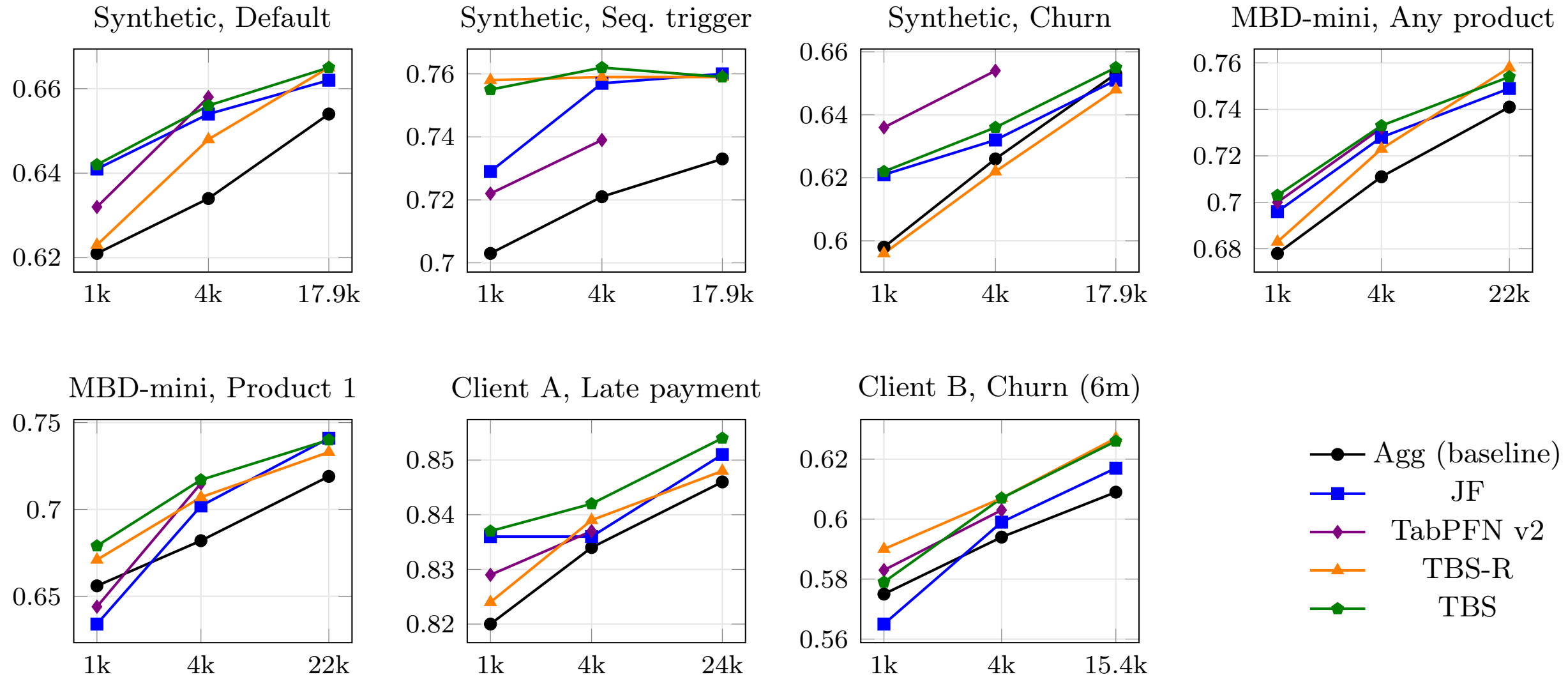
\begin{figure}[t]
  \centering
  \resizebox{\linewidth}{!}{\begin{tikzpicture}
\begin{groupplot}[group style={group size=4 by 2, horizontal sep=1.25cm, vertical sep=1.6cm},
  width=4.6cm, height=4.0cm, xmode=log, log basis x=10, title style={font=\footnotesize, yshift=-1ex},
  tick label style={font=\scriptsize}, yticklabel style={/pgf/number format/fixed, /pgf/number format/precision=2},
  ylabel style={font=\scriptsize}, grid=major, grid style={black!10},
  legend style={font=\footnotesize, draw=none}]
\nextgroupplot[legend to name=auclegend, legend columns=1, title={Synthetic, Default}, xtick={1000,4000,17879}, xticklabels={1k,4k,17.9k}]
\addplot[color=black, mark=*, mark size=1.6pt, thick] coordinates {(1000,0.621) (4000,0.634) (17879,0.654)};\addlegendentry{Agg (baseline)}
\addplot[color=blue, mark=square*, mark size=1.6pt, thick] coordinates {(1000,0.641) (4000,0.654) (17879,0.662)};\addlegendentry{JF}
\addplot[color=violet, mark=diamond*, mark size=1.6pt, thick] coordinates {(1000,0.632) (4000,0.658)};\addlegendentry{TabPFN v2}
\addplot[color=orange, mark=triangle*, mark size=1.6pt, thick] coordinates {(1000,0.623) (4000,0.648) (17879,0.665)};\addlegendentry{TBS-R}
\addplot[color=green!50!black, mark=pentagon*, mark size=1.6pt, thick] coordinates {(1000,0.642) (4000,0.656) (17879,0.665)};\addlegendentry{TBS}
\nextgroupplot[title={Synthetic, Seq.\ trigger}, xtick={1000,4000,17879}, xticklabels={1k,4k,17.9k}]
\addplot[color=black, mark=*, mark size=1.6pt, thick] coordinates {(1000,0.703) (4000,0.721) (17879,0.733)};
\addplot[color=blue, mark=square*, mark size=1.6pt, thick] coordinates {(1000,0.729) (4000,0.757) (17879,0.760)};
\addplot[color=violet, mark=diamond*, mark size=1.6pt, thick] coordinates {(1000,0.722) (4000,0.739)};
\addplot[color=orange, mark=triangle*, mark size=1.6pt, thick] coordinates {(1000,0.758) (4000,0.759) (17879,0.759)};
\addplot[color=green!50!black, mark=pentagon*, mark size=1.6pt, thick] coordinates {(1000,0.755) (4000,0.762) (17879,0.759)};
\nextgroupplot[title={Synthetic, Churn}, xtick={1000,4000,17879}, xticklabels={1k,4k,17.9k}]
\addplot[color=black, mark=*, mark size=1.6pt, thick] coordinates {(1000,0.598) (4000,0.626) (17879,0.653)};
\addplot[color=blue, mark=square*, mark size=1.6pt, thick] coordinates {(1000,0.621) (4000,0.632) (17879,0.651)};
\addplot[color=violet, mark=diamond*, mark size=1.6pt, thick] coordinates {(1000,0.636) (4000,0.654)};
\addplot[color=orange, mark=triangle*, mark size=1.6pt, thick] coordinates {(1000,0.596) (4000,0.622) (17879,0.648)};
\addplot[color=green!50!black, mark=pentagon*, mark size=1.6pt, thick] coordinates {(1000,0.622) (4000,0.636) (17879,0.655)};
\nextgroupplot[title={MBD-mini, Any product}, xtick={1000,4000,22000}, xticklabels={1k,4k,22k}]
\addplot[color=black, mark=*, mark size=1.6pt, thick] coordinates {(1000,0.678) (4000,0.711) (22000,0.741)};
\addplot[color=blue, mark=square*, mark size=1.6pt, thick] coordinates {(1000,0.696) (4000,0.728) (22000,0.749)};
\addplot[color=violet, mark=diamond*, mark size=1.6pt, thick] coordinates {(1000,0.700) (4000,0.732)};
\addplot[color=orange, mark=triangle*, mark size=1.6pt, thick] coordinates {(1000,0.683) (4000,0.723) (22000,0.758)};
\addplot[color=green!50!black, mark=pentagon*, mark size=1.6pt, thick] coordinates {(1000,0.703) (4000,0.733) (22000,0.754)};
\nextgroupplot[title={MBD-mini, Product 1}, xtick={1000,4000,22000}, xticklabels={1k,4k,22k}]
\addplot[color=black, mark=*, mark size=1.6pt, thick] coordinates {(1000,0.656) (4000,0.682) (22000,0.719)};
\addplot[color=blue, mark=square*, mark size=1.6pt, thick] coordinates {(1000,0.634) (4000,0.702) (22000,0.741)};
\addplot[color=violet, mark=diamond*, mark size=1.6pt, thick] coordinates {(1000,0.644) (4000,0.715)};
\addplot[color=orange, mark=triangle*, mark size=1.6pt, thick] coordinates {(1000,0.671) (4000,0.707) (22000,0.733)};
\addplot[color=green!50!black, mark=pentagon*, mark size=1.6pt, thick] coordinates {(1000,0.679) (4000,0.717) (22000,0.740)};
\nextgroupplot[title={Client A, Late payment}, xtick={1000,4000,24000}, xticklabels={1k,4k,24k}]
\addplot[color=black, mark=*, mark size=1.6pt, thick] coordinates {(1000,0.820) (4000,0.834) (24000,0.846)};
\addplot[color=blue, mark=square*, mark size=1.6pt, thick] coordinates {(1000,0.836) (4000,0.836) (24000,0.851)};
\addplot[color=violet, mark=diamond*, mark size=1.6pt, thick] coordinates {(1000,0.829) (4000,0.837)};
\addplot[color=orange, mark=triangle*, mark size=1.6pt, thick] coordinates {(1000,0.824) (4000,0.839) (24000,0.848)};
\addplot[color=green!50!black, mark=pentagon*, mark size=1.6pt, thick] coordinates {(1000,0.837) (4000,0.842) (24000,0.854)};
\nextgroupplot[title={Client B, Churn (6m)}, xtick={1000,4000,15376}, xticklabels={1k,4k,15.4k}]
\addplot[color=black, mark=*, mark size=1.6pt, thick] coordinates {(1000,0.575) (4000,0.594) (15376,0.609)};
\addplot[color=blue, mark=square*, mark size=1.6pt, thick] coordinates {(1000,0.565) (4000,0.599) (15376,0.617)};
\addplot[color=violet, mark=diamond*, mark size=1.6pt, thick] coordinates {(1000,0.583) (4000,0.603)};
\addplot[color=orange, mark=triangle*, mark size=1.6pt, thick] coordinates {(1000,0.590) (4000,0.607) (15376,0.627)};
\addplot[color=green!50!black, mark=pentagon*, mark size=1.6pt, thick] coordinates {(1000,0.579) (4000,0.607) (15376,0.626)};
\end{groupplot}
\node[anchor=west, align=left] at ($(group c3r2.east)+(1.6cm,0)$) {\pgfplotslegendfromname{auclegend}};
\end{tikzpicture}}
  \caption{Test AUC by number of training labels (log scale), transactions-only scenario, mean over five seeds: the baseline (Agg) and four of the methods. TabPFN v2 runs only with 1,000 and 4,000 labels. Paired differences against Agg with their intervals are in Table~\ref{tab:res-all} and, for the smaller budgets, in Tables~\ref{tab:res-1k} and~\ref{tab:res-4k}.}
  \label{fig:auc-budgets}
\end{figure}

With 1,000 labels (Table~\ref{tab:res-1k} in the appendix), TabPFN v2 on the aggregates has an interval above zero on 5 of the 7 targets, but on only 3 with the seeds resampled, since a 1,000-label subsample varies from seed to seed; the hybrids are strong as well.
At this budget the fine-tuned network alone ranges from $-2.1$ to +2.6 points, and the GRU and the network without pretraining are often far below the baseline.
As labels grow, the boosting baseline improves faster than most alternatives (Figure~\ref{fig:auc-budgets}).
The gains of TBS shrink but persist: +0.4 to +5.2 points with 1,000 labels and +0.2 to +2.6 with all labels, above zero on 6 of 7 targets.

When every method also receives the register and statement data (synthetic), the static attributes of the point of sale (client A) or the twelve variables of the production model (client B), the sequence adds less (Table~\ref{tab:res-full} in the appendix).
Over the 15 target--budget cells available in both scenarios, the median gain of TBS falls from +1.6 to +1.2 points, with 11 cells still above zero.
On synthetic default, where the register carries the latent risk, the baseline AUC rises from 0.654 to 0.741 and no sequence method adds a detectable amount.
The answer to RQ4 is that the gain of the sequence models depends on both coordinates: it is largest with few labels and without rich tabular data, and it shrinks, without disappearing for TBS, as either grows, which supports H4.

\subsection{Distance to the latent-information oracle}
\label{sec:res-oracle}

On the synthetic data we know the latent-information oracle (Section~\ref{sec:setup-datasets}).
With all labels and transactions only, the baseline is 12.3 points below it for default, 4.6 for the sequence trigger and 3.2 for churn.
These gaps measure what a model that knew the latent variables would gain, not what a better transaction model could recover, because part of the latent signal never reaches the transactions.
On default and churn the best method closes only 10\% and 15\% of the gap, and for default much of the missing signal is in the register.

The sequence trigger is the case that window aggregates cannot express.
There, the 96 expert features, written with knowledge of the pattern, close 78\% of the gap (AUC 0.769), the fine-tuned network 59\%, TBS 57\% and the transition features, mined without the label, 52\%.
For this target we can also say how much of the signal the transactions reveal: the count of the pattern, which can be computed exactly from the transactions, reaches AUC 0.763 on its own.
The expert features exceed it because the transactions also carry part of the credit risk that enters the label, so the remaining gap is partly but not wholly recoverable from transactions.

\subsection{Re-examination of public benchmarks (RQ5)}
\label{sec:benchmarks}

Reported results for foundation models on transactions often rest on two public synthetic datasets from IBM: the TabFormer card transactions~\citep{padhi2021tabbert} and the anti-money-laundering (AML) transfers of \citet{altman2023aml}.
We re-examined both with the questions of Section~\ref{sec:methodology}: what the test measures, what the baseline sees, and whether the compared models see the same information.
Unlike the main study, each benchmark keeps its own split and metrics, AUPRC and F1, so that our numbers can be compared with the published ones, and our runs use two or three seeds for the boosting models and one for the graph networks.

\subsubsection{Card fraud (TabFormer)}
\label{sec:res-tabformer}

TabFormer has 24.4 million transactions from 2,000 users and 6,139 cards between 1991 and 2020, with 29,757 frauds (0.12\% of transactions).
Published evaluations use the last 10\% of transactions by date as the test set, here 2.4 million transactions from October 12, 2018 on, with 2,718 frauds.
NeoSpace, a commercial vendor of transaction foundation models, publishes results of its model Cortex on TabFormer.
Our main analysis audits the version in commit \texttt{3215436} of its public repository~\citep{neospace2026ldm}, unchanged since June 2026 and current when we downloaded the scores.
That version states training on 1991--2017, validation on 2017--2018 and a test on 2018--2020 with 2,412,326 transactions and 2,698 frauds.
It reports AUPRC 0.989 and F1 0.959, with the F1 threshold chosen on the test set (its notebook takes the maximum over the test precision--recall curve), against gradient boosting on the 13 raw columns with AUPRC 0.19.
NeoSpace also published per-transaction Cortex scores for the whole dataset, as four Parquet files, which allowed a paired comparison on the same rows.
We downloaded them on 29 October and again on 6 November 2026, with identical SHA-256 digests (Appendix~\ref{app:repro}).
On those files, the last 10\% of transactions gives AUPRC 0.989 and F1 0.961, and it is the window closest to the repository's counts; we use it as the test and could not identify which rows those counts exclude.
Released scores cannot show the training period of a model, so we take the period NeoSpace states as given.
A comparison on released scores does not amount to running both models under the same conditions.
The end of this subsection evaluates the different protocol that NeoSpace's benchmark page describes today.

\begin{figure}[t]
  \centering
  \includegraphics[width=0.8\linewidth]{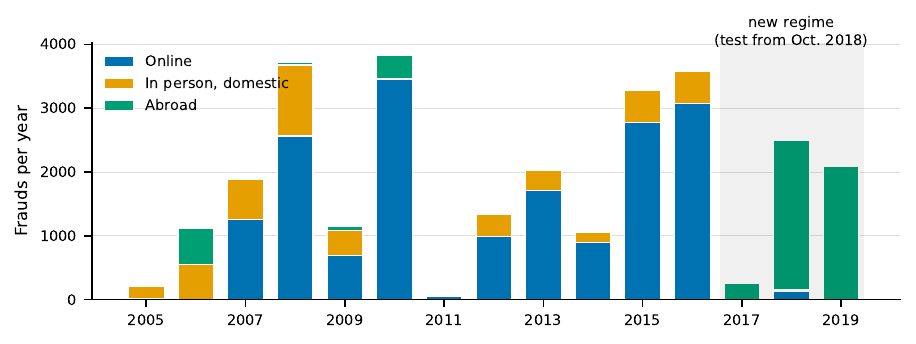}
  \caption{TabFormer frauds per year by channel, 2005--2019. No fraud occurs in the first three quarters of 2017; from the fourth quarter on, almost all frauds happen abroad. The test set (from October 2018) lies entirely in the new regime.}
  \label{fig:tabformer}
\end{figure}

The fraud regime changes before the test.
No fraud occurs in the first three quarters of 2017, and from the fourth quarter of 2017 on, 97\% of frauds happen abroad, while in 2014--2016 none did and 85\% were online purchases.
The test falls entirely in the new regime, and the only examples of it before 2018 are the 255 frauds of late 2017 (Figure~\ref{fig:tabformer}).
Whatever a model scores on this test therefore depends first on whether its training data reach past the regime change.
NeoSpace states that Cortex was trained through 2017, which contains only those frauds of the new regime.
We measure what baselines reach under each assumption, with three training cut-offs against the same test (Table~\ref{tab:res-tabformer}): up to May 2017 (strict), up to December 2017 (calendar, the period NeoSpace states), and up to the day before the test (through, which gives our models about nine months of data that NeoSpace states Cortex did not use for training).

LightGBM is trained with all frauds and 10\% of legitimate transactions, with two seeds, on three feature levels: the 13 raw columns of the transaction; the generic window summary over 1, 7 and 30 days that our platform generates automatically; and 72 hand-written card-history features, such as the time since the previous purchase, the purchases per hour, and whether the merchant and the place are new to the card.
Under the strict cut-off no feature set recovers the new regime (AUPRC 0.060 with card history).
With training through the day before the test, the 13 raw columns reach 0.862 and card history reaches 0.980.
On the same rows, card history minus the Cortex scores is $-0.0082$ AUPRC (95\% CI $-0.0145$ to $-0.0024$), so card history stays below Cortex even with those nine extra months.
Under the calendar cut-off, the period NeoSpace states, card history reaches 0.809 and its difference from the Cortex scores is $-0.180$ (CI $-0.203$ to $-0.158$).
Here and below, intervals come from a paired bootstrap over users rather than transactions, because frauds come in runs within a user.

\begin{table}[t]
  \centering
  \caption{TabFormer, last 10\% of transactions as test. AUPRC and F1 at a threshold chosen on validation, by training cut-off: strict (training data up to May 2017), calendar (up to December 2017) and through (up to the day before the test). LightGBM rows: mean of two seeds. FM: per-transaction foundation model, out-of-fold scores. Published Cortex scores~\citep{neospace2026ldm} on the same test: AUPRC 0.989, F1 0.961 at the test-optimal threshold.}
  \label{tab:res-tabformer}
  \footnotesize
  \begin{tabular}{lrrrrrr}
    \toprule
    & \multicolumn{2}{c}{Strict} & \multicolumn{2}{c}{Calendar} & \multicolumn{2}{c}{Through} \\
    \cmidrule(lr){2-3} \cmidrule(lr){4-5} \cmidrule(lr){6-7}
    Features & AUPRC & F1 & AUPRC & F1 & AUPRC & F1 \\
    \midrule
    13 raw columns & 0.022 & 0.017 & 0.135 & 0.103 & 0.862 & 0.580 \\
    Window summary (1/7/30 days) & -- & -- & 0.605 & 0.545 & 0.957 & 0.879 \\
    Card history (72 features) & 0.060 & 0.060 & 0.809 & 0.656 & 0.980 & 0.925 \\
    FM, first version, alone & -- & -- & 0.861 & 0.742 & 0.977 & 0.941 \\
    Window summary + FM score (v1) & -- & -- & 0.896 & 0.826 & 0.983 & 0.941 \\
    Card history + FM score (v1) & -- & -- & 0.936 & 0.850 & 0.990 & 0.955 \\
    FM, field encoder, alone & -- & -- & 0.961 & 0.871 & 0.994 & 0.979 \\
    Card history + FM score (field) & -- & -- & 0.965 & 0.916 & 0.994 & 0.969 \\
    \bottomrule
  \end{tabular}
\end{table}

We also trained a per-transaction foundation model: a causal Transformer over the transactions of each user, with all cards interleaved, 13 categorical fields and 3 continuous values per transaction, and windows of 256 transactions with stride 128, for about 8.6 million parameters.
It is pretrained to predict the next transaction on data before 2018 and fine-tuned to score every transaction, and three folds by user give out-of-fold scores, so that the score can enter LightGBM as a feature without leaking labels.
As in Section~\ref{sec:models-backbone}, a first version sums the field embeddings and a second encodes the fields of each transaction with a two-layer Transformer.
The second version was selected by screening three changes on one third of the users, and the gain holds on the other two thirds (+0.0125, CI +0.0077 to +0.0176, in the through protocol).

In the through protocol, the second version alone reaches AUPRC 0.994, above the Cortex scores on the same rows by +0.0054 (CI +0.0025 to +0.0091), with F1 0.979 at a threshold chosen on validation.
That comparison favors our model, which saw training data that NeoSpace states Cortex did not.
In the calendar protocol, where the new regime has only 255 training examples, the foundation model makes a larger difference for us: 0.961 alone against 0.809 for card history.
It remains below the Cortex scores: the difference is $-0.027$ (CI $-0.035$ to $-0.019$) alone and $-0.023$ (CI $-0.031$ to $-0.016$) combined with card history.
Pretraining helps only there: on the users unseen by fold 0, the pretrained first version scores 0.775 against 0.719 without pretraining in the calendar protocol, and 0.955 against 0.956 in the through protocol.

As of 9 November 2026, the benchmark page~\citep{neospace2026cortex} describes training on 1991--2017, validation on 2018 and a test on 2019--2020 (2,034,720 transactions, 2,068 frauds), an F1 threshold set on validation before 2019, and a raw-column baseline of AUPRC 0.14.
These differ from the audited version; the score files do not.
On those files, transactions dated 2019--2020 are 2,060,438 with 2,087 frauds, and we could not identify which rows the page's counts exclude.
On them, Cortex has AUPRC 0.989 and F1 0.960 at a threshold chosen on 2018 (0.962 at the test-optimal one), so the choice of threshold moves its F1 by about 0.002.
With training through 2017, card history has AUPRC 0.759, a difference from Cortex of $-0.231$ (CI $-0.260$ to $-0.203$), and our foundation model combined with card history 0.962, a difference of $-0.027$ (CI $-0.039$ to $-0.018$).

The size of the published gap depends on the comparison.
The baseline NeoSpace reports has no history.
Our own model on the same 13 raw columns reaches 0.135 with training through 2017, close to the 0.19 of the audited version and the 0.14 of the current page, while card history reaches 0.809, so a comparison with a raw-column baseline does not indicate the gap to a history-aware one.
Against a history-aware baseline, the size of the gap depends on the training cut-off.
With the period NeoSpace states, the models in Table~\ref{tab:res-tabformer} stay below the Cortex scores, by 0.180 AUPRC for card history and 0.023 for the best combination in the table.
Even when our baseline trains through the day before the test, on about nine more months of the new regime that NeoSpace states Cortex did not use, it stays below the Cortex scores, by 0.008 AUPRC.
As with any released scores, the stated training period cannot be verified from them.
The test-set threshold of the audited version matters little for Cortex.

After these comparisons, we ran an additional exploratory experiment with the training period NeoSpace states (Table~\ref{tab:res-tabformer-periods}).
The 18 candidates were variants of the field-encoder foundation model, CatBoost alone, an ensemble of three CatBoost and three LightGBM seeds on 93 point-in-time features, and rank averages of these.
CatBoost hyperparameters come from a Bayesian search (Optuna, 13 trials) that maximized AUPRC on all of 2018, on a closely related set of 94 features, and an analysis of errors on 2018 guided part of the feature design.
The selected candidate is the rank average of that ensemble and the foundation model with three seeds and early stopping evaluated four times per epoch.
It was selected on validation and evaluated on a test set previously explored by this study, because earlier models had already been evaluated on these rows, so this is not an independent confirmation.
On 2019--2020, the window disjoint from the search, it has AUPRC 0.999 against 0.989 for the Cortex scores, a difference of $+0.0092$ (CI $+0.0055$ to $+0.0129$).
On the last 10\% of transactions, whose first weeks, up to December 31, 2018, fall inside the search period, it has AUPRC 0.993 against 0.989, a difference of $+0.0045$ (CI $+0.0011$ to $+0.0088$).
The reported confidence intervals do not account for model-selection or adaptive-development bias.
All frauds from 2019 on come from a single simulated fraud pattern, and the stated training period of Cortex cannot be verified from its scores, so we do not read this result as conclusive against that product.

\begin{table}[t]
  \centering
  \caption{Data periods of the additional TabFormer experiment. The hyperparameter search and the error analysis overlap the first test window from October 12 to December 31, 2018, and are disjoint from the 2019--2020 window.}
  \label{tab:res-tabformer-periods}
  \footnotesize
  \begin{tabular}{ll}
    \toprule
    Use & Period \\
    \midrule
    Pretraining and model training & up to December 31, 2017 \\
    Early stopping & January 1 to June 30, 2018 \\
    CatBoost hyperparameter search and error analysis & January 1 to December 31, 2018 \\
    Selection among the 18 candidates & July 1, 2018 to the start of the test \\
    Test, last 10\% of transactions & from October 12, 2018 \\
    Test, 2019--2020 & from January 1, 2019 \\
    \bottomrule
  \end{tabular}
\end{table}

\subsubsection{Anti-money laundering (IBM AML)}
\label{sec:res-aml}

The IBM AML datasets simulate transfers between accounts of many banks, with laundering patterns injected by the simulator~\citep{altman2023aml}.
We use HI-Small, with 5.1 million transfers and 5,177 laundering transfers, and LI-Small, with 6.9 million and 3,565, with the day-based split of Multi-GNN~\citep{egressy2024multignn}: training on days 0--5, validation on days 6--7 and test from day 8.

\begin{figure}[t]
  \centering
  \includegraphics[width=0.85\linewidth]{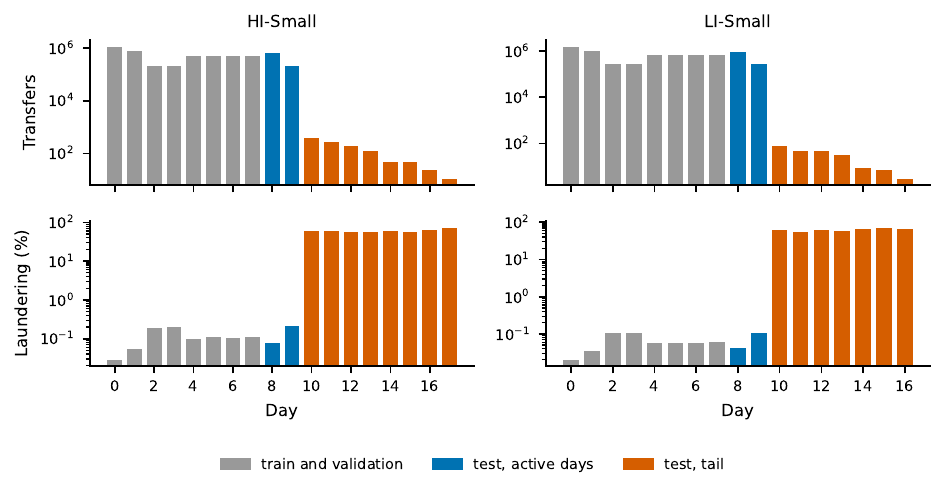}
  \caption{IBM AML transfers per day (top) and share of laundering per day (bottom), both on a log scale. From day 10 on, normal activity stops while laundering chains continue, so the last days of the test are mostly laundering.}
  \label{fig:aml}
\end{figure}

From day 10 on, normal activity stops and only laundering chains continue (Figure~\ref{fig:aml}).
In the HI-Small test, the 1,108 transfers of this tail are 59\% laundering and hold 41\% of the laundering of the test; in LI-Small, the tail has 223 transfers, 60\% laundering and 17\% of the laundering of the test.
The rule ``flag every transfer dated day 10 or later'' alone has F1 0.482 on HI-Small and 0.261 on LI-Small.
We therefore report the whole test, to compare with published numbers, and the active days 8 and 9 separately.

The published numbers come from different protocols.
The gradient-boosting results of \citet{altman2023aml} use graph features computed from past transactions only.
The Multi-GNN code builds the test graph with all transactions, including those after the transfer being classified, samples neighbors without regard to time and uses the absolute timestamp as an edge attribute, and its F1 is computed at the argmax of the two-class output.\footnote{Commit 252b025 of \url{https://github.com/IBM/Multi-GNN}: \texttt{data\_loading.py}, lines 32 and 102; \texttt{train\_util.py}, lines 86--93.}
The published table therefore compares models that see different information.

As a point-in-time baseline, we built LightGBM features from the history of the sender and of the receiver over 6, 24 and 72 hours and the whole past, the history of the pair in both directions and the amount relative to the mean of the account: 70 features computed only from transactions of earlier minutes, which we checked by brute force on sampled rows.
The threshold is chosen on validation.
Table~\ref{tab:res-aml} compares this baseline with the published results and with our runs of the Multi-GNN code of the authors, one seed each, rescored on the same slices.

\begin{table}[t]
  \centering
  \caption{IBM AML, day-based split of Multi-GNN. F1 of the laundering class on the whole test, and AUPRC and F1 on the active days (8--9), where the simulation tail no longer helps. Published: mean $\pm$ s.d.\ as reported~\citep{altman2023aml,egressy2024multignn}, argmax threshold for the GNNs. Ours: LightGBM mean $\pm$ s.d.\ over three seeds, threshold chosen on validation; Multi-GNN, our run of the authors' code (one seed), argmax / validation threshold.}
  \label{tab:res-aml}
  \footnotesize
  \setlength{\tabcolsep}{3pt}
  \resizebox{\linewidth}{!}{%
  \begin{tabular}{lcccccc}
    \toprule
    & \multicolumn{3}{c}{HI-Small} & \multicolumn{3}{c}{LI-Small} \\
    \cmidrule(lr){2-4} \cmidrule(lr){5-7}
    & F1 test & AUPRC act. & F1 act. & F1 test & AUPRC act. & F1 act. \\
    \midrule
    Published: LightGBM+GFs & 0.629 $\pm$ 0.003 & -- & -- & 0.208 $\pm$ 0.015 & -- & -- \\
    Published: XGBoost+GFs & 0.632 $\pm$ 0.002 & -- & -- & 0.273 $\pm$ 0.003 & -- & -- \\
    Published: Multi-GIN+EU & 0.648 $\pm$ 0.012 & -- & -- & 0.269 $\pm$ 0.066 & -- & -- \\
    Published: Multi-PNA+EU & 0.682 $\pm$ 0.027 & -- & -- & 0.331 $\pm$ 0.026 & -- & -- \\
    LightGBM, transfer fields only & 0.223 $\pm$ 0.005 & 0.069 $\pm$ 0.000 & 0.147 $\pm$ 0.001 & 0.080 $\pm$ 0.007 & 0.014 $\pm$ 0.000 & 0.049 $\pm$ 0.004 \\
    LightGBM, point-in-time history & 0.666 $\pm$ 0.045 & 0.519 $\pm$ 0.037 & 0.526 $\pm$ 0.065 & 0.373 $\pm$ 0.018 & 0.218 $\pm$ 0.007 & 0.263 $\pm$ 0.016 \\
    Multi-GIN+EU, our run & 0.584 / 0.663 & 0.498 & 0.418 / 0.530 & 0.342 / 0.323 & 0.155 & 0.220 / 0.205 \\
    Multi-PNA+EU, our run & 0.687 / 0.701 & 0.516 & 0.552 / 0.573 & 0.372 / 0.367 & 0.180 & 0.258 / 0.253 \\
    Rule: day $\geq$ 10 & 0.482 & -- & -- & 0.261 & -- & -- \\
    \bottomrule
  \end{tabular}}
\end{table}

On the whole test, the point-in-time baseline reaches F1 0.666 on HI-Small and 0.373 on LI-Small, at the level of the best published GNN on HI-Small and above it on LI-Small, using only earlier transactions.
On the active days, where the tail no longer helps, Multi-PNA has AUPRC 0.516 against 0.519 for the baseline on HI-Small, and 0.180 against 0.218 on LI-Small.
The published F1 also depends on the threshold: the argmax corresponds to a probability of 0.5, while on HI-Small the F1-optimal validation threshold of Multi-GIN is 0.84.
This says nothing about calibration, since even a calibrated model need not maximize F1 at 0.5; it means that the published F1 uses an untuned threshold.

We also carried the per-transaction foundation model of Section~\ref{sec:res-tabformer} to accounts, with each transfer an event in the sequences of both the sender and the receiver and the counterparty as a hashed field.
Attention cannot see events of the same minute, and fields that look back count only earlier minutes.
On the active days of HI-Small, the rank ensemble of this model and the baseline reaches AUPRC 0.557 and 0.566 with two of three model seeds, against 0.519 for the baseline, and 0.513 with the third; on LI-Small the ensemble stays below the baseline (0.204).
Without pretraining, the model alone drops from 0.480 to 0.317 AUPRC.
The evidence for a gain from sequence models on this benchmark is thus seed-dependent and limited to one of the two datasets.

The answer to RQ5 is that, on both benchmarks, protocol choices produce differences larger than those between the models themselves: the training cut-off and a baseline without history on TabFormer, and a simulation tail, an untuned threshold and a test graph that includes later transactions on IBM AML.
On IBM AML, a point-in-time baseline reaches the published F1 of the graph networks.
On TabFormer, with the training period the vendor states, the published model stays ahead of our history-aware baseline and of our foundation model, the baseline stays below it even when it receives later data, and an additional exploratory candidate scored above it on a test set previously explored by this study.
This supports H5, except for the decision threshold on TabFormer, which moves the F1 of Cortex by about 0.002, while a foundation model can still add a measurable amount where the protocol leaves it room, as with few examples of a new fraud regime.

\subsection{Computational cost}
\label{sec:cost}

Cost is often cited as a reason to doubt or to defer transaction foundation models, so we measured it.
All times are wall-clock on one NVIDIA A10G (AWS g5.4xlarge, US\$1.624 per hour on demand in us-east-1, price checked on 2026-11-04), from an earlier execution of the pipeline in which every dataset ran on that GPU; the final execution of Section~\ref{sec:setup-config} used an L4 for client A.
Some runs shared the GPU with other jobs, so we report ranges.
For the per-client models of the main study, pretraining the Nep backbone took between 81\,s and 10\,min per dataset, one fine-tuning run with all labels between 29\,s and 66\,s, and one LightGBM fit a few seconds (Table~\ref{tab:res-cost}).
Because each client is truncated to its last 128 events, GPU time grows with the number of clients, not with the number of transactions.

On the 24 million transactions of TabFormer, the first version of the per-transaction model pretrained in 11\,min, fine-tuned each of the three folds in 7\,min and scored the whole dataset in 29\,s, about US\$0.86 in all, and the field encoder took 37\,min, 21\,min and 84\,s, about US\$2.73.
At the scale we tested, compute is therefore not an obstacle to trying these models.
Appendix~\ref{app:cost} extrapolates these rates to a bank with 100 million clients and estimates the cost of models of the size reported in the literature; both are estimates conditional on stated assumptions, not measurements.

\begin{table}[t]
  \centering
  \caption{Measured wall-clock time on one A10G (US\$1.624 per hour). Per-client models: pretraining of the Nep backbone, one joint-fusion fine-tuning run and one LightGBM fit with all labels.}
  \label{tab:res-cost}
  \footnotesize
  \begin{tabular}{lrrrr}
    \toprule
    Dataset & Labels & Pretraining & Fine-tuning & LightGBM \\
    \midrule
    Synthetic & 17,879 & 10 min & 53 s & 2.5 s \\
    MBD-mini & 22,000 & 8 min & 66 s & 2.8 s \\
    Client A & 24,000 & 81 s & 31 s & 1.9 s \\
    Client B & 15,376 & 5 min & 29 s & 1.5 s \\
    \bottomrule
  \end{tabular}
\end{table}

\section{Discussion}
\label{sec:discussion}

The results support a small number of practical statements for models of the size we tested, with fixed configurations and five training runs per cell.
This section interprets them together, relates the main study to the re-examination of the public benchmarks, and turns both into recommendations for practitioners who evaluate a transaction foundation model.

\subsection{What the evidence supports}
\label{sec:disc-evidence}

Representations learned without the label and appended to a boosting model on window aggregates add little in the configurations we evaluated.
Where they help, as on the synthetic sequence trigger or on the B2B churn of client B, the target depends on the order or the rhythm of events, which window aggregates summarize poorly; where the signal is in the levels and counts that aggregates already capture, the boosting model has little left to learn from them.
A sequence model fine-tuned on the target adds a small amount, and it does so most consistently when it is combined with the boosting model rather than used in its place.
In that combination, the controls suggest that most of the observed gain comes from adding a sequence network to the boosting model, pretraining that network adds a median of +0.3 points, the transition features add +0.1 once the network is present, and averaging two boosting models gives a smaller but consistent gain.

These statements depend on two coordinates that a single headline number usually omits: the number of labels and what the baseline already knows.
With 1,000 labels, TabPFN on the aggregates has an interval above a boosting model trained from scratch on 5 of 7 targets and the fine-tuned sequence network on 5; with all labels and with the tabular attributes of a production model, the boosting baseline closes much of the gap, and only the combination keeps a consistent advantage.
A reported gain without these two coordinates, and without an estimate of its uncertainty, therefore says little about what a given institution can expect.

\subsection{Protocol choices and measured differences}
\label{sec:disc-protocol}

The differences we measured between methods in the main study are of one to a few AUC points.
The re-examination of the public benchmarks shows that protocol choices can produce much larger differences: a training cut-off before a change in the fraud pattern, a baseline without history, an untuned decision threshold, a simulation tail in which a date rule finds most of the laundering, and a test graph that includes transactions after the one being classified.
None of these choices is specific to foundation models, and each can change the comparison between a more complex model and a simpler one.

The two parts of the study complement each other.
In the main study, where the baseline is history-aware, point in time and trained on the same period as the candidate, the gap between foundation models and boosting is small enough that its sign and size depend on the number of labels, the tabular information and the seed, and it can only be measured with paired runs and intervals.
On TabFormer, with the training period its vendor states, the gap between the published model and our history-aware baseline remains large, and it narrows without closing when the baseline receives later data.
At the same time, a foundation model can add a measurable amount where the protocol leaves it room, as with the 255 examples of a new fraud regime on TabFormer or with 1,000 labels in the main study, which is consistent with the expected role of pretraining when labels are scarce.

\subsection{Recommendations}
\label{sec:disc-recommendations}

For practitioners evaluating a transaction foundation model, our results suggest starting from a history-aware, point-in-time boosting baseline, which in our runs closes most of the gap that weaker baselines leave, and checking, field by field, that no input is dated by when it refers to rather than by when it became known.
Candidates should be evaluated at several label budgets with paired seeds and intervals, reading an interval that covers zero as an absence of evidence rather than as a tie.
Fine-tuning on the target, alone or through joint fusion, should be evaluated as an alternative to frozen embeddings rather than assuming that frozen embeddings are enough, and combining the sequence model with the boosting model should be tried before replacing it.
Finally, when relying on a published comparison, it is worth checking what each side was allowed to see: until which date it was trained, which history its features use, how its threshold was chosen and what the test graph contains.

Two practical considerations make the combination easier to adopt than a replacement.
At the scale we tested, the compute cost of pretraining and fine-tuning is minutes of GPU time per dataset, so the cost of trying these models is mostly that of building the event pipeline, not of training.
And TBS preserves the explainability of the tabular component, which matters in credit, where models must be explained to risk committees and regulators.
The shift in the rank of each entity shows the effect of the network on the combination, but explaining why the network produced that score requires additional analysis.

\section{Threats to Validity}
\label{sec:threats}

We group the limitations of the study by the type of validity they affect.
Many of them were stated where they arise; this section collects them so that the conclusions of Sections~\ref{sec:results} and~\ref{sec:discussion} can be read with their limits.

\subsection{Internal validity}
\label{sec:threats-internal}

The transition features, the Mlm-Cl encoder and the two variants of TBS were designed after seeing earlier results on these same datasets, and the field encoder was selected on TabFormer before being run on the other datasets.
The hypotheses of Section~\ref{sec:introduction} were also formulated after those earlier results, not registered before the experiments.
The results of TBS and of its components are therefore not an independent confirmation, which requires a dataset or a period not used during their development.
The authors also have a commercial interest in TBS, which we disclose in the Declarations; to limit its influence, we report every method we ran, including those that did not help, and generate every number from the per-seed predictions.

No method in the main study received hyperparameter tuning, and a fixed configuration can favor one family over another (Section~\ref{sec:method-interpretation}); a tuned baseline or a tuned network could each move by more than some of the differences we report.
The additional TabFormer experiment of Section~\ref{sec:res-tabformer} is separate from the main study: it includes a hyperparameter search for CatBoost and a selection among 18 candidates.
The baseline uses generic window aggregates rather than features written by a specialist for each target; on TabFormer, that difference alone moves AUPRC from 0.605 to 0.809 in the calendar protocol, so a richer baseline could shrink the gains we report.
With 1,000 labels, the boosting model sees few positive examples, so part of the larger gain at small budgets may reflect a baseline that is weak in that regime.
Some comparisons change more than one component (Section~\ref{sec:models-ablations}), and we did not run the controls that would separate them further: joint fusion with a frozen encoder, an ensemble partner that ignores the order of events, such as an MLP on the aggregates or TabPFN, and JF-rand with more than 12 epochs, which could reduce the advantage of starting from a pretrained network.

\subsection{External validity}
\label{sec:threats-external}

Our sequence backbones have 0.4 million parameters per client and 8.6 million per transaction and are pretrained on tens of thousands of clients, while published production models are two to four orders of magnitude larger and are trained on millions of clients~\citep{braithwaite2026nuformer,ostroukhov2026pragma}.
Nor do we test broad pretraining and reuse, that is, a representation pretrained once on a large and varied corpus, inside one institution or across several, and reused across tasks: each of our backbones is pretrained on one dataset and serves only the targets of that dataset.
Our results therefore say nothing about what scale or reuse buys.

The datasets also limit how far the results travel.
We designed the synthetic dataset, and although its targets represent different sources of signal, varied mechanisms do not make it neutral between methods; it illustrates mechanisms, and its numbers are not market numbers.
MBD-mini is real, but its target is product propensity, its codes are anonymized and our sample is easier than population-level scoring.
The proprietary datasets cover two companies and use one row per entity rather than full panels, and they cannot be released, so their results cannot be reproduced by others.
We did not use free text or geolocation, which some production models include.

\subsection{Construct validity}
\label{sec:threats-construct}

In this study, ``foundation model'' refers to the pretraining and adaptation recipes of foundation models applied at a small scale and per dataset, not to a model pretrained once on a broad corpus.
Statements about foundation models in general should be read through this definition.
Similarly, the expert features represent the knowledge of one person about the synthetic generator and of one company about its own churn, not the best features that domain knowledge could produce.

The primary metric, the AUC, measures ranking quality, not calibration or operational value.
TBS produces a ranking that must be calibrated before it is used as a probability, and a gain of one AUC point may be worth a great deal or very little depending on the decision it supports, the cut-off and the cost of errors, none of which we evaluate.
The latent-information oracle of the synthetic data is a reference, not a ceiling, because part of the latent signal never reaches the transactions and because its empirical value can be exceeded by chance on a finite test set.

\subsection{Conclusion validity}
\label{sec:threats-conclusion}

Each cell of the main study is run with five seeds, and the bootstrap intervals hold the trained models fixed, so they measure the uncertainty of the finite test set and not the variation that new training samples or initializations would produce; the sensitivity analysis with resampled seeds is a coarse estimate of that variation.
Some tasks have few test positives, such as the 600 of client B, for which differences of about one AUC point are below what the test can detect.
The study makes many comparisons, which we control with the Benjamini--Hochberg procedure in two families, and cells are not independent, because the budgets of a target share its test set and the targets of a dataset share entities and pretraining.

The re-examination of the public benchmarks has weaker controls than the main study.
On TabFormer, we compared our models with the published scores of a commercial model whose stated training period we cannot verify from the scores, so the comparison is between our model and released scores on the same rows, not between two models trained under the same conditions.
On IBM AML, our runs of the Multi-GNN code use one seed each and the evidence for a gain from our sequence model depends on the seed, so those results indicate what the published comparison measures rather than a ranking of models.

\section{Conclusion}
\label{sec:conclusion}

This paper asked how much the recipes of foundation models for financial event sequences add to a baseline of point-in-time window aggregates and gradient boosting, and proposed TBS, a hybrid that combines the two.
On four datasets, at three label budgets and with paired seeds and intervals, frozen self-supervised representations rarely improved the baseline, and a fine-tuned sequence network alone cleared it in fewer than half of the target--budget cells.
TBS, which averages the ranks of a boosting model enriched with label-free transition features and of a pretrained network fine-tuned jointly with the aggregates, cleared the baseline in 19 of 21 cells with a median gain of +2.1 AUC points, and the controls suggest that most of this gain comes from combining boosting with a fine-tuned sequence network, with a smaller part from pretraining it.
The gains were largest with few labels and without rich tabular data, and shrank as either grew.

The re-examination of two public benchmarks showed that protocol choices, such as the training cut-off, the information given to the baseline, the choice of the decision threshold and the information visible at test time, can produce differences between models larger than those measured under a point-in-time protocol.
On IBM AML, a history-aware, point-in-time baseline reached the F1 of published graph neural networks.
On TabFormer, in the comparisons reported in Table~\ref{tab:res-tabformer}, with the training period the vendor states, the published model stayed ahead of our history-aware baseline and of our foundation model, and the baseline remained below it even when trained on later data that the vendor states its model did not use.

These conclusions hold for sequence models of up to a few million parameters, pretrained per dataset, evaluated at fixed configurations on four datasets, and the method we propose was developed on the same data on which it was evaluated.
Three directions follow from these limits.
The first is to tune the main competitors under the same search budget.
The second is to confirm TBS on datasets and periods not used during its development.
The third is to measure what scale and reuse add, by pretraining larger models on broad corpora and evaluating them under the same paired, label-budgeted and point-in-time protocol.

\section*{Declarations}
\phantomsection
\label{sec:declarations}
\addcontentsline{toc}{section}{Declarations}

Both authors work for WeCogno, which sells modeling services and offers TBS, together with the transition features and the Mlm-Cl encoder, commercially under the name WeCogno Temporal Booster.
WeCogno also competes with NeoSpace, whose published results we re-examine in Section~\ref{sec:res-tabformer}.
The proprietary datasets belong to clients of the company and are used with their permission.

The synthetic data generator, the computation of its latent-information oracle, the code of the paired statistics, and the per-seed test predictions and labels of the synthetic and MBD-mini datasets are available at \url{https://github.com/datarisk-io/fm-vs-gbdt-eval} under the Apache 2.0 license (Section~\ref{sec:setup-config}).
MBD-mini is publicly available under the CC BY 4.0 license~\citep{mollaev2025mbd}, and the TabFormer and IBM AML datasets are publicly available from their authors~\citep{padhi2021tabbert,altman2023aml}.
The proprietary datasets and the rest of the pipeline are not released.

\bibliographystyle{plainnat}
\bibliography{references}

\appendix
\section{Configurations}
\label{app:config}

No configuration below was tuned on the test set or searched on validation; all were fixed before the runs reported in this paper.
Table~\ref{tab:config} lists them for every component of the main study.

\begin{table}[ht]
  \centering
  \caption{Fixed configurations of the main study.}
  \label{tab:config}
  \footnotesize
  \begin{tabularx}{\linewidth}{>{\raggedright\arraybackslash}p{3.0cm}X}
    \toprule
    Component & Configuration \\
    \midrule
    LightGBM & binary objective, learning rate 0.05, 31 leaves, min.\ 40 rows per leaf, feature fraction 0.8, bagging fraction 0.8 every iteration, L2 1.0, deterministic, up to 3,000 rounds, early stopping after 100 rounds on a stratified 20\% split of the training rows \\
    Tokens & last 128 events; categorical vocabularies capped at the most frequent training-period levels (8--64 per field), amount in 32 quantile bands, gap 16, recency 16, hour 24; three standardized continuous values \\
    Backbone & width 96, 4 layers, 4 heads, feed-forward 256, dropout 0.1, pre-norm, GELU, learned positions; representation = last position $\oplus$ mean (192) \\
    Field encoder & per event: field embeddings and the three continuous values as tokens, field position embeddings, 2-layer Transformer (width 96, 4 heads, feed-forward 192), concatenation and linear projection to 96 \\
    Pretraining & AdamW, learning rate $10^{-3}$, weight decay 0.01, one-cycle schedule (5\% warm-up), batch 128, gradient clipping 1.0; MLM and Nep: 10 epochs (synthetic) or 8; CoLES: 12 or 10 epochs, 4 sub-sequences of 20--120 events per client, temperature 0.1; Mlm-Cl: 10 epochs, contrastive weight 0.5, temperature 0.1, projection to 64 \\
    Reconstruction AEs & inputs: last 50 events, amount and hour (plus one-hot event type and currency in the second variant); Transformer width 64, 4 heads, 2 layers, feed-forward 128, 15\% of events masked, 5 epochs; MLP autoencoders with layers 256--128 and latent dimension 64, 50 epochs \\
    GRU & hidden size 128 over the event embeddings \\
    Joint fusion & tabular branch 256--128 with GELU and dropout 0.2 on quantile-normalized aggregates; head 128--1; encoder learning rate $2 \times 10^{-4}$ (pretrained) or $10^{-3}$, other parameters $10^{-3}$; batch 128; up to 12 epochs, patience 2 on validation AUC (20\% stratified split) \\
    TBS-R & offset from three-fold out-of-fold LightGBM logits; head 192--64--1, last layer initialized at zero; encoder learning rate $10^{-4}$ ($5 \times 10^{-4}$ without pretraining), head $10^{-3}$; up to 8 epochs, patience 1; the initial state is kept unless validation AUC improves \\
    TabPFN v2 & released v2 weights, default classifier; at most 300 features, selected by LightGBM gain on the training rows \\
    Trans & $\Delta t \in \{1, 24\}$ hours, windows 30/90/180 days, minimum support 1\% of training-period clients, at most 150 pairs, 30 event types plus a pooled level \\
    \bottomrule
  \end{tabularx}
\end{table}

\section{Reproducibility of the benchmark re-examination}
\label{app:repro}

The Cortex scores on TabFormer are four Parquet files, \texttt{000.parquet} to \texttt{003.parquet},\footnote{\url{https://embeddings.neospace.ai/cortex_score/}} which the notebook \texttt{cortex\_score.ipynb} of the repository of NeoSpace downloads~\citep{neospace2026ldm}.
Their SHA-256 digests, identical on 29 October and 6 November 2026, were the following.

\begin{center}
  \footnotesize
  \begin{tabular}{ll}
    \texttt{000.parquet} & \texttt{7629ec7aaea9ba7f58abb872013dfdbd1b891064fa6f384093300a7ea18ecc22} \\
    \texttt{001.parquet} & \texttt{e71c73b97c22491390261ca8d1c6c44445f3a2db9f9f063319195e8bec41c34e} \\
    \texttt{002.parquet} & \texttt{0d1339c10354ec667fc9631bce859263a5f807069179159fcd9e767b8a21ceba} \\
    \texttt{003.parquet} & \texttt{72cb1ef2677a79b441518d8c89894cb5161fe50186e95758bd364858d892a9f8} \\
  \end{tabular}
\end{center}

The Multi-GNN runs use commit \texttt{252b0252afca109d1d216c411c59ff70753b25fc} of the repository of the authors (\url{https://github.com/IBM/Multi-GNN}), with their training loop, settings, split and loaders unchanged.
A wrapper replaces their evaluation functions by copies that return the same F1 and also save, for every scored edge, its laundering probability at the epoch with the best validation F1, which our rescoring reads.

\section{Cost at larger scale}
\label{app:cost}

At the rates measured in Section~\ref{sec:cost}, that is, 624 to 4,741 client-epochs per second in pretraining, 1.3 to 3.0\,ms per labeled client per fine-tuning run and 0.12 to 0.82\,ms per client to score, a bank with 100 million clients, half of them labeled, would need 2.8 to 17.5 days of A10G time per training round with our backbone, about US\$111 to 683.
The binding constraint at that scale is memory: the token tensors of 100 million clients take 1,001\,GiB of host memory, against 64\,GB on our machine, so data would have to be streamed from disk.

Our backbones are small, with 0.4 million parameters per client and 8.6 million for the card model, whereas published production models have hundreds of millions to billions of parameters~\citep{braithwaite2026nuformer,ostroukhov2026pragma}.
For an order of magnitude, we use the training-compute approximation of $6ND$ floating-point operations for $N$ parameters and $D$ tokens~\citep{kaplan2020scaling}, one token per field (9 fields per transaction) with up to 2,048 tokens per client, and one H100 at 40\% of its 989 dense bf16 TFLOPS, priced at one eighth of a p5.48xlarge (US\$6.88 per hour).
One pretraining epoch over the same 100 million clients with 120 transactions each would take 45.5\,h and cost US\$313 with 100 million parameters, 5.7 days and US\$939 with 300 million, and 19.0 days and US\$3,130 with one billion.
These figures ignore the cost of attention, a further 20 to 40\% at this context, so they are lower bounds.
Under these assumptions, a pretraining epoch at this scale would cost hundreds to a few thousand dollars.
The estimate is conditional on the utilization, the token count and the price we assumed, and what makes such projects expensive is the number of epochs and experimental runs, data engineering, real-time serving and the team, which this estimate does not include.

\section{Full results}
\label{app:results}

This appendix gives the differences against the baseline at 1,000 and 4,000 labels (Tables~\ref{tab:res-1k} and~\ref{tab:res-4k}), the differences in the full scenario (Table~\ref{tab:res-full}), the targeted comparisons and ensemble controls of Sections~\ref{sec:res-finetune} and~\ref{sec:res-tbs} by label budget (Tables~\ref{tab:res-targeted} and~\ref{tab:res-controls}), and the absolute test AUCs of every method, target, scenario and budget, grouped by dataset (Tables~\ref{tab:abs-first} onward).
Markers are as in Table~\ref{tab:res-all}.

\begin{table}[p]
  \centering
  \caption{As Table~\ref{tab:res-all}, with 1,000 training labels.}
  \label{tab:res-1k}
  \footnotesize
  \setlength{\tabcolsep}{3.5pt}
  \begin{tabular}{lrrrrrrr}
    \toprule
    & \multicolumn{3}{c}{Synthetic} & \multicolumn{2}{c}{MBD-mini} & \multicolumn{1}{c}{Client A} & \multicolumn{1}{c}{Client B} \\
    \cmidrule(lr){2-4} \cmidrule(lr){5-6} \cmidrule(lr){7-7} \cmidrule(lr){8-8}
    Method & Default & \shortstack{Seq.\\trigger} & Churn & \shortstack{Any\\product} & \shortstack{Product\\1} & \shortstack{Late\\payment} & \shortstack{Churn\\(6m)} \\
    \midrule
    Agg AUC & 0.621 & 0.703 & 0.598 & 0.678 & 0.656 & 0.820 & 0.575 \\
    \multicolumn{8}{l}{\textit{Frozen representation + GBDT}} \\
    AE, Transformer, masked numeric & $-$0.5 & +0.3 & +0.3 & +0.3 & $-$0.1 & +0.1 & +0.2 \\
    AE, Transformer, + one-hot & +0.0 & \textbf{+2.0$^{\uparrow*}$} & $-$0.3 & $-$0.7 & +0.7 & +0.6 & $-$0.4 \\
    AE, MLP, whole sequence & +0.3 & $-$0.2 & $-$0.4 & +0.3 & +0.2 & $-$0.2 & $-$0.5 \\
    AE, MLP, masked events & $-$0.9 & $-$0.7$^{\downarrow}$ & $-$0.2 & +0.1 & $-$1.0 & $-$0.4$^{\downarrow}$ & $-$0.1 \\
    MLM (TabBERT-style) & \textbf{+1.3$^{\uparrow*}$} & \textbf{+1.3$^{\uparrow*}$} & \textbf{+1.2$^{\uparrow}$} & $-$0.8 & $-$0.4 & +0.0 & $-$0.7 \\
    Nep (nuFormer-style) & \textbf{+1.5$^{\uparrow*}$} & +0.3 & +0.8 & +0.4 & $-$0.3 & +0.5 & +0.1 \\
    CoLES & +0.6 & \textbf{+4.4$^{\uparrow*}$} & $-$1.0 & $-$0.6 & $-$0.2 & \textbf{+0.9$^{\uparrow}$} & $-$0.2 \\
    Mlm-Cl & \textbf{+1.1$^{\uparrow}$} & +0.6 & \textbf{+1.1$^{\uparrow}$} & $-$0.8 & +0.0 & +0.6 & +0.9 \\
    \multicolumn{8}{l}{\textit{End-to-end}} \\
    GRU & $-$3.9$^{\downarrow*}$ & $-$16.1$^{\downarrow*}$ & +1.3 & $-$3.2$^{\downarrow*}$ & $-$5.2$^{\downarrow*}$ & +0.7 & $-$0.6 \\
    JF-rand (no pretraining) & +0.2 & $-$2.1$^{\downarrow*}$ & \textbf{+1.7$^{\uparrow*}$} & \textbf{+2.7$^{\uparrow*}$} & +2.2 & $-$0.8 & $-$1.8$^{\downarrow*}$ \\
    JF (pretrained Nep + joint fusion) & \textbf{+2.1$^{\uparrow*}$} & \textbf{+2.6$^{\uparrow*}$} & \textbf{+2.3$^{\uparrow*}$} & \textbf{+1.8$^{\uparrow*}$} & $-$2.1 & \textbf{+1.7$^{\uparrow*}$} & $-$1.0 \\
    TabPFN v2 & \textbf{+1.1$^{\uparrow*}$} & \textbf{+1.9$^{\uparrow*}$} & \textbf{+3.9$^{\uparrow*}$} & \textbf{+2.2$^{\uparrow*}$} & $-$1.2 & \textbf{+0.9$^{\uparrow*}$} & +0.8 \\
    \multicolumn{8}{l}{\textit{Expert features}} \\
    Expert features alone & $-$0.7 & \textbf{+5.2$^{\uparrow*}$} & \textbf{+3.3$^{\uparrow*}$} & -- & -- & -- & $-$0.5 \\
    Agg + expert features & +0.3 & \textbf{+5.5$^{\uparrow*}$} & \textbf{+2.4$^{\uparrow*}$} & -- & -- & -- & -- \\
    \multicolumn{8}{l}{\textit{Transition features}} \\
    Agg + Trans & $-$0.8$^{\downarrow}$ & \textbf{+5.4$^{\uparrow*}$} & +0.1 & +0.4 & +0.9 & +0.2 & \textbf{+0.7$^{\uparrow}$} \\
    Agg + Trans + Mlm-Cl & \textbf{+1.1$^{\uparrow}$} & \textbf{+4.5$^{\uparrow*}$} & \textbf{+1.2$^{\uparrow}$} & +0.2 & $-$1.0 & +0.6 & +1.1 \\
    \multicolumn{8}{l}{\textit{TBS and hybrids}} \\
    JF + Trans & \textbf{+2.1$^{\uparrow*}$} & \textbf{+3.1$^{\uparrow*}$} & \textbf{+2.0$^{\uparrow*}$} & \textbf{+1.5$^{\uparrow*}$} & \textbf{+2.4$^{\uparrow}$} & \textbf{+1.7$^{\uparrow*}$} & $-$0.5 \\
    TBS & \textbf{+2.1$^{\uparrow*}$} & \textbf{+5.2$^{\uparrow*}$} & \textbf{+2.4$^{\uparrow*}$} & \textbf{+2.4$^{\uparrow*}$} & \textbf{+2.3$^{\uparrow*}$} & \textbf{+1.7$^{\uparrow*}$} & +0.4 \\
    TBS-R & +0.3 & \textbf{+5.5$^{\uparrow*}$} & $-$0.2 & +0.5 & \textbf{+1.6$^{\uparrow*}$} & \textbf{+0.4$^{\uparrow}$} & \textbf{+1.5$^{\uparrow*}$} \\
    TBS-R, no pretraining & $-$2.2$^{\downarrow*}$ & \textbf{+5.6$^{\uparrow*}$} & $-$0.2 & \textbf{+0.7$^{\uparrow*}$} & +1.0 & +0.3 & +0.3 \\
    TBS-R, no Trans in offset & \textbf{+1.5$^{\uparrow*}$} & \textbf{+0.9$^{\uparrow*}$} & +0.1 & $-$0.6$^{\downarrow*}$ & \textbf{+1.1$^{\uparrow*}$} & \textbf{+0.4$^{\uparrow*}$} & \textbf{+1.1$^{\uparrow*}$} \\
    TBS-R, Nep encoder & +0.0 & \textbf{+5.6$^{\uparrow*}$} & \textbf{+1.3$^{\uparrow*}$} & \textbf{+1.0$^{\uparrow*}$} & \textbf{+1.4$^{\uparrow*}$} & +0.2 & \textbf{+1.2$^{\uparrow*}$} \\
    \multicolumn{8}{l}{\textit{Field encoder (second version)}} \\
    Nep, field encoder & \textbf{+1.6$^{\uparrow*}$} & \textbf{+1.1$^{\uparrow*}$} & +0.1 & $-$0.9 & $-$0.5 & +0.7 & $-$0.5 \\
    JF, field encoder & \textbf{+1.4$^{\uparrow}$} & \textbf{+3.6$^{\uparrow*}$} & \textbf{+2.3$^{\uparrow*}$} & \textbf{+1.1$^{\uparrow}$} & $-$0.1 & \textbf{+1.8$^{\uparrow*}$} & $-$0.6 \\
    JF + Trans, field encoder & \textbf{+2.5$^{\uparrow*}$} & \textbf{+4.1$^{\uparrow*}$} & \textbf{+1.7$^{\uparrow*}$} & \textbf{+1.4$^{\uparrow*}$} & \textbf{+4.7$^{\uparrow*}$} & \textbf{+1.7$^{\uparrow*}$} & +0.0 \\
    TBS, field encoder & \textbf{+1.8$^{\uparrow*}$} & \textbf{+5.5$^{\uparrow*}$} & \textbf{+2.4$^{\uparrow*}$} & \textbf{+2.1$^{\uparrow*}$} & \textbf{+3.1$^{\uparrow*}$} & \textbf{+1.8$^{\uparrow*}$} & +0.6 \\
 
    \bottomrule
  \end{tabular}
\end{table}

\begin{table}[p]
  \centering
  \caption{As Table~\ref{tab:res-all}, with 4,000 training labels.}
  \label{tab:res-4k}
  \footnotesize
  \setlength{\tabcolsep}{3.5pt}
  \begin{tabular}{lrrrrrrr}
    \toprule
    & \multicolumn{3}{c}{Synthetic} & \multicolumn{2}{c}{MBD-mini} & \multicolumn{1}{c}{Client A} & \multicolumn{1}{c}{Client B} \\
    \cmidrule(lr){2-4} \cmidrule(lr){5-6} \cmidrule(lr){7-7} \cmidrule(lr){8-8}
    Method & Default & \shortstack{Seq.\\trigger} & Churn & \shortstack{Any\\product} & \shortstack{Product\\1} & \shortstack{Late\\payment} & \shortstack{Churn\\(6m)} \\
    \midrule
    Agg AUC & 0.634 & 0.721 & 0.626 & 0.711 & 0.682 & 0.834 & 0.594 \\
    \multicolumn{8}{l}{\textit{Frozen representation + GBDT}} \\
    AE, Transformer, masked numeric & $-$0.1 & $-$0.2 & +0.1 & +0.1 & +1.1 & $-$0.1 & +0.3 \\
    AE, Transformer, + one-hot & +0.1 & +1.0 & $-$0.7 & +0.2 & +0.4 & +0.1 & +0.3 \\
    AE, MLP, whole sequence & +0.5 & $-$0.2 & $-$0.3 & +0.3 & +1.0 & +0.1 & +0.2 \\
    AE, MLP, masked events & +0.5 & $-$0.9$^{\downarrow*}$ & $-$0.3 & +0.0 & +0.6 & $-$0.3 & +0.0 \\
    MLM (TabBERT-style) & \textbf{+1.2$^{\uparrow*}$} & \textbf{+1.3$^{\uparrow}$} & +0.6 & +0.0 & +0.8 & $-$0.1 & +0.6 \\
    Nep (nuFormer-style) & \textbf{+1.2$^{\uparrow*}$} & +0.7 & +0.5 & +0.7 & +0.8 & +0.0 & +0.8 \\
    CoLES & +0.7 & \textbf{+3.5$^{\uparrow*}$} & $-$0.4 & +0.0 & $-$0.1 & +0.5 & \textbf{+1.4$^{\uparrow*}$} \\
    Mlm-Cl & \textbf{+1.1$^{\uparrow}$} & \textbf{+0.9$^{\uparrow}$} & +0.9 & +0.5 & +0.2 & +0.0 & \textbf{+1.6$^{\uparrow*}$} \\
    \multicolumn{8}{l}{\textit{End-to-end}} \\
    GRU & $-$0.8 & \textbf{+3.8$^{\uparrow*}$} & +0.4 & $-$0.8 & $-$0.5 & $-$0.1 & +0.9 \\
    JF-rand (no pretraining) & +0.4 & $-$0.3 & $-$0.8 & \textbf{+1.4$^{\uparrow*}$} & \textbf{+2.4$^{\uparrow}$} & $-$0.6 & $-$1.6$^{\downarrow*}$ \\
    JF (pretrained Nep + joint fusion) & \textbf{+2.0$^{\uparrow*}$} & \textbf{+3.6$^{\uparrow*}$} & +0.6 & \textbf{+1.6$^{\uparrow*}$} & \textbf{+2.0$^{\uparrow}$} & +0.2 & +0.5 \\
    TabPFN v2 & \textbf{+2.4$^{\uparrow*}$} & \textbf{+1.8$^{\uparrow*}$} & \textbf{+2.8$^{\uparrow*}$} & \textbf{+2.1$^{\uparrow*}$} & \textbf{+3.4$^{\uparrow*}$} & +0.3 & +0.9 \\
    \multicolumn{8}{l}{\textit{Expert features}} \\
    Expert features alone & +0.7 & \textbf{+3.9$^{\uparrow*}$} & \textbf{+1.5$^{\uparrow*}$} & -- & -- & -- & $-$0.3 \\
    Agg + expert features & $-$0.7 & \textbf{+3.8$^{\uparrow*}$} & \textbf{+1.9$^{\uparrow*}$} & -- & -- & -- & -- \\
    \multicolumn{8}{l}{\textit{Transition features}} \\
    Agg + Trans & +0.3 & \textbf{+3.6$^{\uparrow*}$} & +0.0 & +0.2 & \textbf{+1.8$^{\uparrow*}$} & +0.0 & \textbf{+0.7$^{\uparrow}$} \\
    Agg + Trans + Mlm-Cl & +0.7 & \textbf{+3.0$^{\uparrow*}$} & +0.2 & \textbf{+1.2$^{\uparrow*}$} & +1.0 & $-$0.1 & \textbf{+2.0$^{\uparrow*}$} \\
    \multicolumn{8}{l}{\textit{TBS and hybrids}} \\
    JF + Trans & \textbf{+2.0$^{\uparrow*}$} & \textbf{+4.2$^{\uparrow*}$} & +0.2 & +1.0 & +1.7 & +0.1 & +0.2 \\
    TBS & \textbf{+2.2$^{\uparrow*}$} & \textbf{+4.1$^{\uparrow*}$} & \textbf{+1.0$^{\uparrow}$} & \textbf{+2.2$^{\uparrow*}$} & \textbf{+3.5$^{\uparrow*}$} & \textbf{+0.8$^{\uparrow*}$} & \textbf{+1.3$^{\uparrow*}$} \\
    TBS-R & \textbf{+1.4$^{\uparrow*}$} & \textbf{+3.8$^{\uparrow*}$} & $-$0.3 & \textbf{+1.2$^{\uparrow*}$} & \textbf{+2.5$^{\uparrow*}$} & +0.4 & \textbf{+1.4$^{\uparrow*}$} \\
    TBS-R, no pretraining & +0.1 & \textbf{+3.5$^{\uparrow*}$} & $-$0.3 & \textbf{+1.0$^{\uparrow*}$} & \textbf{+2.7$^{\uparrow*}$} & +0.3 & +0.4 \\
    TBS-R, no Trans in offset & \textbf{+1.8$^{\uparrow*}$} & \textbf{+4.1$^{\uparrow*}$} & $-$0.5 & \textbf{+1.1$^{\uparrow*}$} & +0.2 & \textbf{+0.4$^{\uparrow}$} & \textbf{+1.4$^{\uparrow*}$} \\
    TBS-R, Nep encoder & \textbf{+1.1$^{\uparrow*}$} & \textbf{+3.6$^{\uparrow*}$} & +0.7 & \textbf{+1.1$^{\uparrow*}$} & \textbf{+2.9$^{\uparrow*}$} & \textbf{+0.9$^{\uparrow*}$} & \textbf{+1.3$^{\uparrow*}$} \\
    \multicolumn{8}{l}{\textit{Field encoder (second version)}} \\
    Nep, field encoder & +0.4 & +0.6 & +0.6 & \textbf{+0.9$^{\uparrow}$} & \textbf{+1.9$^{\uparrow}$} & +0.1 & +0.3 \\
    JF, field encoder & \textbf{+2.2$^{\uparrow*}$} & \textbf{+4.6$^{\uparrow*}$} & \textbf{+1.7$^{\uparrow*}$} & \textbf{+1.3$^{\uparrow*}$} & \textbf{+2.3$^{\uparrow}$} & +0.8 & +0.1 \\
    JF + Trans, field encoder & \textbf{+2.1$^{\uparrow*}$} & \textbf{+4.4$^{\uparrow*}$} & +0.8 & \textbf{+2.1$^{\uparrow*}$} & \textbf{+3.2$^{\uparrow*}$} & +0.8 & $-$0.2 \\
    TBS, field encoder & \textbf{+2.3$^{\uparrow*}$} & \textbf{+4.5$^{\uparrow*}$} & \textbf{+1.6$^{\uparrow*}$} & \textbf{+2.2$^{\uparrow*}$} & \textbf{+3.5$^{\uparrow*}$} & \textbf{+1.0$^{\uparrow*}$} & \textbf{+1.2$^{\uparrow*}$} \\
 
    \bottomrule
  \end{tabular}
\end{table}

\begin{table}[p]
  \centering
  \caption{As Table~\ref{tab:res-all}, in the full scenario: register and statement aggregates (synthetic), static attributes of the point of sale (client A) or the twelve variables of the production model (client B) are added to every method.}
  \label{tab:res-full}
  \footnotesize
  \begin{tabular}{lrrrrr}
    \toprule
    & \multicolumn{3}{c}{Synthetic} & \multicolumn{1}{c}{Client A} & \multicolumn{1}{c}{Client B} \\
    \cmidrule(lr){2-4} \cmidrule(lr){5-5} \cmidrule(lr){6-6}
    Method & Default & Seq.\ trigger & Churn & Late payment & Churn (6m) \\
    \midrule
    Agg AUC & 0.741 & 0.732 & 0.649 & 0.848 & 0.622 \\
    AE, Transformer, masked numeric & $-$0.1 & +0.3 & $-$0.1 & +0.0 & +0.0 \\
    AE, Transformer, + one-hot & $-$0.3 & \textbf{+1.3$^{\uparrow*}$} & +0.0 & $-$0.1 & +0.0 \\
    AE, MLP, whole sequence & $-$0.2 & +0.1 & $-$0.1 & $-$0.3 & +0.0 \\
    AE, MLP, masked events & $-$0.1 & $-$0.2 & +0.1 & $-$0.1 & $-$0.3 \\
    MLM (TabBERT-style) & $-$0.2 & \textbf{+1.5$^{\uparrow*}$} & +0.1 & $-$0.5 & +0.1 \\
    Nep (nuFormer-style) & $-$0.4 & \textbf{+1.4$^{\uparrow*}$} & +0.0 & +0.2 & +0.6 \\
    CoLES & $-$0.3 & \textbf{+3.4$^{\uparrow*}$} & $-$0.2 & +0.0 & +1.0 \\
    Mlm-Cl & $-$0.1 & \textbf{+1.7$^{\uparrow*}$} & $-$0.1 & +0.0 & +0.7 \\
    JF-rand (no pretraining) & $-$0.3 & \textbf{+2.6$^{\uparrow*}$} & $-$0.1 & +0.1 & $-$0.7 \\
    JF (pretrained Nep + joint fusion) & $-$0.4 & \textbf{+3.9$^{\uparrow*}$} & $-$0.4 & +0.6 & $-$0.1 \\
    Expert features alone & $-$0.1 & \textbf{+3.4$^{\uparrow*}$} & +0.1 & -- & -- \\
    Agg + expert features & $-$0.3 & \textbf{+3.8$^{\uparrow*}$} & +0.4 & -- & -- \\
    Agg + Trans & $-$0.1 & \textbf{+3.5$^{\uparrow*}$} & +0.4 & $-$0.2 & +0.2 \\
    Agg + Trans + Mlm-Cl & $-$0.3 & \textbf{+3.1$^{\uparrow*}$} & +0.1 & $-$0.1 & +1.0 \\
    JF + Trans & $-$0.8 & \textbf{+3.3$^{\uparrow*}$} & +0.2 & +0.5 & $-$0.2 \\
    TBS & +0.2 & \textbf{+4.1$^{\uparrow*}$} & +0.5 & \textbf{+0.7$^{\uparrow}$} & +0.8 \\
    TBS-R & $-$0.1 & \textbf{+3.7$^{\uparrow*}$} & +0.2 & $-$0.4 & \textbf{+0.9$^{\uparrow}$} \\
    TBS-R, no pretraining & $-$0.1 & \textbf{+3.4$^{\uparrow*}$} & +0.4 & $-$0.3 & +0.2 \\
    TBS-R, no Trans in offset & +0.0 & \textbf{+3.0$^{\uparrow*}$} & $-$0.2 & +0.0 & \textbf{+0.8$^{\uparrow*}$} \\
    TBS-R, Nep encoder & $-$0.1 & \textbf{+3.7$^{\uparrow*}$} & +0.5 & $-$0.1 & +0.6 \\
    Nep, field encoder & $-$0.5$^{\downarrow}$ & \textbf{+1.5$^{\uparrow*}$} & $-$0.4 & +0.1 & +0.3 \\
    JF, field encoder & $-$0.5 & \textbf{+3.6$^{\uparrow*}$} & +0.1 & +0.7 & $-$0.1 \\
    JF + Trans, field encoder & $-$1.1$^{\downarrow}$ & \textbf{+3.8$^{\uparrow*}$} & +0.0 & +0.6 & +0.3 \\
    TBS, field encoder & +0.2 & \textbf{+3.9$^{\uparrow*}$} & +0.7 & \textbf{+0.7$^{\uparrow}$} & \textbf{+0.9$^{\uparrow}$} \\
 
    \bottomrule
  \end{tabular}
\end{table}

\begin{table}[p]
  \centering
  \caption{Targeted comparisons in the transactions-only scenario, in AUC points, by label budget. Markers as in Table~\ref{tab:res-all}, except that $*$ marks Benjamini--Hochberg $q < 0.05$ within the family of targeted comparisons (Section~\ref{sec:method-stats}).}
  \label{tab:res-targeted}
  \footnotesize
  \setlength{\tabcolsep}{3pt}
  \begin{tabular}{llrrrrrrr}
    \toprule
    Comparison & Labels & Default & \shortstack{Seq.\\trigger} & Churn & \shortstack{Any\\product} & \shortstack{Product\\1} & \shortstack{Late\\payment} & \shortstack{Churn\\(6m)} \\
    \midrule
    \multirow{3}{*}{\parbox{3.7cm}{\raggedright JF vs.\ JF-rand (pretraining)}} & 1k & \textbf{+1.9$^{\uparrow*}$} & \textbf{+4.6$^{\uparrow*}$} & +0.6 & $-$0.9 & $-$4.3$^{\downarrow*}$ & \textbf{+2.5$^{\uparrow*}$} & \textbf{+0.9$^{\uparrow}$} \\
     & 4k & \textbf{+1.6$^{\uparrow*}$} & \textbf{+3.9$^{\uparrow*}$} & \textbf{+1.4$^{\uparrow*}$} & +0.2 & $-$0.3 & \textbf{+0.8$^{\uparrow*}$} & \textbf{+2.1$^{\uparrow*}$} \\
     & all & \textbf{+0.9$^{\uparrow}$} & +0.9 & +0.4 & $-$0.3 & +0.5 & +0.5 & +0.3 \\
    \midrule
    \multirow{3}{*}{\parbox{3.7cm}{\raggedright JF vs.\ frozen Nep in boosting}} & 1k & +0.5 & \textbf{+2.2$^{\uparrow*}$} & \textbf{+1.6$^{\uparrow}$} & \textbf{+1.4$^{\uparrow*}$} & $-$1.8 & \textbf{+1.2$^{\uparrow*}$} & $-$1.0 \\
     & 4k & +0.8 & \textbf{+2.9$^{\uparrow*}$} & +0.1 & +1.0 & +1.2 & +0.2 & $-$0.3 \\
     & all & +0.3 & \textbf{+1.6$^{\uparrow*}$} & $-$0.5 & +0.0 & +0.6 & +0.3 & $-$0.5 \\
    \midrule
    \multirow{3}{*}{\parbox{3.7cm}{\raggedright TBS vs.\ Agg + Trans}} & 1k & \textbf{+2.9$^{\uparrow*}$} & $-$0.2 & \textbf{+2.4$^{\uparrow*}$} & \textbf{+2.0$^{\uparrow*}$} & +1.4 & \textbf{+1.5$^{\uparrow*}$} & $-$0.2 \\
     & 4k & \textbf{+1.9$^{\uparrow*}$} & +0.5 & \textbf{+1.0$^{\uparrow*}$} & \textbf{+2.0$^{\uparrow*}$} & \textbf{+1.7$^{\uparrow*}$} & \textbf{+0.8$^{\uparrow*}$} & +0.6 \\
     & all & \textbf{+1.2$^{\uparrow*}$} & +0.3 & +0.4 & \textbf{+1.2$^{\uparrow*}$} & \textbf{+2.4$^{\uparrow*}$} & \textbf{+0.6$^{\uparrow*}$} & +0.5 \\
    \midrule
    \multirow{3}{*}{\parbox{3.7cm}{\raggedright TBS vs.\ JF}} & 1k & +0.1 & \textbf{+2.7$^{\uparrow*}$} & +0.1 & \textbf{+0.6$^{\uparrow}$} & \textbf{+4.4$^{\uparrow*}$} & +0.1 & \textbf{+1.4$^{\uparrow*}$} \\
     & 4k & +0.2 & +0.5 & +0.4 & +0.6 & \textbf{+1.5$^{\uparrow*}$} & \textbf{+0.6$^{\uparrow*}$} & \textbf{+0.8$^{\uparrow*}$} \\
     & all & +0.3 & $-$0.1 & +0.4 & +0.6 & $-$0.1 & +0.3 & \textbf{+0.8$^{\uparrow*}$} \\
    \midrule
    \multirow{3}{*}{\parbox{3.7cm}{\raggedright TBS-R vs.\ Agg + Trans}} & 1k & \textbf{+1.1$^{\uparrow*}$} & +0.1 & $-$0.3 & +0.1 & \textbf{+0.7$^{\uparrow*}$} & \textbf{+0.2$^{\uparrow}$} & \textbf{+0.8$^{\uparrow*}$} \\
     & 4k & \textbf{+1.0$^{\uparrow*}$} & +0.2 & $-$0.3 & \textbf{+0.9$^{\uparrow*}$} & +0.7 & \textbf{+0.4$^{\uparrow}$} & \textbf{+0.7$^{\uparrow*}$} \\
     & all & \textbf{+1.1$^{\uparrow*}$} & +0.3 & $-$0.3$^{\downarrow}$ & \textbf{+1.5$^{\uparrow*}$} & \textbf{+1.7$^{\uparrow*}$} & +0.0 & \textbf{+0.7$^{\uparrow*}$} \\
    \midrule
    \multirow{3}{*}{\parbox{3.7cm}{\raggedright TBS-R vs.\ TBS-R without pretraining}} & 1k & \textbf{+2.5$^{\uparrow*}$} & $-$0.1 & $-$0.1 & $-$0.2 & \textbf{+0.6$^{\uparrow}$} & +0.1 & \textbf{+1.2$^{\uparrow*}$} \\
     & 4k & \textbf{+1.2$^{\uparrow*}$} & +0.3 & +0.0 & +0.2 & $-$0.2 & +0.1 & \textbf{+0.9$^{\uparrow*}$} \\
     & all & \textbf{+1.0$^{\uparrow*}$} & +0.4 & $-$0.2$^{\downarrow}$ & \textbf{+0.6$^{\uparrow*}$} & +0.4 & +0.0 & \textbf{+0.7$^{\uparrow*}$} \\
    \midrule
    \multirow{3}{*}{\parbox{3.7cm}{\raggedright JF, field encoder vs.\ sum}} & 1k & $-$0.7$^{\downarrow*}$ & \textbf{+1.0$^{\uparrow*}$} & +0.0 & $-$0.7 & \textbf{+2.0$^{\uparrow}$} & +0.1 & +0.4 \\
     & 4k & +0.2 & \textbf{+1.0$^{\uparrow*}$} & +1.1 & $-$0.3 & +0.3 & \textbf{+0.6$^{\uparrow*}$} & $-$0.4 \\
     & all & +0.4 & +0.1 & $-$0.4 & +0.0 & $-$0.2 & +0.4 & \textbf{+0.9$^{\uparrow}$} \\
    \midrule
    \multirow{3}{*}{\parbox{3.7cm}{\raggedright TBS, field encoder vs.\ sum}} & 1k & $-$0.3$^{\downarrow*}$ & \textbf{+0.3$^{\uparrow}$} & +0.0 & $-$0.3 & +0.8 & +0.1 & +0.2 \\
     & 4k & +0.0 & \textbf{+0.4$^{\uparrow}$} & +0.6 & +0.0 & +0.0 & +0.2 & $-$0.1 \\
     & all & +0.2 & +0.1 & $-$0.2 & +0.0 & $-$0.1 & +0.1 & \textbf{+0.5$^{\uparrow}$} \\
 
    \bottomrule
  \end{tabular}
\end{table}

\begin{table}[p]
  \centering
  \caption{Ensemble controls, transactions-only scenario, in AUC points, by label budget. Each control is a rank average of two saved predictions, built like TBS (Agg + Trans with JF). Markers as in Table~\ref{tab:res-targeted}.}
  \label{tab:res-controls}
  \footnotesize
  \setlength{\tabcolsep}{3pt}
  \begin{tabular}{llrrrrrrr}
    \toprule
    Comparison & Labels & Default & \shortstack{Seq.\\trigger} & Churn & \shortstack{Any\\product} & \shortstack{Product\\1} & \shortstack{Late\\payment} & \shortstack{Churn\\(6m)} \\
    \midrule
    \multirow{3}{*}{\parbox{3.7cm}{\raggedright Ensemble of Agg and Agg + Trans vs.\ Agg}} & 1k & +0.2 & \textbf{+4.0$^{\uparrow*}$} & +0.3 & \textbf{+0.7$^{\uparrow*}$} & \textbf{+1.2$^{\uparrow*}$} & \textbf{+0.4$^{\uparrow*}$} & \textbf{+0.6$^{\uparrow*}$} \\
     & 4k & \textbf{+0.7$^{\uparrow*}$} & \textbf{+3.2$^{\uparrow*}$} & +0.3 & \textbf{+0.6$^{\uparrow*}$} & \textbf{+1.7$^{\uparrow*}$} & \textbf{+0.3$^{\uparrow*}$} & \textbf{+0.7$^{\uparrow*}$} \\
     & all & \textbf{+0.3$^{\uparrow}$} & \textbf{+2.3$^{\uparrow*}$} & +0.2 & \textbf{+0.4$^{\uparrow*}$} & +0.5 & \textbf{+0.3$^{\uparrow*}$} & \textbf{+0.8$^{\uparrow*}$} \\
    \midrule
    \multirow{3}{*}{\parbox{3.7cm}{\raggedright Ensemble with JF-rand vs.\ Agg + Trans}} & 1k & \textbf{+2.1$^{\uparrow*}$} & $-$1.6$^{\downarrow*}$ & \textbf{+2.0$^{\uparrow*}$} & \textbf{+2.2$^{\uparrow*}$} & \textbf{+2.9$^{\uparrow*}$} & +0.3 & $-$0.6 \\
     & 4k & \textbf{+1.1$^{\uparrow*}$} & $-$0.2 & +0.5 & \textbf{+1.8$^{\uparrow*}$} & \textbf{+1.7$^{\uparrow*}$} & \textbf{+0.5$^{\uparrow}$} & $-$0.5 \\
     & all & \textbf{+0.7$^{\uparrow}$} & +0.3 & +0.2 & \textbf{+1.1$^{\uparrow*}$} & \textbf{+2.0$^{\uparrow*}$} & +0.3 & +0.3 \\
    \midrule
    \multirow{3}{*}{\parbox{3.7cm}{\raggedright TBS vs.\ ensemble with JF-rand (pretraining)}} & 1k & \textbf{+0.8$^{\uparrow*}$} & \textbf{+1.3$^{\uparrow*}$} & +0.4 & $-$0.2 & $-$1.5$^{\downarrow*}$ & \textbf{+1.2$^{\uparrow*}$} & +0.4 \\
     & 4k & \textbf{+0.8$^{\uparrow*}$} & \textbf{+0.7$^{\uparrow}$} & +0.5 & +0.2 & +0.0 & +0.3 & \textbf{+1.1$^{\uparrow*}$} \\
     & all & \textbf{+0.5$^{\uparrow}$} & $-$0.1 & +0.2 & +0.0 & +0.3 & \textbf{+0.3$^{\uparrow}$} & +0.2 \\
    \midrule
    \multirow{3}{*}{\parbox{3.7cm}{\raggedright TBS vs.\ ensemble of Agg and JF (transitions)}} & 1k & $-$0.2 & \textbf{+2.5$^{\uparrow*}$} & +0.1 & +0.2 & +0.2 & +0.1 & \textbf{+0.3$^{\uparrow}$} \\
     & 4k & +0.2 & \textbf{+0.9$^{\uparrow}$} & +0.0 & +0.1 & +0.7 & +0.0 & +0.3 \\
     & all & +0.0 & +0.1 & $-$0.1 & +0.1 & $-$0.1 & +0.1 & \textbf{+0.5$^{\uparrow*}$} \\
    \midrule
    \multirow{3}{*}{\parbox{3.7cm}{\raggedright TBS vs.\ ensemble of Agg and Agg + Trans (network)}} & 1k & \textbf{+1.9$^{\uparrow*}$} & \textbf{+1.2$^{\uparrow*}$} & \textbf{+2.1$^{\uparrow*}$} & \textbf{+1.7$^{\uparrow*}$} & +1.1 & \textbf{+1.3$^{\uparrow*}$} & $-$0.1 \\
     & 4k & \textbf{+1.5$^{\uparrow*}$} & \textbf{+0.9$^{\uparrow*}$} & +0.7 & \textbf{+1.6$^{\uparrow*}$} & \textbf{+1.8$^{\uparrow*}$} & \textbf{+0.5$^{\uparrow}$} & +0.6 \\
     & all & \textbf{+0.8$^{\uparrow*}$} & +0.3 & +0.0 & \textbf{+1.0$^{\uparrow*}$} & \textbf{+1.6$^{\uparrow*}$} & +0.5 & \textbf{+0.8$^{\uparrow}$} \\
 
    \bottomrule
  \end{tabular}
\end{table}

\begin{table}[p]
  \centering
  \caption{Test AUC, mean over seeds, with the standard deviation in thousandths in parentheses: Synthetic, transactions-only scenario. Columns give the label budget (1k, 4k, all).}
  \label{tab:abs-first}
  \scriptsize
  \setlength{\tabcolsep}{2.2pt}
  \resizebox{\linewidth}{!}{%
  \begin{tabular}{lrrrrrrrrr}
    \toprule
    & \multicolumn{3}{c}{Default} & \multicolumn{3}{c}{Seq.\ trigger} & \multicolumn{3}{c}{Churn} \\
    \cmidrule(lr){2-4} \cmidrule(lr){5-7} \cmidrule(lr){8-10}
    Method & 1k & 4k & all & 1k & 4k & all & 1k & 4k & all \\
    \midrule
    Agg (baseline) & 0.621\,(8) & 0.634\,(7) & 0.654\,(6) & 0.703\,(7) & 0.721\,(5) & 0.733\,(3) & 0.598\,(20) & 0.626\,(8) & 0.653\,(4) \\
    AE, Transformer, masked numeric & 0.615\,(14) & 0.633\,(10) & 0.660\,(6) & 0.706\,(5) & 0.718\,(7) & 0.733\,(2) & 0.601\,(18) & 0.627\,(13) & 0.650\,(5) \\
    AE, Transformer, + one-hot & 0.620\,(13) & 0.636\,(8) & 0.657\,(7) & 0.723\,(8) & 0.730\,(7) & 0.741\,(4) & 0.595\,(22) & 0.619\,(10) & 0.649\,(5) \\
    AE, MLP, whole sequence & 0.624\,(10) & 0.639\,(9) & 0.652\,(7) & 0.701\,(12) & 0.719\,(7) & 0.731\,(4) & 0.594\,(18) & 0.623\,(12) & 0.648\,(10) \\
    AE, MLP, masked events & 0.612\,(22) & 0.639\,(3) & 0.657\,(6) & 0.696\,(5) & 0.711\,(7) & 0.726\,(6) & 0.596\,(17) & 0.623\,(10) & 0.651\,(10) \\
    MLM (TabBERT-style) & 0.633\,(10) & 0.646\,(9) & 0.660\,(4) & 0.716\,(9) & 0.733\,(8) & 0.748\,(4) & 0.610\,(22) & 0.632\,(14) & 0.653\,(6) \\
    Nep (nuFormer-style) & 0.636\,(7) & 0.646\,(7) & 0.659\,(3) & 0.706\,(8) & 0.728\,(6) & 0.744\,(7) & 0.605\,(17) & 0.631\,(12) & 0.657\,(7) \\
    CoLES & 0.626\,(5) & 0.641\,(7) & 0.661\,(3) & 0.747\,(6) & 0.756\,(6) & 0.767\,(3) & 0.588\,(19) & 0.622\,(8) & 0.646\,(6) \\
    Mlm-Cl & 0.632\,(7) & 0.645\,(10) & 0.665\,(2) & 0.709\,(10) & 0.730\,(5) & 0.746\,(3) & 0.609\,(10) & 0.635\,(9) & 0.651\,(5) \\
    GRU & 0.581\,(26) & 0.626\,(24) & 0.662\,(6) & 0.542\,(79) & 0.759\,(5) & 0.756\,(7) & 0.611\,(33) & 0.630\,(12) & 0.654\,(7) \\
    JF-rand (no pretraining) & 0.623\,(9) & 0.638\,(9) & 0.654\,(4) & 0.682\,(23) & 0.718\,(12) & 0.751\,(12) & 0.615\,(17) & 0.617\,(22) & 0.647\,(5) \\
    JF & 0.641\,(9) & 0.654\,(8) & 0.662\,(5) & 0.729\,(35) & 0.757\,(16) & 0.760\,(7) & 0.621\,(8) & 0.632\,(15) & 0.651\,(5) \\
    TabPFN v2 & 0.632\,(14) & 0.658\,(6) & -- & 0.722\,(6) & 0.739\,(5) & -- & 0.636\,(21) & 0.654\,(5) & -- \\
    Expert features alone & 0.614\,(19) & 0.641\,(8) & 0.659\,(5) & 0.755\,(5) & 0.760\,(5) & 0.769\,(4) & 0.631\,(18) & 0.641\,(6) & 0.653\,(5) \\
    Agg + expert features & 0.623\,(15) & 0.628\,(19) & 0.657\,(3) & 0.758\,(3) & 0.758\,(9) & 0.765\,(5) & 0.622\,(20) & 0.644\,(6) & 0.658\,(8) \\
    Agg + Trans & 0.613\,(16) & 0.638\,(10) & 0.654\,(4) & 0.757\,(9) & 0.757\,(7) & 0.756\,(4) & 0.598\,(18) & 0.626\,(8) & 0.651\,(4) \\
    Agg + Trans + Mlm-Cl & 0.632\,(7) & 0.642\,(8) & 0.665\,(3) & 0.748\,(4) & 0.751\,(7) & 0.755\,(6) & 0.610\,(17) & 0.628\,(14) & 0.653\,(4) \\
    JF + Trans & 0.642\,(17) & 0.655\,(8) & 0.660\,(4) & 0.734\,(53) & 0.763\,(7) & 0.762\,(5) & 0.618\,(8) & 0.628\,(17) & 0.649\,(6) \\
    TBS & 0.642\,(6) & 0.656\,(6) & 0.665\,(2) & 0.755\,(11) & 0.762\,(11) & 0.759\,(5) & 0.622\,(8) & 0.636\,(10) & 0.655\,(4) \\
    TBS-R & 0.623\,(27) & 0.648\,(13) & 0.665\,(3) & 0.758\,(9) & 0.759\,(8) & 0.759\,(5) & 0.596\,(19) & 0.622\,(11) & 0.648\,(5) \\
    TBS-R, no pretraining & 0.599\,(25) & 0.636\,(12) & 0.655\,(4) & 0.759\,(9) & 0.755\,(9) & 0.755\,(5) & 0.596\,(14) & 0.622\,(11) & 0.651\,(4) \\
    TBS-R, no Trans in offset & 0.636\,(15) & 0.653\,(8) & 0.666\,(7) & 0.712\,(18) & 0.762\,(4) & 0.761\,(4) & 0.599\,(19) & 0.621\,(13) & 0.651\,(4) \\
    TBS-R, Nep encoder & 0.621\,(21) & 0.645\,(14) & 0.663\,(5) & 0.759\,(9) & 0.757\,(7) & 0.760\,(4) & 0.611\,(13) & 0.633\,(8) & 0.649\,(6) \\
    Nep, field encoder & 0.637\,(12) & 0.638\,(17) & 0.659\,(3) & 0.714\,(12) & 0.727\,(4) & 0.746\,(5) & 0.599\,(19) & 0.632\,(11) & 0.645\,(7) \\
    JF, field encoder & 0.635\,(9) & 0.656\,(11) & 0.667\,(3) & 0.739\,(23) & 0.767\,(8) & 0.761\,(6) & 0.621\,(16) & 0.643\,(3) & 0.647\,(2) \\
    JF + Trans, field encoder & 0.646\,(9) & 0.655\,(10) & 0.662\,(4) & 0.744\,(36) & 0.765\,(13) & 0.759\,(7) & 0.615\,(21) & 0.633\,(15) & 0.648\,(9) \\
    TBS, field encoder & 0.638\,(6) & 0.657\,(7) & 0.667\,(3) & 0.758\,(5) & 0.766\,(8) & 0.760\,(3) & 0.622\,(15) & 0.642\,(3) & 0.653\,(3) \\
    \bottomrule
  \end{tabular}}
\end{table}

\begin{table}[p]
  \centering
  \caption{Test AUC, mean over seeds, with the standard deviation in thousandths in parentheses: Synthetic, full scenario. Columns give the label budget (1k, 4k, all).}
  \scriptsize
  \setlength{\tabcolsep}{2.2pt}
  \resizebox{\linewidth}{!}{%
  \begin{tabular}{lrrrrrrrrr}
    \toprule
    & \multicolumn{3}{c}{Default} & \multicolumn{3}{c}{Seq.\ trigger} & \multicolumn{3}{c}{Churn} \\
    \cmidrule(lr){2-4} \cmidrule(lr){5-7} \cmidrule(lr){8-10}
    Method & 1k & 4k & all & 1k & 4k & all & 1k & 4k & all \\
    \midrule
    Agg (baseline) & 0.702\,(11) & 0.722\,(5) & 0.741\,(3) & 0.699\,(7) & 0.718\,(5) & 0.732\,(4) & 0.602\,(14) & 0.626\,(11) & 0.649\,(4) \\
    AE, Transformer, masked numeric & 0.701\,(16) & 0.721\,(5) & 0.741\,(2) & 0.708\,(7) & 0.720\,(6) & 0.734\,(2) & 0.597\,(17) & 0.624\,(11) & 0.648\,(8) \\
    AE, Transformer, + one-hot & 0.702\,(11) & 0.721\,(3) & 0.739\,(3) & 0.724\,(8) & 0.729\,(8) & 0.745\,(4) & 0.591\,(17) & 0.627\,(16) & 0.649\,(7) \\
    AE, MLP, whole sequence & 0.703\,(13) & 0.720\,(4) & 0.739\,(4) & 0.703\,(11) & 0.714\,(5) & 0.733\,(3) & 0.586\,(19) & 0.625\,(8) & 0.648\,(8) \\
    AE, MLP, masked events & 0.701\,(12) & 0.722\,(7) & 0.741\,(1) & 0.704\,(7) & 0.714\,(7) & 0.730\,(5) & 0.596\,(17) & 0.626\,(14) & 0.650\,(4) \\
    MLM (TabBERT-style) & 0.696\,(10) & 0.722\,(5) & 0.739\,(3) & 0.718\,(5) & 0.727\,(6) & 0.747\,(3) & 0.605\,(20) & 0.624\,(11) & 0.650\,(13) \\
    Nep (nuFormer-style) & 0.695\,(15) & 0.719\,(4) & 0.738\,(4) & 0.707\,(6) & 0.731\,(6) & 0.746\,(2) & 0.602\,(16) & 0.628\,(14) & 0.649\,(4) \\
    CoLES & 0.700\,(12) & 0.721\,(2) & 0.738\,(4) & 0.749\,(12) & 0.764\,(5) & 0.766\,(6) & 0.585\,(21) & 0.617\,(9) & 0.647\,(11) \\
    Mlm-Cl & 0.700\,(12) & 0.716\,(4) & 0.740\,(3) & 0.709\,(14) & 0.727\,(5) & 0.749\,(3) & 0.602\,(18) & 0.635\,(6) & 0.648\,(6) \\
    JF-rand (no pretraining) & 0.706\,(9) & 0.723\,(6) & 0.738\,(2) & 0.681\,(16) & 0.714\,(17) & 0.758\,(9) & 0.590\,(19) & 0.617\,(15) & 0.648\,(3) \\
    JF & 0.709\,(6) & 0.726\,(5) & 0.737\,(1) & 0.738\,(33) & 0.763\,(15) & 0.770\,(3) & 0.615\,(14) & 0.634\,(11) & 0.645\,(7) \\
    TabPFN v2 & 0.725\,(5) & 0.737\,(5) & -- & 0.720\,(1) & 0.734\,(4) & -- & 0.631\,(22) & 0.655\,(8) & -- \\
    Expert features alone & 0.704\,(12) & 0.722\,(5) & 0.740\,(2) & 0.761\,(3) & 0.763\,(7) & 0.766\,(5) & 0.610\,(41) & 0.642\,(9) & 0.650\,(7) \\
    Agg + expert features & 0.709\,(7) & 0.719\,(4) & 0.739\,(4) & 0.762\,(6) & 0.760\,(5) & 0.770\,(7) & 0.615\,(36) & 0.638\,(6) & 0.653\,(7) \\
    Agg + Trans & 0.702\,(15) & 0.722\,(7) & 0.740\,(2) & 0.763\,(6) & 0.758\,(4) & 0.767\,(4) & 0.599\,(20) & 0.626\,(9) & 0.653\,(8) \\
    Agg + Trans + Mlm-Cl & 0.699\,(15) & 0.717\,(3) & 0.738\,(2) & 0.754\,(5) & 0.756\,(5) & 0.763\,(3) & 0.609\,(17) & 0.632\,(8) & 0.649\,(2) \\
    JF + Trans & 0.686\,(17) & 0.720\,(5) & 0.733\,(3) & 0.750\,(25) & 0.765\,(6) & 0.765\,(11) & 0.616\,(14) & 0.620\,(24) & 0.650\,(5) \\
    TBS & 0.719\,(8) & 0.733\,(5) & 0.744\,(1) & 0.763\,(9) & 0.766\,(7) & 0.773\,(4) & 0.618\,(16) & 0.638\,(8) & 0.654\,(5) \\
    TBS-R & 0.702\,(16) & 0.721\,(6) & 0.740\,(2) & 0.763\,(6) & 0.762\,(4) & 0.769\,(6) & 0.599\,(17) & 0.624\,(11) & 0.651\,(7) \\
    TBS-R, no pretraining & 0.701\,(14) & 0.719\,(11) & 0.740\,(2) & 0.763\,(6) & 0.759\,(4) & 0.766\,(3) & 0.590\,(15) & 0.625\,(11) & 0.653\,(8) \\
    TBS-R, no Trans in offset & 0.703\,(12) & 0.722\,(4) & 0.742\,(3) & 0.709\,(16) & 0.761\,(3) & 0.762\,(5) & 0.599\,(18) & 0.623\,(14) & 0.647\,(5) \\
    TBS-R, Nep encoder & 0.702\,(15) & 0.719\,(6) & 0.740\,(2) & 0.764\,(6) & 0.760\,(3) & 0.769\,(5) & 0.605\,(22) & 0.635\,(3) & 0.654\,(6) \\
    Nep, field encoder & 0.698\,(16) & 0.715\,(4) & 0.737\,(4) & 0.714\,(10) & 0.728\,(4) & 0.747\,(4) & 0.603\,(13) & 0.628\,(12) & 0.645\,(3) \\
    JF, field encoder & 0.710\,(12) & 0.725\,(3) & 0.737\,(2) & 0.748\,(19) & 0.767\,(10) & 0.768\,(7) & 0.603\,(17) & 0.636\,(15) & 0.650\,(8) \\
    JF + Trans, field encoder & 0.690\,(11) & 0.718\,(4) & 0.731\,(4) & 0.743\,(31) & 0.770\,(9) & 0.770\,(5) & 0.614\,(26) & 0.630\,(21) & 0.648\,(6) \\
    TBS, field encoder & 0.719\,(10) & 0.733\,(5) & 0.743\,(1) & 0.766\,(7) & 0.767\,(5) & 0.771\,(6) & 0.613\,(15) & 0.639\,(8) & 0.656\,(5) \\
    \bottomrule
  \end{tabular}}
\end{table}

\begin{table}[p]
  \centering
  \caption{Test AUC, mean over seeds, with the standard deviation in thousandths in parentheses: MBD-mini, transactions-only scenario. Columns give the label budget (1k, 4k, all).}
  \scriptsize
  \setlength{\tabcolsep}{2.2pt}
  \begin{tabular}{lrrrrrr}
    \toprule
    & \multicolumn{3}{c}{Any product} & \multicolumn{3}{c}{Product 1} \\
    \cmidrule(lr){2-4} \cmidrule(lr){5-7}
    Method & 1k & 4k & all & 1k & 4k & all \\
    \midrule
    Agg (baseline) & 0.678\,(18) & 0.711\,(7) & 0.741\,(3) & 0.656\,(25) & 0.682\,(12) & 0.719\,(6) \\
    AE, Transformer, masked numeric & 0.681\,(26) & 0.712\,(10) & 0.740\,(3) & 0.654\,(27) & 0.693\,(12) & 0.724\,(4) \\
    AE, Transformer, + one-hot & 0.671\,(55) & 0.713\,(6) & 0.740\,(2) & 0.662\,(34) & 0.686\,(9) & 0.724\,(11) \\
    AE, MLP, whole sequence & 0.682\,(22) & 0.714\,(15) & 0.739\,(5) & 0.658\,(22) & 0.691\,(24) & 0.733\,(7) \\
    AE, MLP, masked events & 0.679\,(23) & 0.711\,(6) & 0.741\,(2) & 0.646\,(28) & 0.687\,(21) & 0.726\,(6) \\
    MLM (TabBERT-style) & 0.671\,(21) & 0.711\,(5) & 0.745\,(5) & 0.651\,(43) & 0.690\,(14) & 0.736\,(7) \\
    Nep (nuFormer-style) & 0.682\,(25) & 0.718\,(7) & 0.749\,(3) & 0.653\,(30) & 0.690\,(11) & 0.735\,(10) \\
    CoLES & 0.673\,(19) & 0.711\,(5) & 0.751\,(4) & 0.654\,(24) & 0.681\,(9) & 0.738\,(6) \\
    Mlm-Cl & 0.670\,(23) & 0.716\,(10) & 0.752\,(3) & 0.655\,(27) & 0.684\,(25) & 0.727\,(5) \\
    GRU & 0.647\,(47) & 0.703\,(7) & 0.744\,(9) & 0.604\,(44) & 0.677\,(31) & 0.735\,(7) \\
    JF-rand (no pretraining) & 0.705\,(17) & 0.725\,(10) & 0.752\,(3) & 0.677\,(45) & 0.705\,(34) & 0.736\,(6) \\
    JF & 0.696\,(21) & 0.728\,(10) & 0.749\,(6) & 0.634\,(66) & 0.702\,(15) & 0.741\,(6) \\
    TabPFN v2 & 0.700\,(15) & 0.732\,(3) & -- & 0.644\,(35) & 0.715\,(25) & -- \\
    Agg + Trans & 0.682\,(14) & 0.714\,(10) & 0.743\,(7) & 0.665\,(19) & 0.700\,(26) & 0.716\,(8) \\
    Agg + Trans + Mlm-Cl & 0.680\,(23) & 0.723\,(7) & 0.756\,(3) & 0.645\,(39) & 0.692\,(20) & 0.729\,(5) \\
    JF + Trans & 0.693\,(26) & 0.721\,(10) & 0.752\,(3) & 0.680\,(14) & 0.699\,(11) & 0.744\,(6) \\
    TBS & 0.703\,(17) & 0.733\,(6) & 0.754\,(4) & 0.679\,(34) & 0.717\,(19) & 0.740\,(9) \\
    TBS-R & 0.683\,(20) & 0.723\,(11) & 0.758\,(7) & 0.671\,(20) & 0.707\,(28) & 0.733\,(11) \\
    TBS-R, no pretraining & 0.686\,(17) & 0.721\,(10) & 0.751\,(12) & 0.666\,(20) & 0.709\,(18) & 0.730\,(8) \\
    TBS-R, no Trans in offset & 0.672\,(17) & 0.722\,(12) & 0.754\,(6) & 0.667\,(20) & 0.684\,(20) & 0.732\,(11) \\
    TBS-R, Nep encoder & 0.688\,(20) & 0.723\,(12) & 0.755\,(6) & 0.670\,(15) & 0.710\,(19) & 0.732\,(7) \\
    Nep, field encoder & 0.669\,(27) & 0.721\,(7) & 0.755\,(3) & 0.651\,(33) & 0.701\,(8) & 0.733\,(5) \\
    JF, field encoder & 0.689\,(27) & 0.724\,(23) & 0.748\,(6) & 0.654\,(86) & 0.704\,(19) & 0.739\,(11) \\
    JF + Trans, field encoder & 0.692\,(33) & 0.732\,(9) & 0.751\,(4) & 0.703\,(24) & 0.713\,(13) & 0.738\,(6) \\
    TBS, field encoder & 0.699\,(18) & 0.733\,(11) & 0.755\,(5) & 0.687\,(37) & 0.717\,(15) & 0.739\,(8) \\
    \bottomrule
  \end{tabular}
\end{table}

\begin{table}[p]
  \centering
  \caption{Test AUC, mean over seeds, with the standard deviation in thousandths in parentheses: Client A, late payment. Columns give the label budget (1k, 4k, all).}
  \scriptsize
  \setlength{\tabcolsep}{2.2pt}
  \begin{tabular}{lrrrrrr}
    \toprule
    & \multicolumn{3}{c}{Transactions only} & \multicolumn{3}{c}{Full} \\
    \cmidrule(lr){2-4} \cmidrule(lr){5-7}
    Method & 1k & 4k & all & 1k & 4k & all \\
    \midrule
    Agg (baseline) & 0.820\,(8) & 0.834\,(9) & 0.846\,(3) & 0.821\,(8) & 0.833\,(6) & 0.848\,(2) \\
    AE, Transformer, masked numeric & 0.821\,(7) & 0.833\,(7) & 0.844\,(2) & 0.817\,(9) & 0.829\,(7) & 0.848\,(2) \\
    AE, Transformer, + one-hot & 0.825\,(3) & 0.835\,(7) & 0.846\,(3) & 0.824\,(1) & 0.834\,(6) & 0.848\,(1) \\
    AE, MLP, whole sequence & 0.817\,(16) & 0.835\,(9) & 0.844\,(2) & 0.816\,(16) & 0.832\,(11) & 0.846\,(4) \\
    AE, MLP, masked events & 0.815\,(16) & 0.831\,(9) & 0.844\,(3) & 0.816\,(22) & 0.832\,(8) & 0.848\,(3) \\
    MLM (TabBERT-style) & 0.819\,(6) & 0.833\,(4) & 0.844\,(2) & 0.818\,(8) & 0.833\,(5) & 0.844\,(4) \\
    Nep (nuFormer-style) & 0.825\,(8) & 0.834\,(9) & 0.848\,(2) & 0.825\,(10) & 0.835\,(7) & 0.850\,(3) \\
    CoLES & 0.829\,(11) & 0.839\,(7) & 0.851\,(1) & 0.826\,(13) & 0.838\,(3) & 0.849\,(3) \\
    Mlm-Cl & 0.825\,(8) & 0.834\,(7) & 0.848\,(3) & 0.824\,(11) & 0.834\,(8) & 0.848\,(2) \\
    GRU & 0.827\,(8) & 0.833\,(11) & 0.853\,(2) & -- & -- & -- \\
    JF-rand (no pretraining) & 0.811\,(6) & 0.828\,(5) & 0.846\,(4) & 0.815\,(7) & 0.830\,(6) & 0.849\,(3) \\
    JF & 0.836\,(6) & 0.836\,(13) & 0.851\,(2) & 0.836\,(7) & 0.844\,(3) & 0.854\,(1) \\
    TabPFN v2 & 0.829\,(10) & 0.837\,(4) & -- & 0.829\,(9) & 0.840\,(4) & -- \\
    Agg + Trans & 0.822\,(3) & 0.834\,(8) & 0.848\,(2) & 0.824\,(10) & 0.832\,(6) & 0.846\,(4) \\
    Agg + Trans + Mlm-Cl & 0.826\,(8) & 0.833\,(7) & 0.848\,(2) & 0.822\,(11) & 0.836\,(6) & 0.847\,(5) \\
    JF + Trans & 0.837\,(10) & 0.835\,(19) & 0.853\,(1) & 0.840\,(9) & 0.842\,(6) & 0.854\,(2) \\
    TBS & 0.837\,(4) & 0.842\,(7) & 0.854\,(1) & 0.838\,(5) & 0.845\,(4) & 0.855\,(1) \\
    TBS-R & 0.824\,(4) & 0.839\,(11) & 0.848\,(2) & 0.827\,(5) & 0.835\,(11) & 0.845\,(2) \\
    TBS-R, no pretraining & 0.823\,(4) & 0.838\,(5) & 0.848\,(2) & 0.824\,(11) & 0.832\,(9) & 0.845\,(3) \\
    TBS-R, no Trans in offset & 0.823\,(11) & 0.838\,(11) & 0.848\,(4) & 0.823\,(9) & 0.835\,(10) & 0.848\,(3) \\
    TBS-R, Nep encoder & 0.822\,(6) & 0.843\,(6) & 0.851\,(1) & 0.826\,(8) & 0.837\,(6) & 0.847\,(3) \\
    Nep, field encoder & 0.826\,(12) & 0.836\,(4) & 0.848\,(2) & 0.823\,(9) & 0.837\,(6) & 0.849\,(3) \\
    JF, field encoder & 0.838\,(11) & 0.842\,(4) & 0.855\,(1) & 0.835\,(5) & 0.841\,(15) & 0.855\,(2) \\
    JF + Trans, field encoder & 0.836\,(9) & 0.843\,(2) & 0.856\,(2) & 0.838\,(11) & 0.842\,(6) & 0.854\,(1) \\
    TBS, field encoder & 0.838\,(6) & 0.844\,(4) & 0.855\,(1) & 0.838\,(6) & 0.843\,(9) & 0.855\,(1) \\
    \bottomrule
  \end{tabular}
\end{table}

\begin{table}[p]
  \centering
  \caption{Test AUC, mean over seeds, with the standard deviation in thousandths in parentheses: Client B, churn (6m). Columns give the label budget (1k, 4k, all).}
  \scriptsize
  \setlength{\tabcolsep}{2.2pt}
  \begin{tabular}{lrrrrrr}
    \toprule
    & \multicolumn{3}{c}{Transactions only} & \multicolumn{3}{c}{Full} \\
    \cmidrule(lr){2-4} \cmidrule(lr){5-7}
    Method & 1k & 4k & all & 1k & 4k & all \\
    \midrule
    Agg (baseline) & 0.575\,(8) & 0.594\,(12) & 0.609\,(3) & 0.579\,(8) & 0.601\,(7) & 0.622\,(5) \\
    AE, Transformer, masked numeric & 0.577\,(8) & 0.597\,(11) & 0.610\,(4) & 0.583\,(10) & 0.605\,(9) & 0.621\,(5) \\
    AE, Transformer, + one-hot & 0.571\,(11) & 0.597\,(13) & 0.614\,(4) & 0.576\,(10) & 0.599\,(13) & 0.622\,(4) \\
    AE, MLP, whole sequence & 0.570\,(11) & 0.595\,(10) & 0.611\,(3) & 0.570\,(11) & 0.601\,(7) & 0.622\,(5) \\
    AE, MLP, masked events & 0.574\,(8) & 0.594\,(6) & 0.611\,(4) & 0.575\,(14) & 0.602\,(5) & 0.619\,(10) \\
    MLM (TabBERT-style) & 0.568\,(10) & 0.600\,(10) & 0.616\,(6) & 0.567\,(21) & 0.602\,(14) & 0.622\,(5) \\
    Nep (nuFormer-style) & 0.576\,(15) & 0.602\,(11) & 0.622\,(5) & 0.577\,(14) & 0.605\,(14) & 0.628\,(6) \\
    CoLES & 0.573\,(12) & 0.608\,(7) & 0.628\,(2) & 0.574\,(20) & 0.605\,(9) & 0.632\,(3) \\
    Mlm-Cl & 0.584\,(6) & 0.610\,(6) & 0.623\,(2) & 0.582\,(6) & 0.608\,(12) & 0.629\,(5) \\
    GRU & 0.569\,(9) & 0.602\,(14) & 0.631\,(4) & -- & -- & -- \\
    JF-rand (no pretraining) & 0.557\,(12) & 0.578\,(15) & 0.614\,(7) & 0.555\,(11) & 0.586\,(13) & 0.614\,(6) \\
    JF & 0.565\,(20) & 0.599\,(15) & 0.617\,(9) & 0.567\,(19) & 0.600\,(14) & 0.621\,(12) \\
    TabPFN v2 & 0.583\,(15) & 0.603\,(12) & -- & 0.586\,(16) & 0.608\,(14) & -- \\
    Expert features alone & 0.570\,(18) & 0.591\,(7) & 0.592\,(5) & -- & -- & -- \\
    Agg + Trans & 0.582\,(15) & 0.601\,(11) & 0.620\,(6) & 0.581\,(9) & 0.608\,(10) & 0.624\,(3) \\
    Agg + Trans + Mlm-Cl & 0.586\,(9) & 0.614\,(8) & 0.627\,(5) & 0.588\,(10) & 0.616\,(8) & 0.632\,(3) \\
    JF + Trans & 0.570\,(17) & 0.596\,(15) & 0.619\,(10) & 0.572\,(20) & 0.596\,(17) & 0.620\,(12) \\
    TBS & 0.579\,(15) & 0.607\,(11) & 0.626\,(5) & 0.580\,(8) & 0.611\,(11) & 0.630\,(7) \\
    TBS-R & 0.590\,(14) & 0.607\,(9) & 0.627\,(8) & 0.591\,(11) & 0.617\,(3) & 0.631\,(6) \\
    TBS-R, no pretraining & 0.578\,(14) & 0.598\,(12) & 0.620\,(6) & 0.577\,(11) & 0.605\,(9) & 0.624\,(3) \\
    TBS-R, no Trans in offset & 0.586\,(11) & 0.607\,(6) & 0.622\,(6) & 0.586\,(6) & 0.611\,(4) & 0.630\,(8) \\
    TBS-R, Nep encoder & 0.587\,(18) & 0.606\,(10) & 0.623\,(8) & 0.588\,(13) & 0.610\,(11) & 0.628\,(8) \\
    Nep, field encoder & 0.570\,(12) & 0.597\,(10) & 0.618\,(8) & 0.573\,(11) & 0.609\,(10) & 0.624\,(8) \\
    JF, field encoder & 0.569\,(17) & 0.595\,(14) & 0.626\,(9) & 0.560\,(18) & 0.596\,(9) & 0.621\,(9) \\
    JF + Trans, field encoder & 0.575\,(16) & 0.592\,(22) & 0.625\,(8) & 0.571\,(17) & 0.599\,(16) & 0.625\,(12) \\
    TBS, field encoder & 0.581\,(17) & 0.606\,(10) & 0.631\,(8) & 0.577\,(7) & 0.611\,(7) & 0.631\,(5) \\
    \bottomrule
  \end{tabular}
\end{table}

\end{document}